\documentclass[accepted]{uai2023} 

\usepackage[american]{babel}
\usepackage[utf8]{inputenc} 
\usepackage[T1]{fontenc}    
\usepackage{hyperref}
\hypersetup{colorlinks = true,
            linkcolor = blue,
            urlcolor  = blue,
            citecolor = blue,
            anchorcolor = blue}
\usepackage{url}            
\usepackage{booktabs}       
\usepackage{amsfonts}       
\usepackage{nicefrac}       
\usepackage{microtype}      
\usepackage{xcolor}         
\usepackage{algorithm2e}
\RestyleAlgo{ruled}
\usepackage{algpseudocode}
\newcommand{\appref}[1]{(\ref*{#1})}
\usepackage{amsmath}
\usepackage{amssymb}
\usepackage{mathtools}
\usepackage{amsthm}
\usepackage{microtype}
\usepackage{graphicx}
\usepackage{booktabs} 
\usepackage{subfigure}
\usepackage{algorithm2e}
\RestyleAlgo{ruled}
\usepackage{algpseudocode}
\usepackage{tabularx}
\usepackage{wrapfig}
\newcolumntype{b}{X}
\newcolumntype{s}{>{\hsize=.5\hsize}X}

\usepackage{placeins}
\theoremstyle{plain}
\newtheorem{theorem}{Theorem}[section]
\newtheorem{proposition}[theorem]{Proposition}

\theoremstyle{definition}

\theoremstyle{remark}

\usepackage{natbib} 
\usepackage{mathtools} 
\usepackage{booktabs} 
\usepackage{tikz} 

\title{Boosting Exploration in Multi-Task Reinforcement Learning \\ using Adversarial Networks}

\author[1]{Ramnath Kumar \thanks{Work done during an internship at Mila; Correspondence author: ramnathk@google.com.}}
\author[2]{Tristan Deleu}
\author[2,3]{Yoshua Bengio}
\affil[1]{%
    Google Research, India
}
\affil[2]{%
    Mila, Qu\'ebec Artificial Intelligence Institute, Université de Montréal
}
\affil[3]{%
    CIFAR, IVADO
  }
  
\begin{document}
\maketitle

\begin{abstract}
Advancements in reinforcement learning (RL) have been remarkable in recent years. However, the limitations of traditional training methods have become increasingly evident, particularly in meta-RL settings where agents face new, unseen tasks. Conventional training approaches are susceptible to failure in such situations as they need more robustness to adversity. Our proposed adversarial training regime for Multi-Task Reinforcement Learning (MT-RL) addresses the limitations of conventional training methods in RL, especially in meta-RL environments where the agent faces new tasks. The adversarial component challenges the agent, forcing it to improve its decision-making abilities in dynamic and unpredictable situations. This component operates without relying on manual intervention or domain-specific knowledge, making it a highly versatile solution. Experiments conducted in multiple MT-RL environments demonstrate that adversarial training leads to better exploration and a deeper understanding of the environment. The adversarial training regime for MT-RL presents a new perspective on training and development for RL agents and is a valuable contribution to the field.


\end{abstract}

\section{Introduction}

Recent years have seen tremendous progress in methods for reinforcement learning with the rise of ``Deep Reinforcement Learning'' \citep[DRL;][]{mnih2015human}. In robotics, DRL holds the promise of automatically learning flexible behaviors end-to-end while dealing with multidimensional data, as mentioned in \cite{arulkumaran2017brief}. The level has risen to such heights that our algorithms are capable of learning policies that can defeat human professionals in many different games such as Chess, Go (e.g., by AlphaZero \citep{silver2017mastering}), and many more complicated games such as Dota-2 (OpenAI, as presented in \cite{berner2019dota}), robot control \citep{gu2016continuous, lillicrap2015continuous, mordatch2015interactive}, and meta-learning \citep{zoph2016neural}.

Despite this recent progress, the predominant paradigm remains to train straightforward algorithms that need to be more robust to adversarial attacks, as shown in \cite{zhou1998essentials, garcia2015comprehensive}. 
In this work, we adopt a game-theoretic approach to adversarial training to enhance the learning process through adversarial competition. The idea is that learning through coevolution with adversaries can lead to a more robust and diverse set of skills. Unlike previous methods that relied on manual intervention to create adversarial examples to make models more robust to outliers  \citep{pinto2017robust, chen2019adversarial}, our proposed algorithm enables the model to learn adversarially without any manual intervention. Our approach learns adversarial skills that provide a better exploration strategy and make the model more adaptable to unseen tasks. This is accomplished by allowing the embedding network to learn multiple skill distributions without prior knowledge of the environment or the task. The result is a more efficient and flexible learning process.

\subsection{Problem Statement} 
In the field of Multi-Task Reinforcement Learning, also known as Meta-RL, the aim is to train an agent to perform a collection of tasks (${\tau_1,\tau_2,..\tau_n}$). Typically, all tasks are designed to have equal rewards for reaching the goal state. However, this presents a challenge since manual intervention is not feasible, and exploration is critical in this setting. A sub-optimal outcome is when the agent only learns to solve one task, $\tau_1$, ignoring all other tasks. In such a scenario, even if all the tasks have proportional rewards, the agent still lacks the incentive to explore and solve other tasks. To address this issue, we introduce an adversarial training regime that enhances robustness and encourages the exploration of the policy.

\begin{figure*}[hbt!]
\centering

\subfigure[Standard Learning Regime]{%
\raisebox{1cm}{\includegraphics[width=0.45\linewidth]{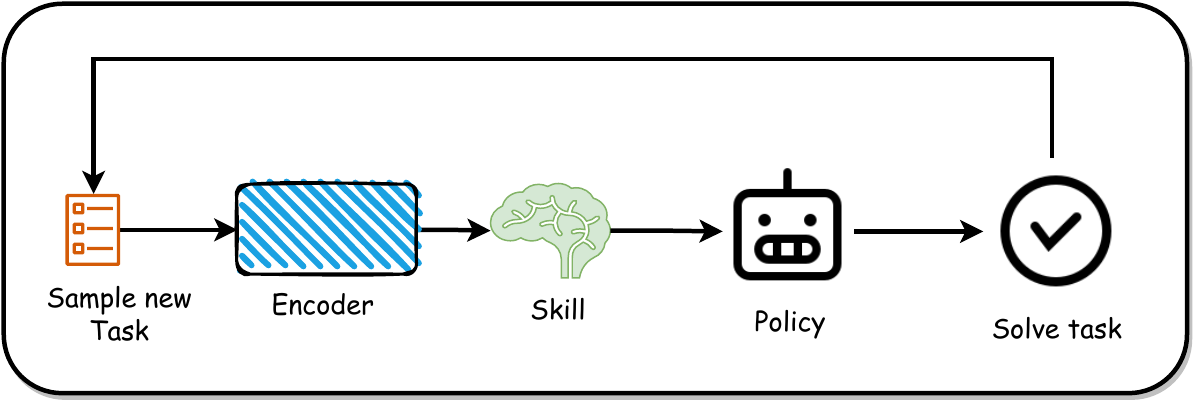}
\label{fig:teppo}}}
\quad
\subfigure[Adversarial Learning Regime]{%
\includegraphics[width=0.45\linewidth]{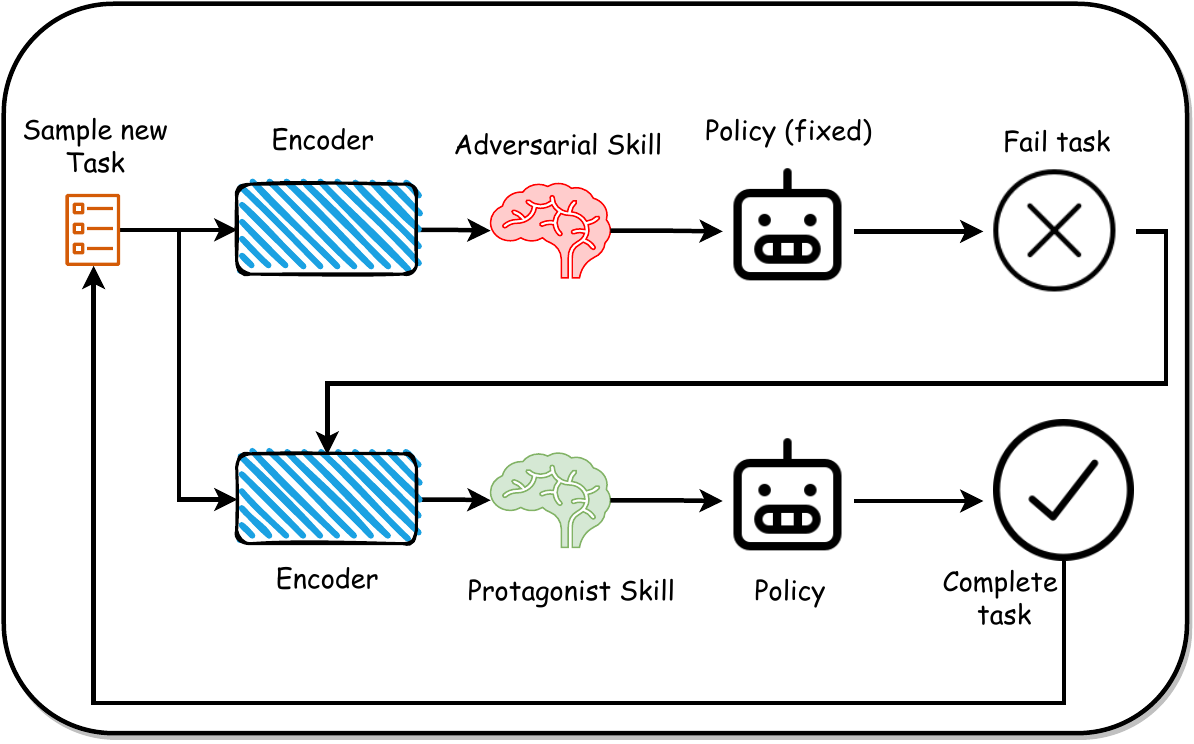}
\label{fig:ateppo}}

\caption{Figure~\ref{fig:teppo} and Figure~\ref{fig:ateppo} depict the straightforward (TE-PPO) and adversarial learning regime (ATE-PPO), respectively. Note that the same task is used in both learning cycles in the Adversarial Training Regime. Furthermore, the top sequence in Figure~\ref{fig:ateppo} tries to minimize reward/maximize regret when the policy is fixed (learning adversarial skills). In contrast, the bottom sequence of the same algorithm tries to correct this adversity by learning both the encoder and the policy.}

\label{fig:preliminary}
\end{figure*}

\subsection{Contributions} 
In this section, we present the main contributions of the paper: \newline
\textbf{A novel adversarial training regime:} The key contribution of our work is the introduction of a new adversarial training methodology for on-policy optimization in Multi-Task Reinforcement Learning (Meta-RL). This approach draws on game theory to produce diverse and robust task representations without requiring manual intervention or knowledge of the environments. This novel approach to adversarial training is the first of its kind in the Meta-RL field and is presented in detail in Section~\ref{adnet}. \newline
\textbf{Theoretical guarantees:} Our theoretical guarantees for the adversarial training regime serve as a useful tool for improving optimization in the process (see Appendix~\appref{the_conv}). The guarantees help ensure that the optimization is well-posed and provide insights into why the training regime works as it does. Additionally, the guarantees show that the learned skills are mutually exclusive to each task. This property is critical when dealing with unseen tasks, as it allows the skills learned to be easily transferred and adapted to these new tasks, leading to better generalization. These theoretical results provide a deeper understanding of the training process and the optimization of the policy. \newline
\textbf{Experimental Results:} Our proposed method has been thoroughly evaluated and shown to outperform the baseline in various simulated environments, including PointMass, 2D navigation, and robotic manipulation tasks from the Meta-World benchmark. This provides strong empirical evidence for the efficacy of our approach. (see Section~\ref{exp_results}) \newline
\textbf{Results on Domain Adaptation:} We also show that the model trained using the adversarial training regime leads to better generalization and more accessible domain adaptation to unseen and complex tasks, even in sparse rewards (see Section~\ref{transfer_mt5}). This result is per the added exploration factor the agent benefits from when trained in the adversarial regime. This exploration factor has been studied under the context of embedding efficiency of latent. (see Section~\ref{emb_efficiency}) \newline

\section{Preliminaries}
Before explaining our proposed adversarial training regime, it is crucial to understand the terminology, conventional reinforcement learning framework, and the concepts of two-player adversarial games that have influenced and motivated our work. This section will provide an overview of these critical elements to lay a solid foundation for the rest of the paper.

\subsection{Standard reinforcement learning in MDPs}

In this study, we focus on the field of reinforcement learning and, specifically, the application of this method in Markov Decision Processes (MDPs). In MDPs, the state of the agent, $s$, is represented as a continuous vector of size $S$, and the action taken by the agent, $a$, is represented as a continuous vector of size $A$. The transition from one state to another, $s_{t+1}$, is determined by the probability distribution $p(s_{t+1}|s_t,a_t)$, which depends on the current state, $s_t$, and the action taken, $a_t$. The policy that the agent follows, $\pi_{\theta}(a|s)$, is represented as a Gaussian distribution, with mean and diagonal covariance parameters that are determined by a neural network with parameters $\theta$. At each step, the agent receives a scalar reward, $r(s_t,a_t)$, and the goal is to maximize the expected sum of discounted rewards, $\mathbb{E}_{\tau _{\pi}}[\sum_{t=0}^{\infty}\gamma^tr(s_t,a_t)]$, where $\gamma$ is the discount factor. As a baseline for our work, we use TE-PPO \citep{hausman2018learning}, a reinforcement learning algorithm that utilizes latent skills for learning the policy. This serves as an ideal baseline to demonstrate the impact of our proposed adversarial training regime in the multi-task reinforcement learning setting since the only difference between our proposed approach and TE-PPO is the adversarial training regime we adopt.

\subsection{Two-player adversarial games}

The adversarial training regime we propose can be modeled as a two-player discounted Markov Game, where the discount factor is represented by $\gamma$. In this game, instead of learning the policy directly in an adversarial manner, we focus on learning the skills used by the policy in an adversarial way. The Markov Decision Process (MDP) remains the same, with the addition of a skill variable from the adversary that reduces the average discounted reward for a given task. Player 1, which consists of the Encoder and the policy network, acts as the protagonist and aims to solve the task. On the other hand, Player 2, which is only the Encoder network, serves as the adversary and learns the skills (\textit{"Worst Faithful Embedding"}) that can best deceive the protagonist for the given task. During Player 2's turn, the policy network remains frozen, and only the embedding network is trained. We call our algorithm \textbf{Adversarial TE-PPO} (or ATE-PPO) and describe our model in more detail in Section~\ref{adnet}. A brief overview of our algorithm is presented in Algorithm~\ref{alg:ateppo}. As discussed in the subsequent section, our end-objective function works in the minimax domain and differentiates our work from TE-PPO. 

Figure~\ref{fig:preliminary} depicts the high-level working of TE-PPO and ATE-PPO, respectively. Please refer to \cite{hausman2018learning} for more information on the TE-PPO algorithm.

\section{Proposed Approach}
\label{adnet}

Our approach, similar to the setup of TE-PPO, and comprised of three different networks which are trained in an adversarial training regime: an Encoder network ($\mathcal{E}$), a Policy network ($\mathbb{\pi}$), and an Inference network ($\mathbb{I}$). The Encoder tries to learn a skill embedding -- something unique for each task at hand, and the Policy network uses this embedding to select actions maximizing its reward. The adversarial modeling framework is most straightforward when the models are multilayer perceptrons. Instead of training two different sets of agents/policies in an adversarial fashion, we do so at the skill/encoder level. We will first learn the encoder parameters that provide the ``Worst Faithful Embedding'' - skills that reduce the discounted rewards for the given policy and minimize mutual information between the skills and one-hot encoding of tasks. By doing so, any spurious correlations between skills would disappear since we are learning for the worst case. We then learn the agent initialized at the given ``Worst Faithful Embedding'' to learn policies immune to adversarial perturbations.

Our mathematical formulation of the above intuition would follow along, as shown. To learn the skill embedding $z$, we represent a mapping between the one-hot encoded task embedding $\tau$ to a continuous skill space $z$ as $z = \mathcal{E}(\tau;\theta_{e})$; where $\mathcal{E}$ is a differentiable function represented by a multilayer perceptron with parameters $\theta_{e}$ and $z$ is the skill required for achieving task $\tau$.
To take advantage of these skill embeddings, we define a second multilayer perceptron $\mathbb{\pi}(z;\phi_{\pi})$ that helps the agent learn the actions to take and maximize its reward. Similar to \cite{hausman2018learning}, we also allow the Encoder to learn in this step to facilitate learning optimal skills for the given task.
To achieve this training regime, our model plays a two-player minimax game where the agent will try to learn a good policy and to embed with a protagonist loss function $\mathcal{L}_{\text{pro}}$:
\begin{equation}
\label{eq:loss_protagonist}
\begin{aligned}
\mathcal{L}_{\text{pro}} ={}&  - \mathbb{E}_{\pi,p_0,\tau\in \mathcal{T}} \left [Q_{\mathbb{\pi}}(s,a;\tau) \right ]
\end{aligned}
\end{equation}

However, the adversary perturbs the embedding slightly to fail at the task, which one cannot detect by computing mutual information between the task and embedding. The adversary will work with a fixed policy and an adversarial loss function $\mathcal{L}_{\text{adv}}$:

\begin{equation}
\label{eq:loss_adversary}
\begin{aligned}
\mathcal{L}_{\text{adv}} ={}& \mathbb{E}_{\pi,p_0,t\in \mathcal{T}} \left [Q_{\mathbb{\pi}}(s,a;\tau) \right ] - \\
{}& \alpha \mathbb{E}_{\tau\in \mathcal{T}}\left [ \mathcal{H}(z) + \mathcal{H}(\tau) - \mathcal{H}(\tau|z)\right ]
\end{aligned}
\end{equation}

Hence, one must learn a good policy and embedding that work despite these perturbations. Here $p_0(s_0)$ is the initial state distribution, $\alpha$ is a weighing term -- a trade-off between deceiving the agent network and minimizing the entropy terms denoted by $\mathcal{H}(.)$. Furthermore, the Q-function $Q_{\mathbb{\pi}}(s,a,\tau)$ is defined as:
\begin{equation}
\label{eq:q-general}
\begin{aligned}
    Q_{\mathbb{\pi}}(s,a,\tau) = \sum_{i=0}^{\infty} \gamma^i(r_\tau(s_i,a_i) + \alpha_0 \mathcal{H}[\pi(a_i|s_i,\tau)])
\end{aligned}
\end{equation}
where action $a_i$ is drawn from the distribution $\pi(.|s,\tau)$, and the new state is $s_{i+1} \sim p(s_{i+1}|a_i,s_i)$. Note that $\alpha_0$ is a constant and is treated as a hyperparameter. The first half of the equation defined by $\sum_{i=0}^{\infty} \gamma^i r_\tau(s_i,a_i)$ is the discounted expected returns and is common to our generic reinforcement learning algorithms. The latter is an entropy regularization term defined by $\sum_{i=0}^{\infty} \gamma^i \mathcal{H}[\pi(a_i|s_i,\tau)]$ which is conventionally applied to many policy gradient schemes, with the critical difference that it takes into account not only the entropy term of the current but also future actions. This approach has been used before by \cite{hausman2018learning}, but with a different objective function and trained in a straightforward reinforcement learning fashion, without any adversarial training dynamics. 

To apply this entropy regularization to our setting of latent variables (skill embeddings), some extra mathematical rigor is required. Borrowing from the toolkit of variational inference \citep{agakov2004algorithm}, we can construct a lower bound on the entropy term from Equation~\ref{eq:q-general} as (see Appendix~\appref{vi_lower_bound}) such that:
\begin{align}
\mathbb{E}_{\pi,p_0,\tau\in \mathcal{T}} \left [Q_{\mathbb{\pi}}(s,a,\tau) \right ] &= \mathbb{E}_{\pi(a,z|s,\tau)} \left [Q_{\pi}^{\varphi}(s,a;z,\tau)  \right ] \nonumber+ \\
{}& \alpha_1 \mathbb{E}_{\tau\in \mathcal{T}} \mathcal{H}[p(z|\tau)]
\end{align}
where, 
\begin{equation*}
Q_{\pi}^{\varphi}(s,a;z,\tau) =  \sum_{i=0}^{\infty} \gamma^i\widehat{r}(s_i,a_i,z,\tau)
\end{equation*}

Here, $s_{i+1}\sim p(s_{i+1}|a_i,s_i)$ and

\begin{equation*}
\begin{aligned}
\widehat{r}(s_i,a_i,z,\tau) =& \left [ r_t(s_i,a_i) + \alpha_2 \log[\mathbb{I}(z|a_i,s_i^H)] + \right . \\
& \left . \alpha_3\mathcal{H}[\pi(a|s,z)] \right ]
\end{aligned}
\end{equation*}



Note that this is where the inference network $\mathbb{I}$ comes into play. Furthermore, $H$ is the history of states saved in the buffer used for the inference. From the above equation, we simplify our objective function as follows and set $\alpha_1$ as $\alpha$, where we denote Mutual Information by $\mathcal{I}$, and Jensen Shannon Divergence by $\mathcal{JSD}$:
\begin{align}
\mathcal{L}_{\text{adv}} &=  \left [Q_{\pi}^{\varphi}(s,a;z,\tau)  \right ] - \\
& \alpha \mathbb{E}_{\tau\in \mathcal{T}}\left [\mathcal{H}(z) - \mathcal{H}(z|\tau) + \mathcal{H}(\tau) - \mathcal{H}(\tau|z)\right ]\nonumber\\
&= \left [Q_{\pi}^{\varphi}(s,a;z,\tau)  \right ]  - \alpha \mathbb{E}_{t\in \mathcal{T}}\left [ \mathcal{I}(z;\tau) + \mathcal{I}(t;\tau) \right ]\nonumber\\
&=  \left [Q_{\pi}^{\varphi}(s,a;z,\tau)  \right ]  - \alpha \mathbb{E}_{\tau\in \mathcal{T}}\left [ \mathcal{JSD}(z,\tau) \right ]\label{eq:q-final}
\end{align}

Although we use the objective presented in Equation~\ref{eq:q-final} to show theoretical guarantees (see Appendix~\appref{the_conv}), we simplify the equation further, making the implementation tractable (see Appendix~\appref{tract_entropy}). In our implementation, we use the modified objective presented in Appendix~\appref{tract_entropy} in an interleaved fashion. This is similar to adversarial networks in the on-policy setting. Furthermore, we would also like to extend this training regime to obtain a data-efficient, off-policy algorithm that could be applied to a natural robotic system in the future. (see Appendix~\appref{extending_off_policy})

\begin{algorithm}
\caption{Our proposed Adversarial Training Regime for Reinforcement Learning}\label{alg:ateppo}
\KwIn{Data sampling mechanism $\mathbb{T}$, Data $\mathcal{D}$, number of episodes $N$, Encoder Networks $\mathcal{E}$, Policy Network $\pi$, Number of Adversarial steps $\mathcal{A}$, and Number of Protagonist steps $\mathcal{P}$.}
$n \leftarrow 0$\;
$\tau_i \leftarrow \mathsf{One\-Hot}(i)$ \Comment{Create one-hot vectors of size k} \\
$\tau \leftarrow \begin{Bmatrix}
\tau_1 & \tau_2 ... & \tau_k
\end{Bmatrix}$ \Comment{Create Task Embedding} \\
\While{$n < N$}
{ $n \leftarrow n+1$\;
$\begin{Bmatrix}
\mathcal{T}_1 & \mathcal{T}_2 ... & \mathcal{T}_k
\end{Bmatrix} \leftarrow \mathbb{T}(\mathcal{D})$ \Comment{Sample k tasks from data distribution.} \\
$a \leftarrow 0$\;
$p \leftarrow 0$\;
\While{$a< \mathcal{A}$}{
$a\leftarrow a+1$\;
$\underset{\mathcal{E}}{\min} \text{ }\mathcal{L}_{\text{adv}}$\Comment{Learn adversarial skills keeping policy fixed} \\
}
\While{$p< \mathcal{P}$}{
$p\leftarrow p+1$\;
$\underset{\mathcal{E}, \pi}{\min} \text{ } \mathcal{L}_{\text{pro}}$\Comment{Learn protagonist skills learning both policy and encoder} \\
}
}

\end{algorithm}

\begin{figure*}[hbt!]
\centering
\subfigure[Learning Curve]{%
\includegraphics[width=0.44\linewidth]{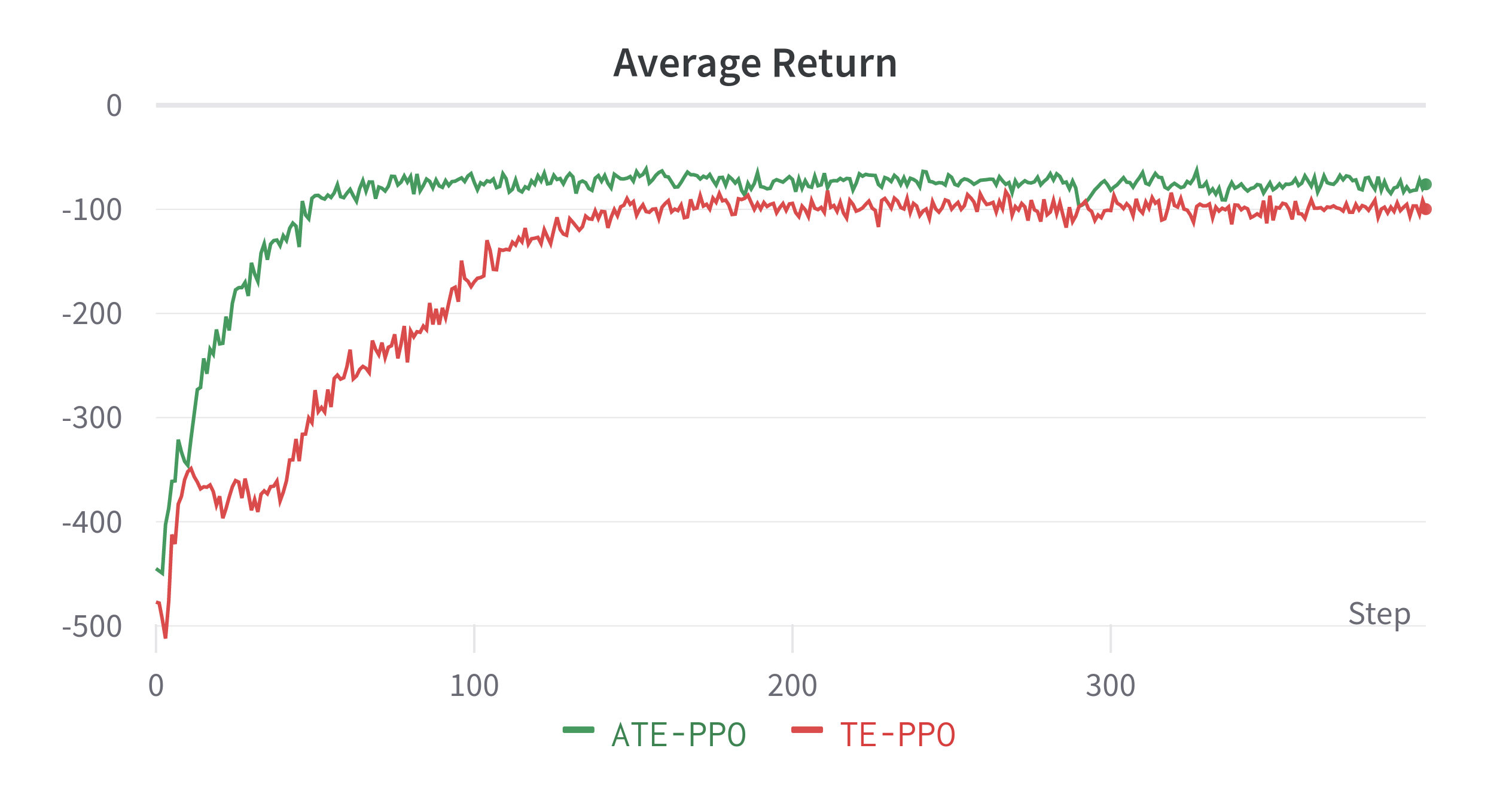}
\label{fig:lc_pointmass}}
\quad
\subfigure[TE-PPO]{%
\includegraphics[width=0.22\linewidth]{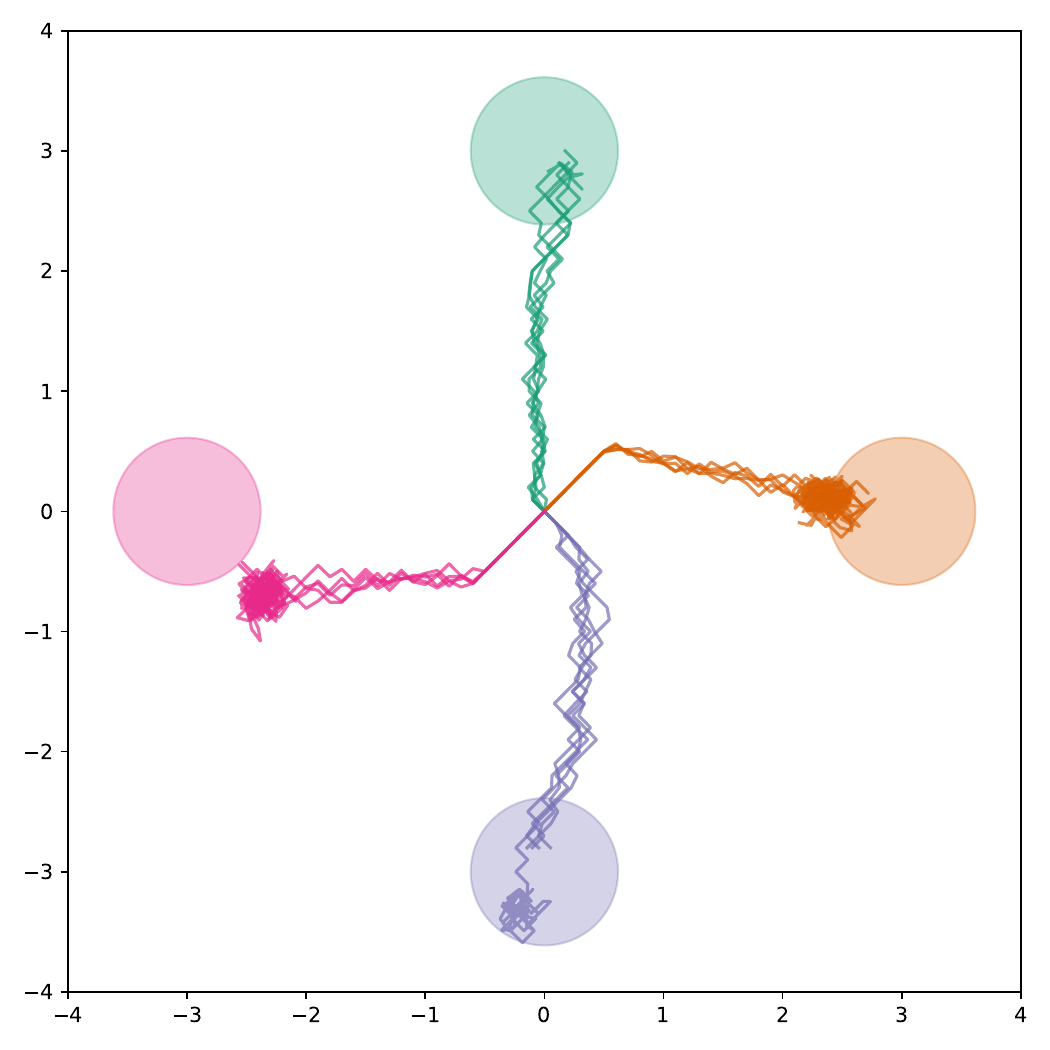}
\label{fig:te_pointmass}}
\quad
\subfigure[ATE-PPO]{%
\includegraphics[width=0.22\linewidth]{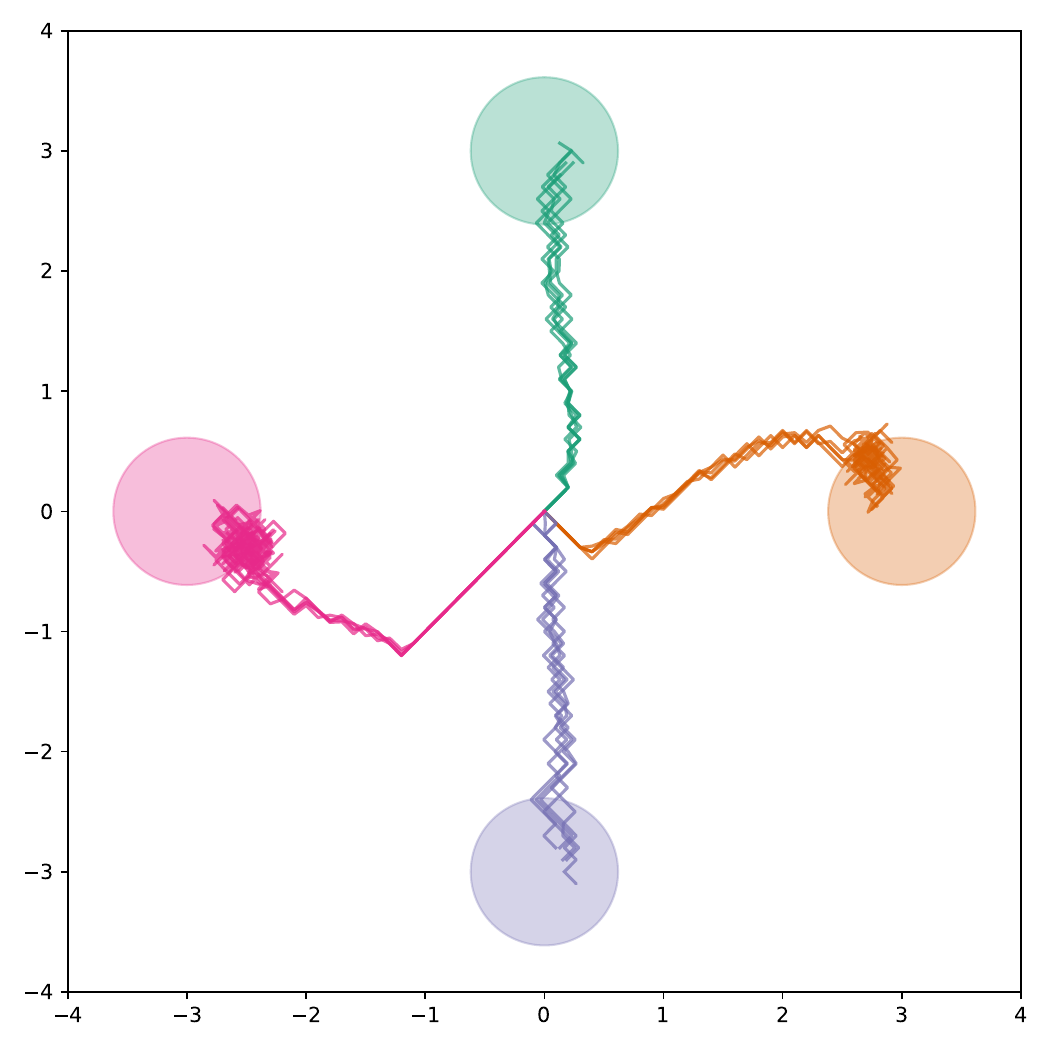}
\label{fig:ate_pointmass}}

\caption{Figure~\ref{fig:lc_pointmass} depicts the Average return on PointMass Environment using TE-PPO and ATE-PPO algorithms. The above illustration shows that the adversarial training regime boosts performance in the PointMass environment. Figure~\ref{fig:te_pointmass} and Figure~\ref{fig:ate_pointmass} show the resulting path taken by the agent trained on TE-PPO and ATE-PPO, respectively (10 random agents trying to solve each task, where the four colored circles are the goals).}
\label{fig:pointmass_results}
\end{figure*}

\section{Causal Perspective into RL}
\label{ace}

In this study, we suggest utilizing latent embeddings as skills to understand their impact on a task better. Our proposed framework calculates the average causal attribution of skills on the task. This is achieved by determining the average outcome difference from different skill-related interventions. The Average Causal Effect (ACE) of a skill (represented by variable $x$) on another outcome (represented by random variable $y$) is mathematically defined as:
\begin{equation}
    ACE^y_{do(x_i=\alpha)} = \mathbb{E}[y|do(x_i=\alpha)] - baseline_{x_i}
\end{equation}
The calculation of ACE involves finding two values: the interventional Expectation ($\mathbb{E}[y|do(x_i=\alpha)]$) and the baseline ($baseline_{x_i}$). The "do-operation" ($do$), as described in \citep{pearl2009causality}, represents fixing the value of a variable $x_i$ to a specific value $\alpha$.
\paragraph{Interventional Expectation} The Interventional Expectation is a value determined by considering the effect of a specific value of $x_i$, while all other variables are disregarded. It is mathematically defined as follows:
\begin{equation}
    \mathbb{E}[y|do(x_i=\alpha)] = \int_y yp(y|do(x_i=\alpha))dy 
\end{equation}
Evaluating the Interventional Expectation naively involves computing the average output values by fixing the value of $X_i$ to $\alpha$ and sampling the other input features from the empirical distribution. However, this method is time-consuming and assumes that the input features do not impact each other. To make the computation more efficient, we assume that the features are independent, given a specific intervention on one variable. This is justified as the learned latent features are time-agnostic. Instead of the naive method, we use the first-order Taylor expansion of the causal mechanism $f(x|do(x_i=\alpha))$. This expansion is performed around the mean vector $\mu = [\mu_1,\mu_2, \cdots, \mu_k]^T$ as follows:
\begin{equation}
\begin{aligned}
     f(x|{}&do(x_i=\alpha)) \approx  f(\mu|do(x_i=\alpha)) +\\
     {}& \nabla f(\mu|do(x_i=\alpha))^{\top}(x-\mu|do(x_i=\alpha))
\end{aligned}
\end{equation}
By taking expectations on both sides and disregarding the impact of all other input variables, we can calculate the Interventional Expectation:
\begin{equation}
    \mathbb{E}[f(x|do(x_i=\alpha))] \approx f(\mu|do(x_i=\alpha)),
\end{equation}

In this equation, $f(.)$ represents the average reward for a specific task, $x$ is the input vector or latent embedding, and $\mu$ is the mean of the input vectors when $x_i=\alpha$. It is important to note that the first-order term vanishes because $\mathbb{E}(x|x_i=\alpha) =\mu$. This eliminates the need for computing gradients through a step in reinforcement learning, as the transition function is generally non-differentiable and would require additional approximations or workarounds to make it differentiable. For this reason, we do not extend the Taylor approximation to the second-order or higher. The approximation of deep non-linear neural networks using Taylor's expansion has been explored in the context of causality \citep{chattopadhyay2019neural}. However, their overall goal was different from the reinforcement learning domain.

\paragraph{Baseline} An ideal baseline would be a point on the decision boundary where the predictions are neutral. However, as shown by the research conducted by \cite{kindermans2019reliability}, using a fixed reference baseline for attribution methods can result in a lack of affine invariance. Instead, we define the baseline as:
\begin{equation}
    baseline_{x_i} = \mathbb{E}_{x_i}[\mathbb{E}_y[y|do(x_i=\alpha)]]
\end{equation}
The baseline is defined as the expected value of $y$ given the intervention of fixing $x_i$ to $\widehat{x_i}$. If the expected value $y$ remains constant for all possible intervention values, the baseline would also be a constant, and the causal attribution, $ ACE^y_{do(x_i=\alpha)}$, would be equal to zero. The L1-norm of the causal attribution, $\mathbb{E}_{x_i}[\left |\mathbb{E}_y[y|do(x_i=\alpha)] \right |]$, can be used as a metric for average importance, which is then normalized for easier interpretation. This framework can be adapted to other models that work with latents, and the attributions were computed using the agent trained on ATE-PPO. The approximation may be rough, but the framework still provides valuable insights. (see Appendix~\appref{ace_results})

\section{Experimental Results and Discussions}
\label{exp_results}

We evaluate our approach in two domains in simulation: simple environments such as the PointMass task, 2-D navigation task, and a set of challenging robot manipulation tasks from Meta-World. We consider the Gaussian embedding space for all experiments for both algorithms (ATE-PPO and TE-PPO).

\subsection{Learning versatile skills}

We highlight this property of learning versatile skills from our experiments on relatively more straightforward environments such as the PointMass environment and 2-D Navigation environment. 

\paragraph{PointMass Environment} Similar to \cite{haarnoja2017reinforcement}, we present a didactic example of multi-goal PointMass tasks demonstrating the variability of solutions that our method can discover. In this experiment, we consider a case where four goals are located around the initial location, and each is equally important to the agent. This leads to a situation where multiple optimal policies exist for a single task. In addition, this task is challenging due to the sparsity of the rewards -- as soon as one solution is discovered, it becomes exceedingly difficult to keep exploring other goals. Due to these challenges, most existing DRL approaches would be content with finding a single solution.
Furthermore, even if a standard policy gradient approach discovered multiple goals, it would have no incentive to represent various solutions. Our proposed approach outperforms the baseline in this scenario, as depicted in Table~\ref{overall_results}. Figure~\ref{fig:pointmass_results} expresses more insights into the learning trajectory and the final agent behavior.

\begin{figure}[t]
\centering
\includegraphics[width=\linewidth]{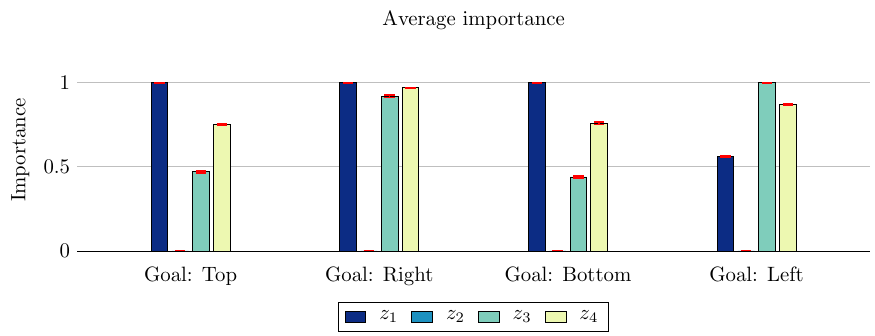}
\caption{The normalized importance of each latent feature on the PointMass tasks across five different seeds.}
\label{fig:imp_pointmass}
\end{figure}

To better understand the skill learned by our model, we use the causal framework proposed in Section~\ref{ace} to compute the average causal attribution of each feature of the skill vector. The average importance of each component is shown in Figure~\ref{fig:imp_pointmass}. This result is interesting because the feature $z_2$ consistently does not contribute to the agent's behavior. Although the latent space is ample, and the model has sufficient capacity, our policy network can learn only the minimum number of skills required to solve the task. 
Furthermore, we discuss the ACE plots in more detail in Appendix~\appref{appendix:point_mass}.

\begin{figure}[hbt!]
\centering

\subfigure[TE-PPO]{%
\includegraphics[width=0.45\linewidth]{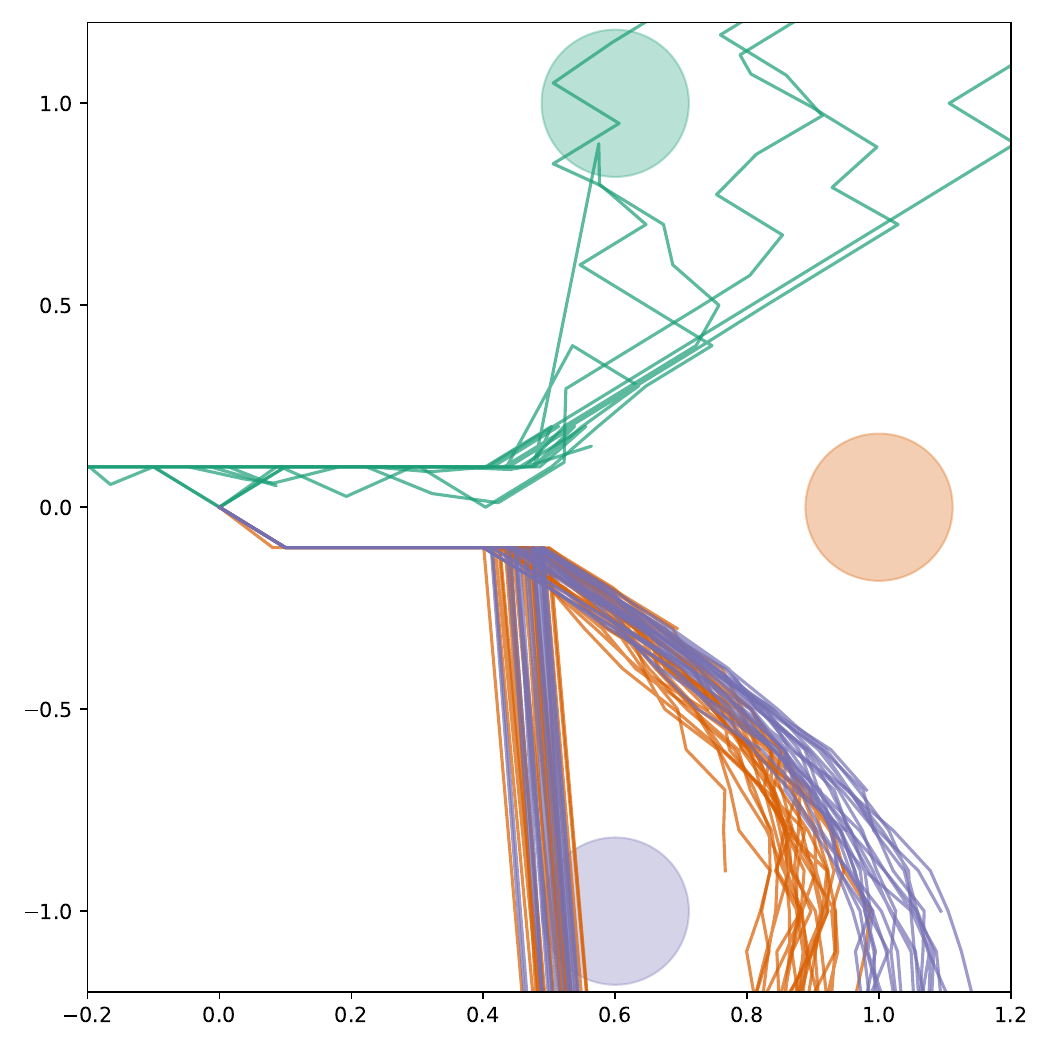}
\label{fig:te_navigation}}
\quad
\subfigure[ATE-PPO]{%
\includegraphics[width=0.45\linewidth]{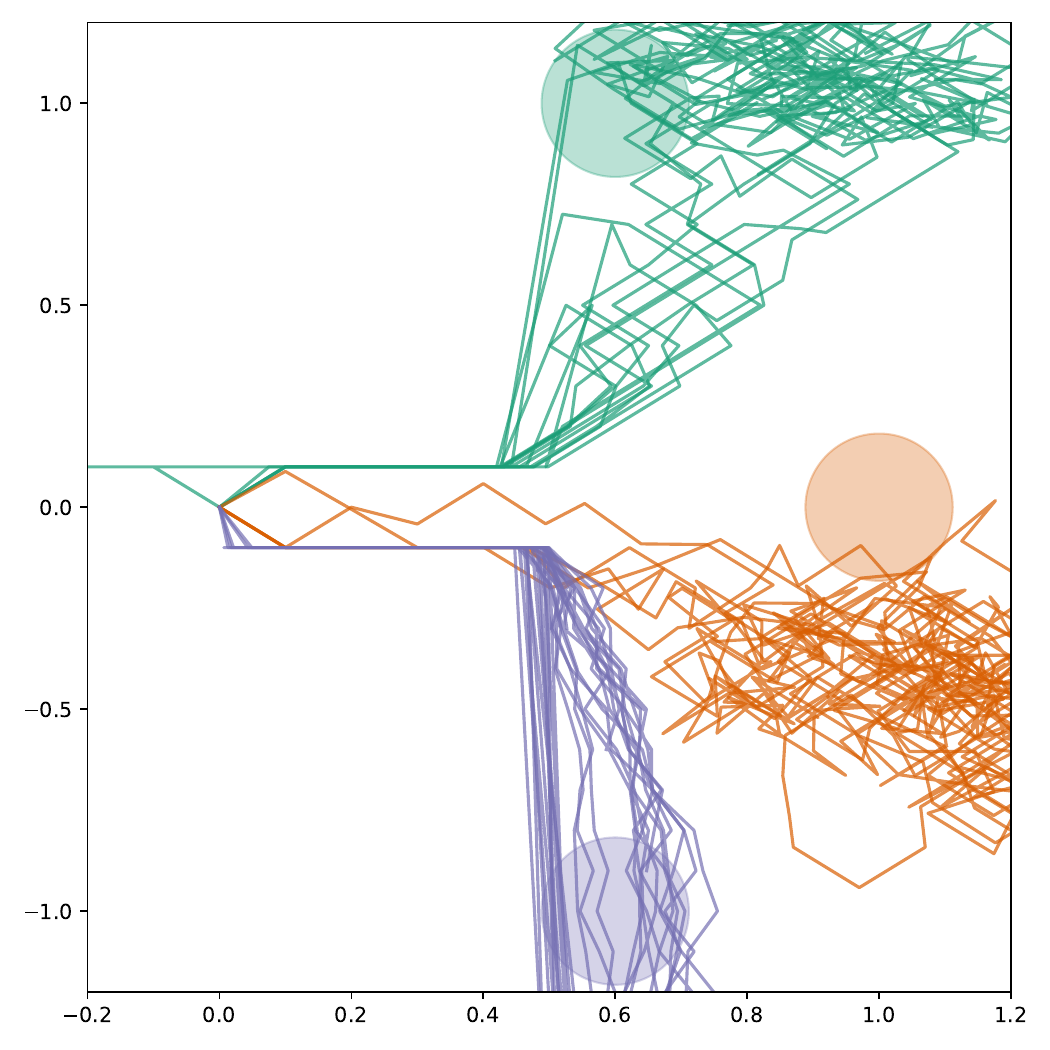}
\label{fig:ate_navigation}}

\caption{Figure~\ref{fig:te_navigation} and Figure~\ref{fig:ate_navigation} show the resulting trajectories by the agent trained on TE-PPO and ATE-PPO, respectively. Note that the three circles represent the goals for different tasks. The above illustration shows that the adversarial training regime boosts performance in the 2-D Navigation environment. Furthermore, the skills learned can create diversity and exploration of the agent in the future instead of the present.}
\label{fig:navigation_results}
\end{figure}

\paragraph{2-D Navigation Environment} Furthermore, we also experiment on the 2-D navigation task \citep{finn2017model} with a slight modification. At each time step, the 2D agent takes action (its velocity, clipped in $[-0.1, 0.1]$) and receives a penalty equal to its L2 distance to the goal position (i.e., the reward is ``$-$distance''). However, we restrict the agent's movement to a small corridor at the beginning, so the vertical state space is clipped between $[-0.2,0.2]$. In this paper, we sample three tasks to solve in the current experimental setup. The agent starts from $(0,0)$, and the goals of the tasks are to reach respective points in the Cartesian plane: $(1,0)$, $(0.6,1)$, and $(0.6,-1)$ denoted by various colors in Figure~\ref{fig:navigation_results}. We find that the ATE-PPO model can successfully learn the task and achieve better performance than the TE-PPO algorithm, as depicted in Table~\ref{overall_results}. We notice that our ATE-PPO model successfully learns skills that are \textcolor{blue}{\textbf{diverse in the future and not the present}}. Figure~\ref{fig:navigation_results} shows the resulting trajectories by the agent trained on TE-PPO and ATE-PPO, respectively. Furthermore, in more detail, we discuss the ACE plots in Appendix~\appref{appendix:navigation}. Our proposed approach outperforms the baseline, as depicted in Table~\ref{overall_results}. As shown in Table~\ref{overall_results}, our proposed approach outperforms the baseline TE-PPO by \textcolor{blue}{\textbf{67.64}} units.

\begin{figure*}[hbt!]
\centering
\includegraphics[width=\columnwidth]{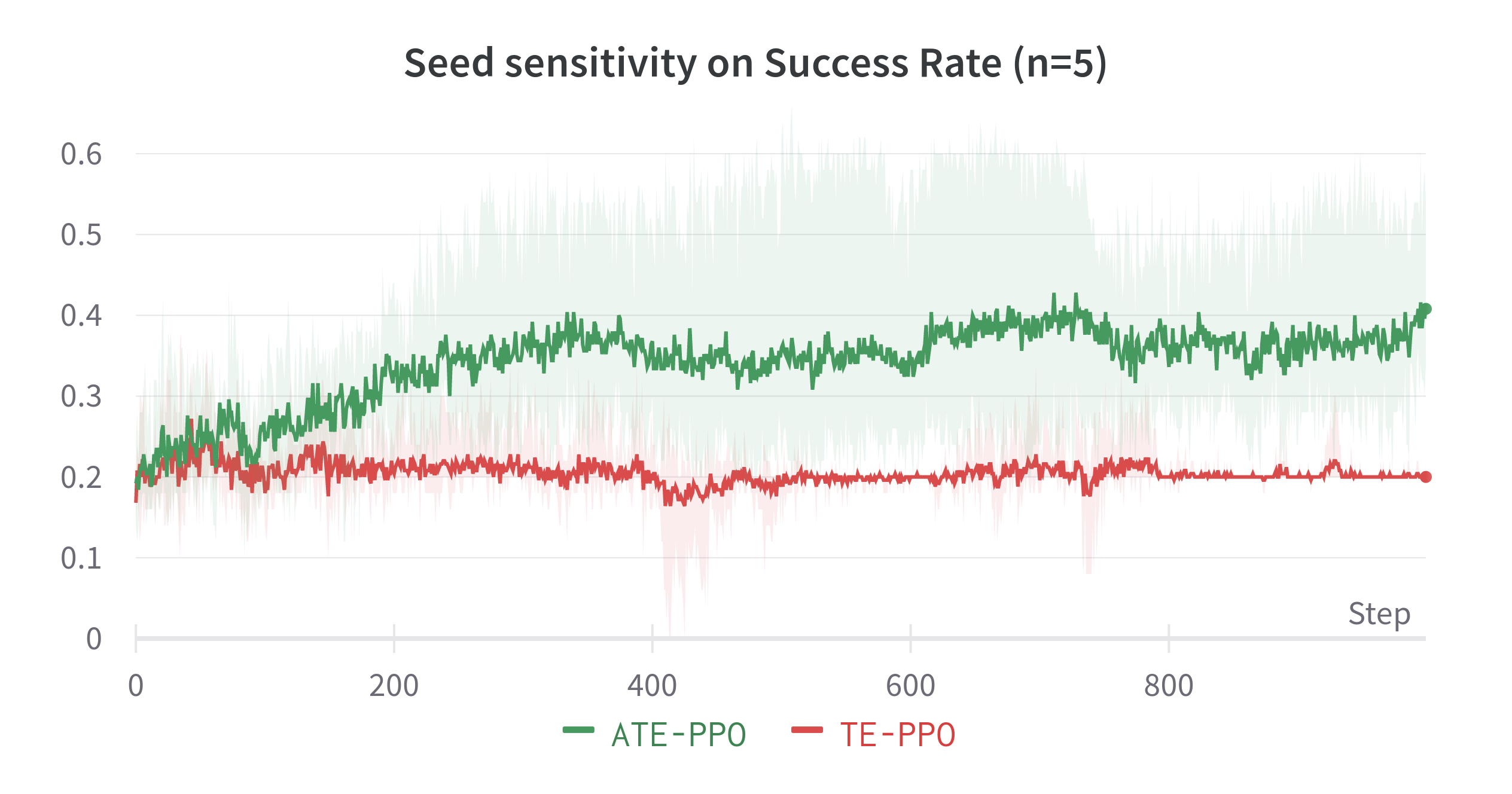}
\caption{Average success rate on MT5 Environment using TE-PPO and ATE-PPO algorithm. The above illustration shows that the adversarial training regime does offer a boost in performance in the MT5 environment.}
\label{fig:mt5_results}
\end{figure*}

\subsection{Learning optimal representations}

Next, we evaluate whether we can learn better representations with the help of the adversarial training regime. We create a pool of tasks from Meta-World with an overlapping skill set for most tasks in this problem. In our experiment, we set this overlapping skill as the act of ``pushing''. With this constraint, we create an environment MT5 by selecting five tasks for training: Push, Open window, Close window, Open drawer, and Close drawer. The multi-task evaluation tests the ability to learn multiple tasks simultaneously without accounting for generalization to new tasks. This task helps evaluate the efficiency of the skills learned by the model. ATE-PPO significantly improves TE-PPO's performance without significant hyperparameter tuning and using common network architecture across both experiments. TE-PPO achieves an average success rate of $0.2 \pm 0.01$, while our ATE-PPO achieves an average success rate of $0.41 \pm 0.08$, i.e., a success rate improvement of \textcolor{blue}{\textbf{+21\%}} over the baseline. We summarize more detailed results of the causal analysis of latent in Appendix~\appref{appendix:mt5}.

\subsection{Generalizing to unseen environments}
\label{transfer_mt5}
Next, we evaluate whether our model can adapt to a new unseen task with minimum adaptation steps. We use our pre-trained model from the previous section for this task, which was trained on the MT5 environment. We selected a diverse set of tasks for this experiment with varying degrees of domain shift:
\begin{itemize}
    \item \textbf{Push Wall:} The goal of this task is similar to the initial Push task from MT5, but we now also have to bypass a wall to reach the goal.
    \item \textbf{Coffee Button:} The goal of this task is to press a button on the coffee machine. We randomize the position of the coffee machine to avoid overfitting.
    \item \textbf{Push Back:} The goal of this task is to push the same object backward, similar to the Push task from MT5. We also randomize puck positions to avoid overfitting.
    \item \textbf{Faucet Open:} The goal of the task is to turn on the faucet. Specifically, we would like to rotate the faucet counter-clockwise and randomize the faucet positions to avoid overfitting.
\end{itemize}

We run the agent for all the above domain adaptation tasks for ten epochs. Although the domain-adapted code needs to be sufficiently trained to solve the task perfectly, our pre-trained ATE-PPO model outperforms the TE-PPO across most target tasks (if not competitive), as shown in Table~\ref{overall_results}. To summarize, our proposed approach ATE-PPO outperforms the baseline TE-PPO by an average return margin of \textcolor{blue}{\textbf{+38.44}}.
Furthermore, we run both models in the standard reinforcement learning setting without any adversarial training regime for better comparison. 

\begin{table}
\caption{Average return on simulated environments over five seeds.}
\label{overall_results}
\vskip 0.15in
\begin{center}
\begin{small}
\begin{tabular}{lcc}
\toprule
\textsc{Environment} & TE-PPO & ATE-PPO\\
\midrule
2D Navigation & -170.03 {\tiny $\pm$ 47.98} & \textbf{-102.39} {\tiny $\pm$ 44.44}\\
\midrule
\multicolumn{3}{l}{\textsc{Generalizing to unseen environments}}\\
\midrule
Push Wall    & 2.54 {\tiny $\pm$ 0.02}& \textbf{2.69} {\tiny $\pm$ 0.18}\\
Coffee Button & 37.21 {\tiny $\pm$ 0.82}& \textbf{80.91} {\tiny $\pm$ 2.80}\\
Push Back    & 0.85 {\tiny $\pm$ 0.01}& \textbf{0.89} {\tiny $\pm$ 0.05}\\
Faucet Open    & 474.41 {\tiny $\pm$ 269.73}& \textbf{584.29} {\tiny $\pm$ 237.43}\\
\bottomrule
\end{tabular}
\end{small}
\end{center}
\vskip -0.1in
\end{table}

\section{Embedding Efficiency}
\label{emb_efficiency}
We want our learned skills to be efficient and diverse in the latent embedding space. To evaluate the efficiency of our embeddings, we compute the diversity of these latents as the volume parallelopiped. Suppose our embedding vector is $z\sim \mathbb{R}^n$. If the volume the skills encompass is higher, more volume of the $\mathbb{R}^n$ space it covers is higher, and we can better adapt to hierarchical skills with the same embedding space. With the given intuition, we compute the efficiency of the embedding as the square of the volume parallelopiped by the embedding. Appendix~\appref{vol} discusses our approach to computing this efficiency in more detail. This approach is motivated by \cite{kumar2022effect}, which uses a similar approach to study the diversity of tasks in meta-learning. Our experiments show that the latent skills learned from the ATE-PPO algorithm are more efficient and diverse than those learned from the TE-PPO algorithm. We summarize our findings in Table~\ref{embedding_efficiency}. Furthermore, when computing the volume of the space it covers, we can scale each latent feature by a constant, if needed, without any loss of information.

\begin{table}[t]
\caption{Embedding Efficiency of learned latents}
\label{embedding_efficiency}
\vskip 0.15in
\begin{center}
\begin{small}
\begin{sc}
\begin{tabular}{lccr}
\toprule
Environment & TE-PPO & ATE-PPO\\
\midrule
PointMass   & 5.88$e^{-15}$ & \textbf{1.87}$\mathbf{e^{-2}}$\\
2D Navigation & 1.97$e^{-1}$ & \textbf{8.13}$\mathbf{e^{-1}}$\\
MT5   & 4.99$e^{-16}$ & \textbf{8.95 $\mathbf{e^{-8}}$}\\
\bottomrule
\end{tabular}
\end{sc}
\end{small}
\end{center}
\vskip -0.1in
\end{table}

\section{Related Works}

This section describes prior studies broadly related to the adversarial training regime of skills. Below, we divide prior research under two prominent subheadings: (i) Adversarial training regime and (ii)  Multi-Task reinforcement learning.

\subsection{Adversarial Training Regime}


In reinforcement learning, learned policies should be robust to uncertainty and parameter variation to ensure predictable behavior. Furthermore, learning policies should employ safe and effective exploration with improved sample efficiency to reduce the risk of costly failure. These issues have long been recognized and studied in reinforcement learning \citep{zhou1998essentials, garcia2015comprehensive}. However, these issues are exacerbated in deep RL using neural networks, which, while more expressible and flexible, often require more data to train and produce potentially unstable policies. 
In \cite{pinto2017supervision}, two supervised agents were trained, with one acting as an adversary for self-supervised learning, which showed an improvement in robot grasping. Other adversarial multiplayer approaches have been proposed, including \cite{heinrich2016deep}. However, these approaches require the training of two agents. Inspired by these, \cite{heinrich2016deep} proposed an approach to learning two policies for a single agent, one as a protagonist and the other as an adversary. However, this requires hard-coding parts of the agent where an adversarial force is applied to fool the model. Although these approaches lead to a more robust model, our goal in this paper is slightly different.

Instead, our goal is to propose an adversarial training regime without significant hard coding. We perform the adversarial training regime at the skill level, allowing better exploration. Since the same policy network is used in both cases, the policy understands the effect of wrong actions, increasing understanding of the environment due to the increased exploration of the state space.

\subsection{Multi-Task reinforcement learning}

Learning multiple tasks at once in reinforcement learning is quite common, with the intuition that more tasks would suggest higher diversity and hence better generalization in downstream tasks. This property is best showcased in work \citep{eysenbach2018diversity}, where they learn diverse skills without any reward function. Furthermore, sequential learning and the need to retain previously known skills have always been a focus \citep{rusu2016progressive, kirkpatrick2017overcoming}. In the space of multi-task reinforcement learning with neural networks, \cite{teh2017distral} proposed a framework that allows sharing knowledge across tasks via a task agnostic prior. Similarly, \cite{cabi2017intentional} uses off-policy learning to learn about many different tasks while following a primary task. Other approaches \citep{devin2017learning,denil2017programmable} propose architectures that can be reconfigured easily to solve various tasks. Subsequently, \cite{finn2017model} uses meta-learning to acquire skills that can be fine-tuned effectively. 

The use of latent variables and entropy constraints to induce diverse skills has been considered before \citep{daniel2012hierarchical, end2017layered} albeit in a different framework and without using neural network function approximators. Further, \cite{konidaris2007building} used the options framework to learn transferable options using the so-called agent space. Inspired by these ideas, \cite{hausman2018learning} introduces a skill embedding learning method that uses deep reinforcement learning techniques and can concisely represent and reuse skills. Their works draw on a connection between entropy-regularized reinforcement learning and variational inference literature \citep{todorov2008general, toussaint2009robot,neumann2011variational, levine2013variational, rawlik2013stochastic, fox2015taming}. The notion of latent variables in policies has been previously explored by works such as controllers \citep{heess2016learning} and options \citep{bacon2017option}. The auxiliary variable perspective introduces an informative-theoretic regularizer that helps the inference model produce more versatile behaviors. Learning these versatile skills has been previously explored by \cite{haarnoja2017reinforcement}, which learns an energy-based, maximum entropy policy via a Q-learning algorithm, and \cite{schulman2017equivalence}, which uses an entropy regularized reinforcement learning policy. \cite{hausman2018learning} uses a similar entropy-regularized reinforcement learning and latent variables but differs in the algorithmic framework.

\section{Conclusion}

In conclusion, our paper presents a novel adversarial training regime for Multi-Task Reinforcement Learning, which enables us to learn distinct and robust latent for each task. This contribution is significant as it is the first of its kind and requires no manual intervention or domain knowledge of the environments in the multi-task reinforcement learning setting. Our method has been shown to outperform the baseline in several simulated environments such as PointMass, 2D navigation, and robotic manipulation tasks from Meta-World. Moreover, we provide theoretical guarantees to aid optimization in the adversarial regime. We show that the skills learned are mutually exclusive to the task, making them reusable for different and unseen tasks. Our results also demonstrate that the model trained using the adversarial training regime leads to better generalization and more accessible domain adaptation to unseen and complex tasks, even with sparse rewards. This is due to the added exploration factor the agent benefits from when trained in the adversarial regime, which has been.

In summary, our work makes a significant contribution to the field of Multi-Task Reinforcement Learning and opens up new avenues for further research and development in this area.



\section*{Reproducibility Statement}

In this paper, we work with three different datasets, which are all open-sourced. Furthermore, we experiment with two different models - TE-PPO and ATE-PPO. The TE-PPO model was run after reproducing from their open-source code\footnote{\url{https://github.com/rlworkgroup/garage}}. Additional details about setting up these models, and the hyperparameters are available in Appendix~\ref{hyperparam}. Our source code is made available for additional reference \footnote{\url{https://github.com/RamnathKumar181/Adversarial-Learning-Dynamics-in-RL}}. 

\begin{acknowledgements} 
We would like to thank Sony Corporation for funding this research through the Sony Research Award Program.

\end{acknowledgements}

\bibliography{uai2023-template}
\newpage

\onecolumn 
\appendix

\section{Variational Bound Inference.}
\label{vi_lower_bound}

We borrow ideas from variational inference literature to introduce an information-theoretical regularization that encourages versatile skills. In particular, we present a lower entropy of marginal entropy $\mathcal{H}[p(x)]$, which will prove helpful when applied to our objective function from Sec.~\ref{adnet}. Note that this section is not novel and is only provided to help readers better understand our proposed methodology. 

\begin{proposition}
The lower bound on the marginal entropy $\mathcal{H}[p(x)]$ corresponds to:
\begin{equation}
    \mathcal{H}[p(x)] \geq \int \int p(x,z)\log\left ( \frac{q(z|x)}{p(x,z)} dz\right )dx,
\end{equation}
where $q(z|x)$ is the variational posterior
\end{proposition} 

\begin{proof}
\begin{equation}
\label{vi_1}
    \begin{aligned}
    H[p(x)] ={}& \int -p(x)\log[p(x)]dx = \int p(x)\log\left ( \int q(z|x)\frac{1}{p(x)} dz\right )dx\\
         = {}& \int p(x)\log \left ( q(z|x) \frac{p(z|x)}{p(x,z)} dz\right )dx \geq \int p(x) \int p(z|x) \log \left ( \frac{q(z|x)}{p(x,z)} dz\right )dx \\
         = {}& \int \int p(x,z) \log \left ( \frac{q(z|x)}{p(x,z)} dz\right )dx 
\end{aligned}
\end{equation}

From the above Equation~\ref{vi_1}, we can construct a lower bound for our entropy term $\mathcal{H}[\pi(a_i|s_i,t)]$ as follows: 
\begin{equation}
\label{vi_2}
\begin{aligned}
    H[\pi(a|s,t)] \geq{}& \mathbb{E}_{\pi_{\theta}(a,z|s,t)}\left [ \log\left ( \frac{q(z|a,s,t)}{\pi(a,z|s,t)} \right ) \right ]\\
    ={}& \int\int p(\pi_{\theta}(a,z|s,t)) \log\left ( \frac{q(z|a,s,t)}{\pi(a,z|s,t)} \right )\partial a \partial z\\
={}& \int\int p(z|a,s,t)\pi(a|s,t)  \log\left ( \frac{q(z|a,s,t)}{\pi(a,z|s,t)} \right )\partial a \partial z\\
={}& \int\int p(z|a,s,t)\pi(a|s,t)  \left [ \log\left ( q(z|a,s,t) \right) -\log\left ( \pi(a,z|s,t)\right) \right ]\partial a \partial z \\
={}& \int\int \pi(a|s,t)p(z|a,s,t)\log\left ( q(z|a,s,t) \right)\partial a \partial z \\
{}& - \int \int \pi(a|s,t)p(z|a,s,t)\log\left ( \pi(a,z|s,t)\right) \partial a \partial z  \\
={}& -\int \pi(a|s,t)\mathcal{CE}[p(z|a,s,t)||q(z|a,s,t)]\partial a  \\
{}& -\int \int \pi(a|s,t)p(z|a,s,t)\log\left ( \pi(a,z|s,t)\right) \partial a \partial z  \\
={}& -\mathbb{E}_{\pi(a|s,t)}\left [  \mathcal{CE}[p(z|a,s,t)||q(z|a,s,t)]\right ] \\
{}& -\int \int \pi(a|s,t)p(z|a,s,t)\log\left ( \pi(a,z|s,t)\right) \partial a \partial z  \\
\end{aligned}
\end{equation}

Where $\mathcal{CE}$ is cross entropy. To simplify the second part of the result from Equation~\ref{vi_2}, we simplify it as follows:

\begin{equation}
\label{vi_3}
\begin{aligned}
={}& -\int \int \pi(a|s,t)p(z|a,s,t)\log\left ( \pi(a,z|s,t)\right) \partial a \partial z  \\
={}& -\int \int \left [  \pi(a|s,t)p(z|a,s,t)\log\left (  p(z|s,t)\right) + \pi(a|s,t)p(z|a,s,t)\log\left (  \pi(a|s,t,z)\right) \right ] \partial a \partial z  \\
={}& -\int \int  p(z,a|s,t)\log\left (  p(z|s,t)\right)\partial a \partial z - \int \int \pi(a|s,t)p(z|a,s,t)\log\left (  \pi(a|s,t,z)\right)  \partial a \partial z  \\
={}& -\int \int  p(z,a|s,t)\log\left (  p(z|s,t)\right)\partial a \partial z - \int \int p(z,a|s,t)\log\left (  \pi(a|s,t,z)\right)  \partial a \partial z  \\
={}& -\int  p(z|s,t)\log\left (  p(z|s,t)\right) \partial z - \int \int p(z|s,t)\pi(a|s,t,z)\log\left (  \pi(a|s,t,z)\right)  \partial a \partial z  \\
\end{aligned}
\end{equation}

Since, the skill embedding is conditionally independent of the state of the agent given task $t$, we can simplify $p(z|s,t)$ as $p(z|t)$. Similarly, since the action $a$ is conditionally independent of the task $t$ given the latent $z$, we can simplify $\pi(a|s,t,z)$ as $\pi(a|s,z)$. This is possible since $z$ carries all the necessary information from $t$, which is required to solve the task. With the above simplifications, Equation~\ref{vi_3} further simplifies to:

\begin{equation}
\label{vi_4}
\begin{aligned}
={}& -\int  p(z|s,t)\log\left (  p(z|s,t)\right) \partial z - \int \int p(z|s,t)\pi(a|s,t,z)\log\left (  \pi(a|s,t,z)\right)  \partial a \partial z  \\
={}& -\int  p(z|t)\log\left (  p(z|t)\right) \partial z - \int \int p(z|t)\pi(a|s,z)\log\left (  \pi(a|s,z)\right)  \partial a \partial z  \\
={}& \mathcal{H}[p(z|t)] + \int p(z|t)\mathcal{H}\left [ \pi(a|s,z) \right ]\partial z  \\
={}& \mathcal{H}[p(z|t)] + \mathbb{E}_{p(z|t)}\left [\mathcal{H}\left [ \pi(a|s,z) \right ]  \right ] \\
\end{aligned}
\end{equation}

Note that $q(z|a,s,t)$ is the variational inference distribution we are free to choose. Since $q(z|a,s,t)$ is intractable, we resort to a sample-based evaluation of the Cross-Entropy term. This bound holds for any $q$. Similar to \cite{hausman2018learning}, we avoid conditioning $q$ on task $t$ to ensure that a given trajectory alone will allow us to identify its skill embedding. Substituting the results from Equation~\ref{vi_3}, Equation~\ref{vi_4}, we can rewrite Equation~\ref{vi_2} as:

\begin{equation}
\label{vi_final}
\begin{aligned}
H[\pi(a|s,t)] ={}& -\mathbb{E}_{\pi(a|s,t)}\left [  \mathcal{CE}[p(z|a,s,t)||q(z|a,s,t)]\right ] + \mathcal{H}[p(z|t)] + \mathbb{E}_{p(z|t)}\left [\mathcal{H}\left [ \pi(a|s,z) \right ]  \right ] \\
={}& -\int \pi(a|s,t) p(z|a,s,t) \log \left [q(z|a,s)  \right ]\partial a + \mathcal{H}[p(z|t)] + \mathbb{E}_{p(z|t)}\left [\mathcal{H}\left [ \pi(a|s,z) \right ]  \right ] \\
={}& \int p(a,z|s,t) \log \left [q(z|a,s)  \right ]\partial a + \mathcal{H}[p(z|t)] + \mathbb{E}_{p(z|t)}\left [\mathcal{H}\left [ \pi(a|s,z) \right ]  \right ] \\
={}& \mathbb{E}_{p(a,z|s,t)} \left [\log \left [q(z|a,s)  \right ]  \right ] + \mathcal{H}[p(z|t)] + \mathbb{E}_{p(z|t)}\left [\mathcal{H}\left [ \pi(a|s,z) \right ]  \right ] \\
\end{aligned}
\end{equation}
\end{proof}

\section{Theoretical Results}
\label{the_conv}

We will first prove that the minimax game has a global optimum for a given policy network and a given embedding network in the case of an on-policy setting, where the optimal policy is already known to us. This proof is presented in Appendix~\ref{on-policy-proof}.

These proofs guarantee the working of our algorithm and set the necessary conditions for the model to train appropriately.

\subsection{On-Policy Optimality}
\label{on-policy-proof}

\begin{proposition}
For a given $\mathbb{E}$, the optimal policy $\mathbb{\pi}$ is
\begin{equation*}
\label{on_1}
\mathbb{\pi}_{\mathbb{E}}^{*}(\tau) = arg\,max_{\pi} Q_{\pi}^{\varphi}(s,a,z,\tau)
\end{equation*}
and the bound on $Q_{\pi}^{\varphi}(s,a,z,\tau)$ is going to be such that:
\begin{equation*}
\label{on_2}
Q_{\pi}^{\varphi}(s,a,z,\tau) \leq \frac{R_{\max}}{1-\gamma} + \alpha_3\frac{\log\left |a_{\max}  \right |}{1-\gamma}\
\end{equation*}
\end{proposition}

\begin{proof}
The training criterion for the given policy network $\mathbb{\pi}$, given any encoder $\mathcal{E}$, is to maximize the quantity $V(\mathcal{E},\mathbb{\pi})$ ($-\mathcal{L}_{\text{pro}}$) from Equation~\ref{eq:loss_protagonist}. Our equation now reduces to:

\begin{equation}
\label{on-proof_1}
    \begin{aligned}
    V(\mathbb{E},\mathbb{\pi}) ={}& \max_{\pi} Q_{\pi}^{\varphi}(s,a,z,\tau) \\
    ={}& \max_{\pi} \sum_{i=0}^{\infty} \gamma^i \left [ r_\tau(s_i,a_i) + \alpha_2 \log[q(z|a_i,s_i^H)] + \alpha_3\mathcal{H}[\pi(a|s,z)] \right ]\\
\end{aligned}
\end{equation}
Here, we will assume the rewards to be bounded by the range $[0, R_{\max}]$. Note that any reward function, sparse or otherwise, can be transformed to the above range and subject to the same proof. Furthermore, we assume perfect optimality of the inference network $q$, and the resulting Cross Entropy loss can be omitted since it would be 0. With the use of theory from Kullback–Leibler divergence and Shannon Entropy, we know that $\mathcal{H}(x) \leq log \left | x \right |$. In our case, the entropy $\mathcal{H}[\pi(a|s,z)]$ can be bounded to $\log\left |a_t  \right |$, where $a_\tau$ is the action space of the given task. For simplification, we approximate $\left |a_t  \right |$ to $\left |a_{\max} \right |$, where $a_{\max}$ denotes the action space with maximum cardinality. We can further simplify the equation above as follows:

\begin{equation}
\label{on-proof_2}
\begin{aligned}
     Q_{\pi}^{\varphi}(s,a,z,\tau) ={}& \sum_{i=0}^{\infty} \gamma^i \left [ r_\tau(s_i,a_i) + \alpha_2 \log[q(z|a_i,s_i^H)] + \alpha_3\mathcal{H}[\pi(a|s,z)] \right ]\\
\leq{}& \sum_{i=0}^{\infty} \gamma^i \left [ R_{\max} + \alpha_3\log\left |a_{\max}  \right | \right ]\\
\leq{}& \frac{R_{\max}}{1-\gamma} + \alpha_3\frac{\log\left |a_{\max}  \right |}{1-\gamma}\\
\end{aligned}
\end{equation}

The Equation~\ref{eq:q-final} can now be reformulated as:
\begin{equation}
\label{on-proof_3}
\begin{aligned}
     C(\mathbb{E}) ={}& \min_{\mathcal{E}}\mathcal{L}_{\text{adv}} \\
={}&\frac{R_{\max}}{1-\gamma} + \alpha_3\frac{\log\left |a_{\max}  \right |}{1-\gamma} - \alpha \mathcal{JSD}(z,\tau) \\
\end{aligned}
\end{equation}
\end{proof}

\begin{proposition}
The global minimum of the training criterion $C(\mathcal{E})$ is achieved if and only if
\begin{equation}
\alpha > \frac{R_{\max}}{1-\gamma} + \alpha_3\frac{\log\left |a_{\max}  \right |}{1-\gamma}
\end{equation}
At this point, $C(\mathbb{E})$ achieves a minimum value bounded by $C(\mathbb{E})<0$ and is not a trivial solution where the embedding function is an identity function. Furthermore, the property that $z$ and $t$ are mutually exclusive is held.
\end{proposition}

\begin{proof}
Since the Jensen–Shannon divergence between two distributions is always non-negative and zero if they are equal. Since we want to minimize Equation~\ref{on-proof_3}, we escape the trivial solution where the embedding network is an identity function. Note that, since the input to the embedding network is a one-hot encoded embedding of the task id, it satisfies the condition of the sharp peak and is uniform in terms of tasks and skills. For the time being, let us assume $\alpha$ is high enough and the $\mathcal{JSD}$ objective is being tuned rather than the term that helps deceive the agent network. In this scenario, we have two unique cases: (i) when $z \sim \tau$, and when (ii) $z \not\sim t$. Note that the Jensen-Shannon divergence would be equal to 0 in the first case and positive in the second. Since our goal is to minimize the objective, the model would converge towards the goal where $z \not \sim \tau$, and escape the trivial solution of the identity function. However, to ensure that the $\mathcal{JSD}$ objective is being optimized with higher precedence, we set $\alpha$ such that:

\begin{equation}
\label{on-proof_4}
\alpha \mathcal{JSD}(z,\tau) > \frac{R_{\max}}{1-\gamma} + \alpha_3\frac{\log\left |a_{\max}  \right |}{1-\gamma} \\
\end{equation}

Since we have already shown that the optimal solution is when $z \not \sim \tau$, or when z and $\tau$ are mutually exclusive, and JSD would be at its maximum value of 1. This mutual exclusivity property is essential since we would like the same skill to be shared across tasks, as long as the tasks are based on similar structures. For instance, turning a doorknob or screwing a cap on a bottle involves identical skills. Using these fundamental skills to improve task structure in meta-learning would facilitate more accessible posterior adaptation and causal inference. With the assurance that our intuition is following the mathematical rigor, we can select $\alpha$ such that:

\begin{equation}
\label{on-proof_5}
\alpha > \frac{R_{\max}}{1-\gamma} + \alpha_3\frac{\log\left |a_{\max}  \right |}{1-\gamma} \\
\end{equation}

Provided this condition is met, the training criterion $C(E)$ achieves a minimum value bounded such that $C(\mathbb{E}) < 0$. We select hyperparameters $(\alpha_3, \alpha)$ such that the following condition is satisfied.
\end{proof}


\subsection{Tractable optimization of objective}
\label{tract_entropy}

Here, we show that the equation presented in Equation~\ref{eq:q-final} might not be the ideal form to compute the loss. This is because the computation of the Jensen-Shanon Divergence forces the shape of both skills $z$ and task $\tau$ to be the same. Furthermore, since $z$ is derived from $\tau$, it is not entirely straightforward how to compute $\mathcal{H}(\tau|z)$. To subvert these issues, we simplify the above equation as follows.

Recall, 
\begin{equation}
\label{tract_1}
\mathcal{H}(\tau)-\mathcal{H}(\tau|z) = \mathcal{H}(z) - \mathcal{H}(z|\tau)
\end{equation}

Note that our original objective equation does have the L.H.S. term. Thus, we simplify the objective to:

We simplify $H(\tau|z)$ as $H(\tau)-H(z)+H(z|\tau)$. Note that the $H(\tau)$ term cancels out and we are left with $2\alpha(H(z)-H(z|\tau))$ as our second term in Equation~\ref{eq:q-final}. This simplification is what we use while computing, but we use the Jensen-Shannon Divergence simplification from Equation~\ref{eq:q-final} to complete our proof of convergence of the model.
\begin{equation}
\label{eq:tract_2}
\begin{aligned}
\mathcal{L}_{\text{adv}} ={}&  \left [Q_{\pi}^{\varphi}(s,a;z,\tau)  \right ] - \alpha \mathbb{E}_{\tau\in \mathcal{T}}\left [\mathcal{H}(z) - \mathcal{H}(z|\tau) + \mathcal{H}(\tau) - \mathcal{H}(\tau|z)\right ]\\
={}& \left [Q_{\pi}^{\varphi}(s,a;z,\tau)  \right ] - \alpha \mathbb{E}_{\tau\in \mathcal{T}}\left [\mathcal{H}(z) - \mathcal{H}(z|\tau) + \mathcal{H}(z) - \mathcal{H}(z|\tau)\right ]\\
={}&  \left [Q_{\pi}^{\varphi}(s,a;z,\tau)  \right ]  - 2*\alpha \mathbb{E}_{\tau\in \mathcal{T}}\left [ \mathcal{H}(z) - \mathcal{H}(z|\tau)\right ]\\
\end{aligned}
\end{equation}

Since $\alpha$ is a hyperparameter of our choice, we can assign $2\alpha$ as $\alpha '$ and simplify the equation as follows:

\begin{equation}
\label{eq:tract_3}
\begin{aligned}
\mathcal{L}_{\text{adv}} ={}& \left [Q_{\pi}^{\varphi}(s,a;z,\tau)  \right ]  - \alpha' \mathbb{E}_{\tau\in \mathcal{T}}\left [ \mathcal{H}(z) - \mathcal{H}(z|\tau)\right ]\\
\end{aligned}
\end{equation}

The above equation is what we use to make the computation tractable.

\subsection{Extending to Off-Policy Setting}
\label{extending_off_policy}
In this section, estimate the discounted sums from Equation~\ref{eq:q-final} from previously gathered data by learning a Q-value function, yielding an off-policy algorithm. Similar to \cite{hausman2018learning}, we assume the availability of a replay buffer $\mathcal{B}$ (containing full trajectory execution traces including states, actions, task id, and reward) that is inherently filled during training. In conjunction with these trajectory traces, we also store the probabilities of each selected action and denote them with the behavior policy probability $b(a|z,s,\tau)$ and the behavior probabilities of the embedding $b(z|\tau)$. Given this replay data, we formulate the off-policy perspective of our algorithm. We start with the notion of a \textit{lower-bound Q-function} that depends on both states $s$ and $a$ and is conditioned on both the embedding $z$ and the task id $\tau$. To learn a parametric representation of $Q_{\pi}^{\varphi}$, we make use of the Retrace algorithm \citep{munos2016safe}, which quickly allows us to propagate entropy augmented rewards across multiple time steps while minimizing the bias of the algorithm relying on the parametric Q-function. Formally, we fit $Q_{\pi}^{\varphi}$ by minimizing the squared loss:
\begin{equation}
\label{off-1}
\begin{aligned}
{}&\min_{\varphi}\mathbb{E}_{\mathcal{B}} \left [ (Q_{\varphi}^{\pi}(s_i,a_i;z,\tau)-Q^{\text{ret}})^2 \right ] \text{, where}\\
{}&Q^{\text{ret}} = \sum_{j=i}^{\infty} \left ( \gamma^{j-i}\prod _{k=i}^j c_k \right ) r(s_j,a_j,z,\tau) {}&\\
\end{aligned}
\end{equation}
wherein,
\begin{equation*}
\label{off-1_sup}
\begin{aligned}
{}& r(s_j,a_j,z,\tau) = \widehat{r}(s_j,a_j,z,\tau) \\
{}&+ \mathbb{E}_{\pi(a|z,s,\tau)} \left [ Q_{\varphi ^{'}}^{\pi} (s_i,.;z,\tau) - Q_{\varphi ^{'}}^{\pi} (s_j,a_j;z,\tau)\right ]\\
& c_k = \min \left ( 1,\frac{\pi(a_k|z,s_k,\tau)p(z|\tau)}{b(a_k|z,s_k,\tau)b(z|\tau)} \right )
\end{aligned}
\end{equation*}
We compute the terms contained in $\widehat{r}$ by using $r_\tau$ and $z$ from the replay buffer and re-compute the (cross-)entropy terms. Here, $\varphi ^{'}$ denotes the parameters of a target Q-network \citep{mnih2015human} that we occasionally copy from the current estimate $\varphi$ and $c_k$ are the per-step importance weights. Further, we bootstrap the infinite sum after $N$-steps with $\mathbb{E}_{\pi} \left [ Q_{\varphi ^{'}}^{\pi}(s_N,.;z_N,\tau)\right]$ instead of introducing a $\lambda$ parameter as in the original paper. Equipped with this Q-function, we can update the policy and embedding network parameters without requiring additional environment interactions by optimizing the same objective as Eq. ~{\ref{eq:q-final}}.

We highlight that the above derivation and steps also hold when the task id is constant, i.e., for the regular reinforcement learning setting rather than the meta-reinforcement learning setting. The following sections in Appendix~\ref{the_conv} present a more detailed theoretical analysis of our adversarial network training regime, essentially showing that the training criterion converges appropriately given the necessary conditions are guaranteed. In practice, we cannot sequentially train the two networks. We must implement the game using an iterative back and forward approach. Optimizing $\mathcal{E}$ to completion would lead to overfitting, and the Encoder would not learn the adequate skill embedding to be used by the policy network. Instead, we alternate between $k$ steps of optimizing the protagonist and one stage of optimizing the adversary. This results in the protagonist being maintained near its optimal solution, so long as adversarial skills change slowly enough. It might seem counter-intuitive to minimize the discounted rewards by the embedding function. However, as long as the hyperparameter $\alpha$ is set correctly, the Jensen-Shannon Divergence objective prioritizes the discounted rewards criterion. With this assumption, we can ensure that minimizing the expected returns would merely act as an implicit noise \citep{hjelm2018learning}.

\section{Desiredata of Loss function}
Intuitively, the resulting objective function bound meets the following desiderata:
\begin{itemize}
    \item \textbf{Discounted Returns $\boldsymbol{r_t(s_i,a_i)}$:} The discounted returns objective is a widely used method in reinforcement learning that enables an agent to learn an optimal policy for a given task and environment. The objective is designed to maximize the reward signal that the agent receives over time, encouraging the agent to make decisions that lead to the highest possible reward. The goal of the discounted returns objective is to train an agent to perform a task in a way that results in a high reward signal, thereby increasing the overall success of the agent. 
    \item \textbf{Cross Entropy $\boldsymbol{\log[\mathbb{I}(z|a_i,s_i^H)]}$:} The term "encourages different embedding vectors" refers to the idea that different values of $z$ should lead to distinct results in terms of the actions taken by the agent and the states visited. Intuitively, the term will have a high value when the inference network ($\mathbb{I}$) can accurately predict the value of $z$ based on the resulting actions ($a$) and states ($s^H$) visited by the agent. Here, $H$ refers to the previously defined notion. This term incentivizes the agent to explore a diverse range of actions and states, leading to a more comprehensive understanding of the task and environment.
    \item \textbf{Entropy of the policy conditioned on the embedding. $\boldsymbol{\mathcal{H}[\pi(a|s,z)]}$:} The Entropy of the policy conditioned on the embedding term is a measure designed to promote the exploration of diverse skills within the embedding space. The goal is to ensure that the policy learned by the agent is not limited to a narrow range of skills, but instead covers a broad spectrum of potential solutions. By maximizing the entropy of the policy conditioned on the embedding, the agent is encouraged to learn and explore a diverse set of skills, which can lead to more robust and adaptable solutions in complex environments.
    \item \textbf{Entropy of the embedding given task $\boldsymbol{\mathcal{H}(z|t)}$:} The minimization of the entropy of the embedding given a task is a key aspect of our proposed method. This objective functions to make the skill embedding highly specific and deterministic for a particular task. By reducing the entropy of the embedding, the agent is forced to focus on a single, well-defined skill that is most likely to produce high rewards for the given task. This results in a sharp peak in the distribution of the embedding, ensuring that the learned policy is highly specific and deterministic for the given task. The minimization of the entropy of the embedding is critical in achieving a consistent and robust learning process, leading to improved performance in challenging environments.
    \item \textbf{Entropy of the embedding $\boldsymbol{\mathcal{H}(z)}$:} To ensure that the algorithm covers a diverse range of skills in the latent space, we maximize the entropy of the skill embedding. Since the tasks are sampled uniformly, we maximize the entropy and are already at the optimal solution.
    \item \textbf{Entropy of task given embedding $\boldsymbol{\mathcal{H}(t|z)}$:} We aim to make the prediction of task $t$ given the skill embedding $z$ deterministic and focused, with a sharp peak. To simplify the computation, we present a modified version of Equation~\ref{eq:q-final} in the appendix~\ref{tract_entropy}. This simplification will eliminate the unnecessary calculation of entropy of task $\mathcal{H}(t)$.

\end{itemize}

\label{limitations}

As discussed briefly in this section, working in an adversarial training regime does pose its own set of limitations. If the hyperparameters are not carefully set, it might become unstable and inconsistent when the adversary completely nullifies the learning from the protagonist. This could occur if the conditions highlighted in Appendix~\ref{the_conv} are not met. This could also happen if the number of iterations the adversary is trained is much greater than the number of iterations of the protagonist forcing the final weights to be closer to an agent trying to minimize the rewards instead of maximizing. Furthermore, setting $\alpha$ in Equation~\ref{eq:loss_adversary} very low will give more importance to learning skills that fool the agent but are not necessarily diverse and distinct. 
Although our adversarial framework does outperform the traditional training regime in multiple environments, it isn't easy to support the need for such a framework due to its inherent counter-intuitiveness. It requires more theoretical studies to understand this behavior.

\section{Computing Volume of latent embedding}
\label{vol}

\begin{proposition}
Let $\Pi$ be an $m$-dimensional parallelotope defined by edge vectors $\mathcal{B}= \left \{ v_1,v_2,...,v_m \right \}$, where $v_i \in \mathbb{R}^n$ for $n\geq m$. That is, we are looking at an $m$-dimensional parallelotope embedded inside $n$-dimensional space. Suppose $\mathcal{A}$ is the $m\times n$ matrix with row vectors $\mathcal{B}$ given by:
\begin{equation*}
    \mathcal{A} = \begin{pmatrix}
v_1^T\\ 
\vdots\\ 
v_m^T
\end{pmatrix}
\end{equation*}
Then the $m$-dimensional volume of the paralleletope is given by:
\begin{equation*}
\left [\text{vol}(\pi)  \right ]^2 = \det(AA^T)
\end{equation*}
\end{proposition}

\begin{proof}
Note that $AA^T$ is an $m\times m$ square matrix. Suppose that $m=1$, then:
\begin{equation*}
\det(AA^T)= \det(v_1v_1^T) = v_1\cdot  v_1 = \left \| v_1 \right \|^2 = \left [\text{vol}_1(v_1)  \right ]^2
\end{equation*}

so the proposition holds for $m=1$. From this base equation, we prove the above theorem by induction. Now, we induct on $m$.

Let us assume the proposition holds for $m^{'}$ such $m^{'}\geq 1$. If we can also prove that the proposition holds for $m^{'}+1$, we would have proved the above theorem. Letting $A_{m^{'}}$ denote the matrix containing rows $v_1$ to $v_{m^{'}}$, we can write $A = A_{m^{'}+1}$ as:
\begin{equation*}
A = \begin{pmatrix}
A_{m^{'}}\\ 
v_{m^{'}+1}^T
\end{pmatrix}
\end{equation*}

We may decompose $v_{m^{'}+1}$ orthogonally as:
\begin{equation*}
v_{m^{'}+1} = v_{\perp } + v_{\parallel }
\end{equation*}
where $v_{\perp}$ lies in the orthogonally complement of the base (i.e., the height of our parallelepipe), and $v_{\perp}\cdot v_i=0$, $\forall$ $1\leq i \leq m^{'}$. Furthermore, $v_{\parallel}$ must be in the span of vectors $\left \{ v_1,v_2,...,v_{m^{'}} \right \}$, such that:
\begin{equation*}
v_{\parallel} = c_1v_1 + ...  c_{m^{'}}v_{m^{'}}
\end{equation*}
We apply a sequence of elementary row operations to $A$, adding a multiple $-c_i$ of row $i$ to row $m^{'}+1$, $\forall$ $1\leq i \leq m^{'}$. We can then write the resulting matrix $B$ as:

\begin{equation*}
    B = \begin{pmatrix}
A_{m^{'}}\\ 
 \\
v_{\perp}^T
\end{pmatrix} = E_{m^{'}}...E_1 A,
\end{equation*}

Each $E_i$ is an elementary matrix adding a multiple of one row to another. Notice that the above operation corresponds to shearing the parallelotope so that the last edge is perpendicular to the base. We see that these operations do not change the determinant as:

\begin{equation*}
\det(BB^T) = \det(E_{m^{'}}...E_1 (AA^T) E_{1}^T...E_{m^{'}}^T) = \det(AA^T)
\end{equation*}

Through block multiplication, we can obtain $BB^T$ as follows:
\begin{equation*}
\begin{aligned}
BB^T ={}& \bigl(\begin{smallmatrix}
 A_{m^{'}}\\ 
 \\
v_{\perp}^T
\end{smallmatrix}\bigr) \bigl(\begin{smallmatrix}
 A_{m^{'}}^T & &
v_{\perp}
\end{smallmatrix}\bigr) \\
={}& \bigl(\begin{smallmatrix}
 A_{m^{'}}A_{m^{'}}^T & & A_{m^{'}}v_{\perp}\\ 
 \\
v_{\perp}^TA_{m^{'}}^T & & v_{\perp}^Tv_{\perp}
\end{smallmatrix}\bigr)\\
={}& \bigl(\begin{smallmatrix}
 A_{m^{'}}A_{m^{'}}^T & & A_{m^{'}}v_{\perp}\\ 
 \\
(A_{m^{'}}v_{\perp})^T & & \left \| v_{\perp} \right \|^2\end{smallmatrix}\bigr)\\
\end{aligned}
\end{equation*}

Furthermore, notice that
\begin{equation*}
\begin{aligned}
A_{m^{'}}v_{\perp} ={}&\begin{pmatrix}
v_1^T\\ 
\vdots\\ 
v_{m^{'}}^T
\end{pmatrix}v_{\perp} =0\\
\end{aligned}
\end{equation*}

Therefore, we have 
\begin{equation*}
\begin{aligned}
BB^T ={}& \bigl(\begin{smallmatrix}
 A_{m^{'}}A_{m^{'}}^T & & 0\\ 
 \\
0^T & & \left \| v_{\perp} \right \|^2\end{smallmatrix}\bigr)\\
\end{aligned}
\end{equation*}

Taking the determinant, we can simplify $\det(BB^T)$ as 
\begin{equation*}
\begin{aligned}
\det(BB^T) ={}&  \left \| v_{\perp} \right \|^2 \det( A_{m^{'}}A_{m^{'}}^T)\\
\end{aligned}
\end{equation*}

By definition, $\left \| v_{\perp} \right \|$ is the height of the parallelotope, and by the induction hypothesis, $\det(A_{m^{'}},A_{m^{'}}^T)$ is the square of the base. Therefore, we have proved the above theorem by induction.
\end{proof}

\section{Analysis of Skills learned}
\label{ace_results}

In this section, we discuss a few additional results, along with a study of the effect of each latent feature on the agent's behavior.

\subsection{PointMass Environment}
\label{appendix:point_mass}

\begin{figure}[h]
\centering
\subfigure[Goal: Left]{%
\includegraphics[width=0.22\linewidth]{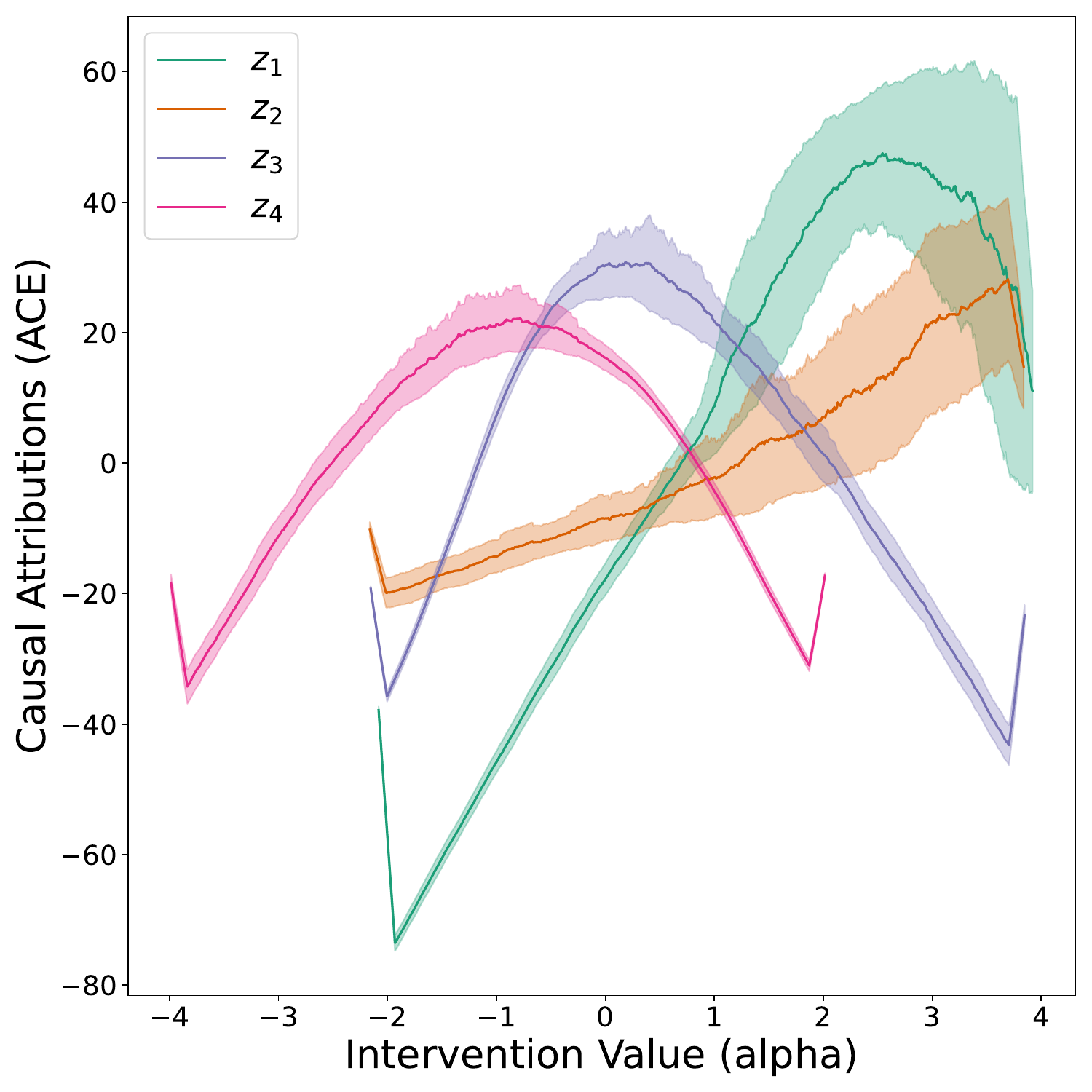}
\label{fig:ate_left}}
\quad
\subfigure[Goal: Right]{%
\includegraphics[width=0.22\linewidth]{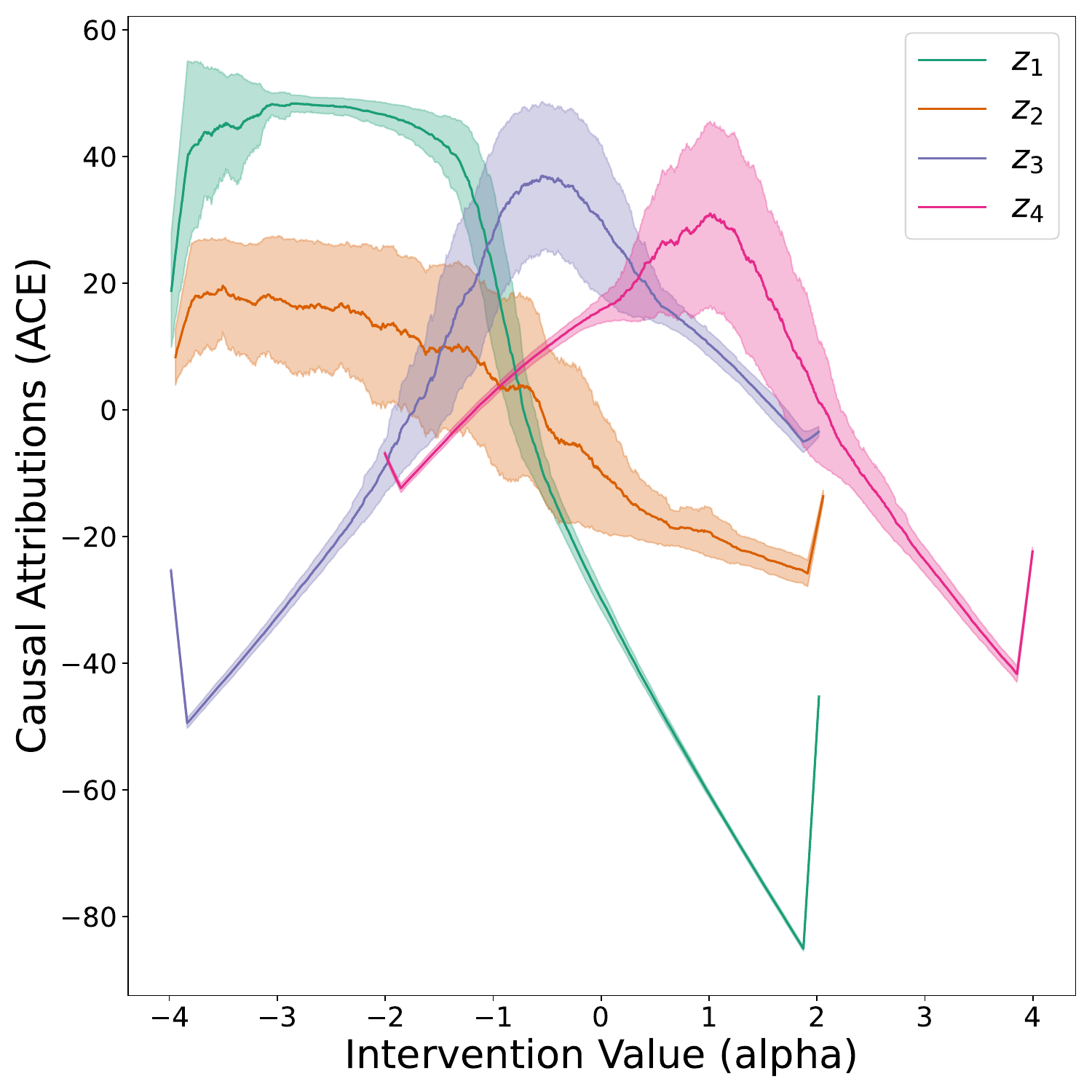}
\label{fig:ate_right}}
\quad
\subfigure[Goal: Top]{%
\includegraphics[width=0.22\linewidth]{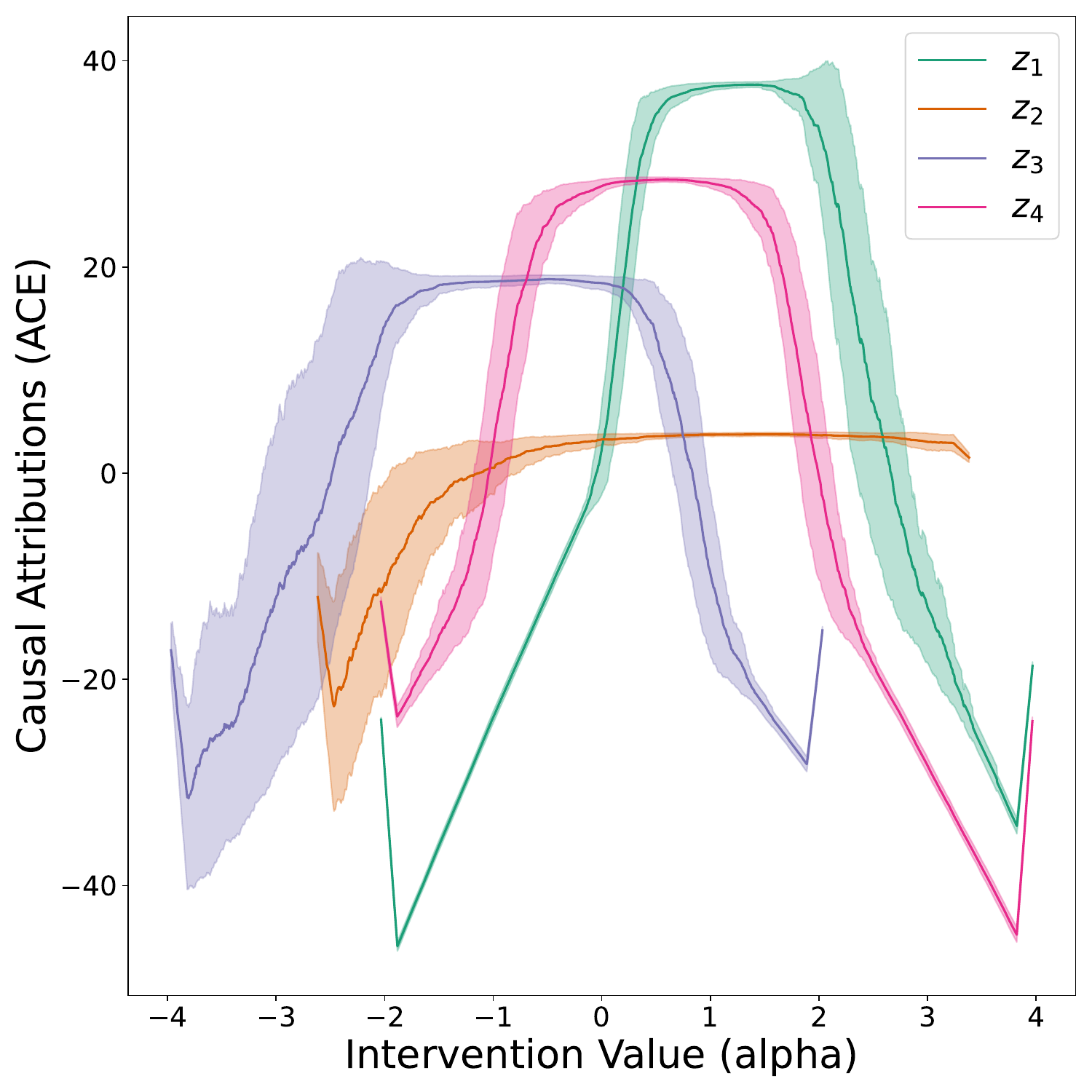}
\label{fig:ate_top}}
\quad
\subfigure[Goal: Bottom]{%
\includegraphics[width=0.22\linewidth]{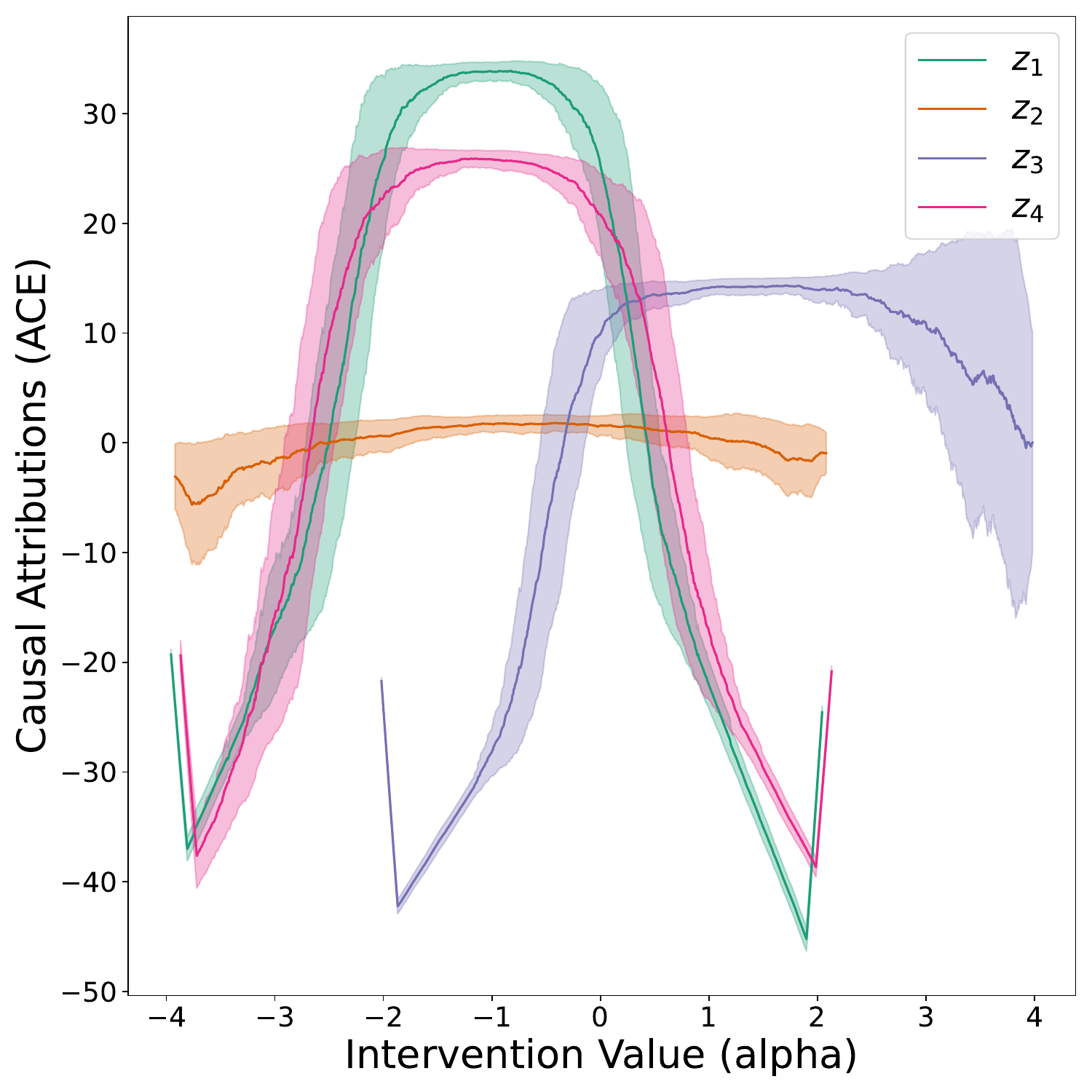}
\label{fig:ate_bottom}}

\caption{The Average Causal Effect of the latent skills for different tasks in the PointMass Environment. From the above plots, we can make a few inferences, such as the fact that embedding $z_2$ seems to be an auxiliary feature and does not play a significant role in the behavior of the agent.}
\label{fig:ace_pointmass}
\end{figure}

From the ACE plots, we can successfully assert that feature $z_2$ is an auxiliary variable and does not play any significant role in the model's behavior since the ACE value is close to 0. Furthermore, we notice that the $z_3$ and $z_4$ causal analysis are pretty similar, and $z_1$ seems to be an essential feature in the pool of latent features. 

\begin{figure}[hbt!]
    \centering
    \subfigure[The effect of $z_1$ on the behavior of the environment. As we can notice from the above plots, $z_1$ seems to have a significant effect on the behavior of the agent. We hypothesize that the $z_1$ feature is used to determine the top-left and bottom-right segment of tasks. This would explain why some trajectories, specifically Top/Left and Bottom/Right are similar to each other.]{
    \begin{tabular}{cccc}
        \includegraphics[width=0.2\linewidth]{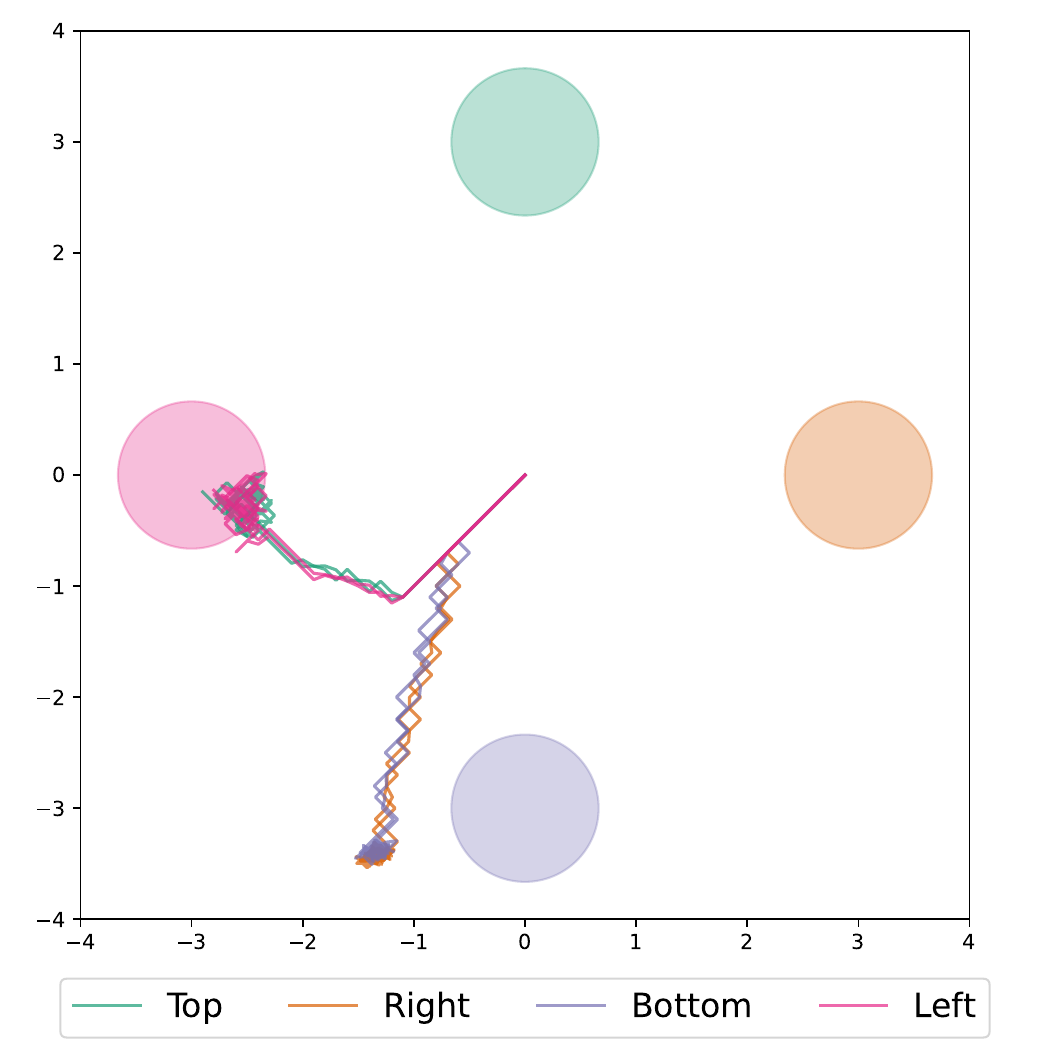} & 
        \includegraphics[width=0.2\linewidth]{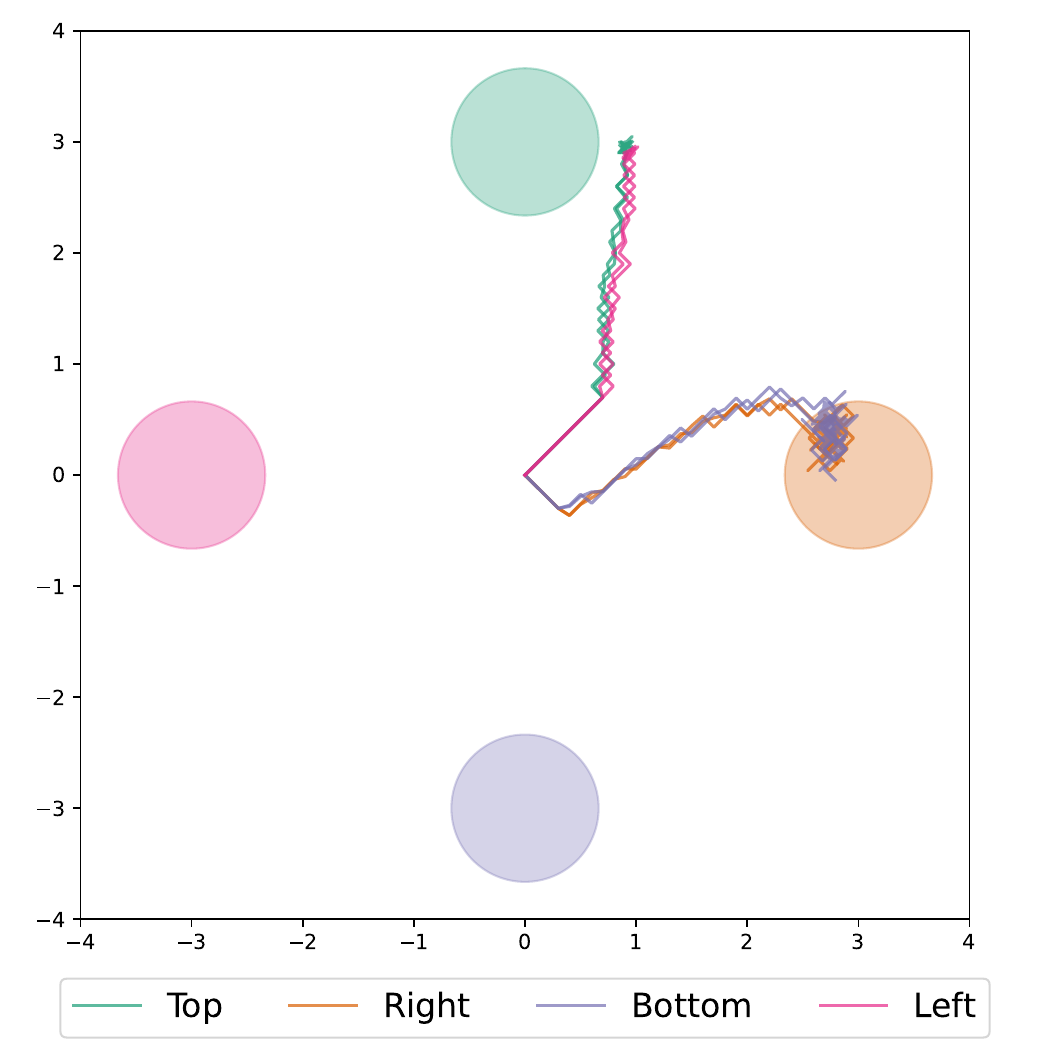} &
        \includegraphics[width=0.2\linewidth]{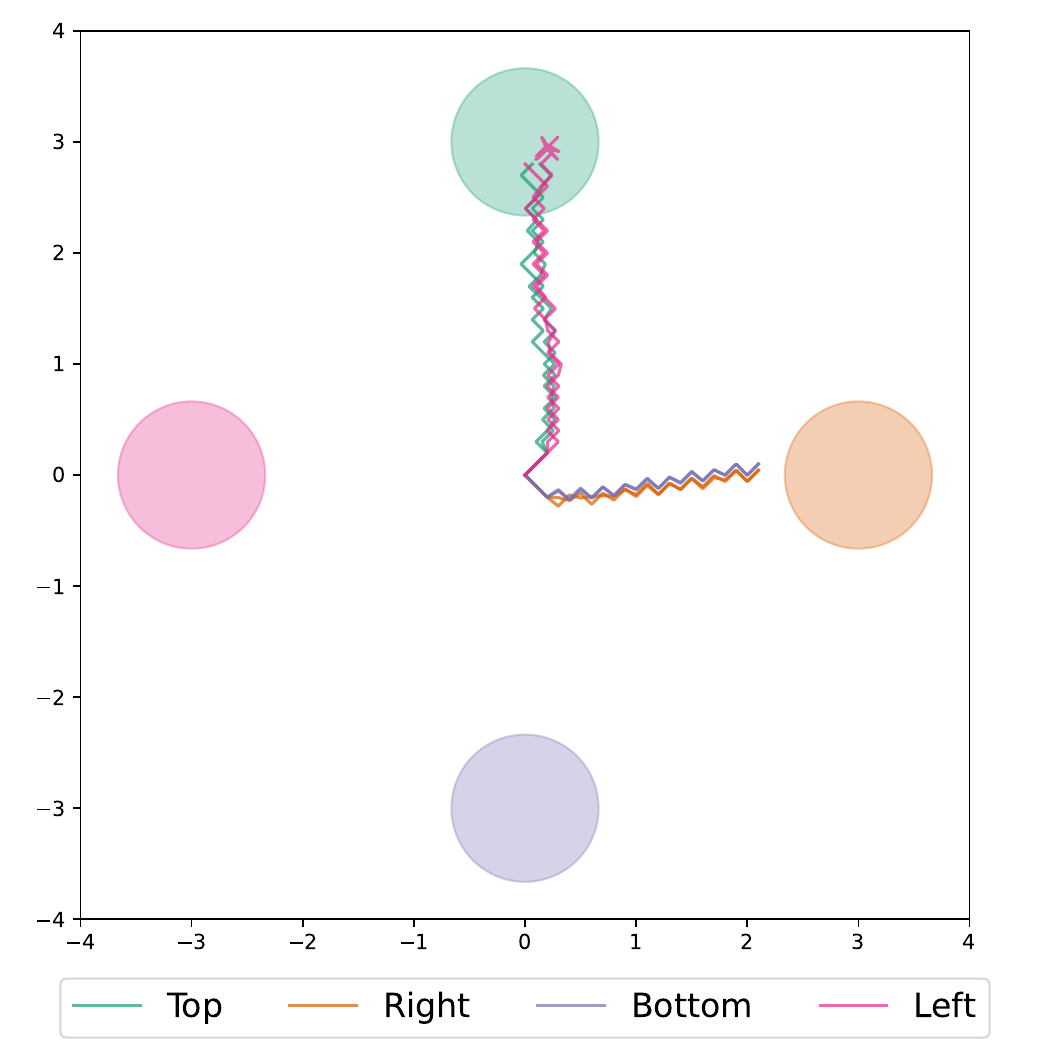} &
        \includegraphics[width=0.2\linewidth]{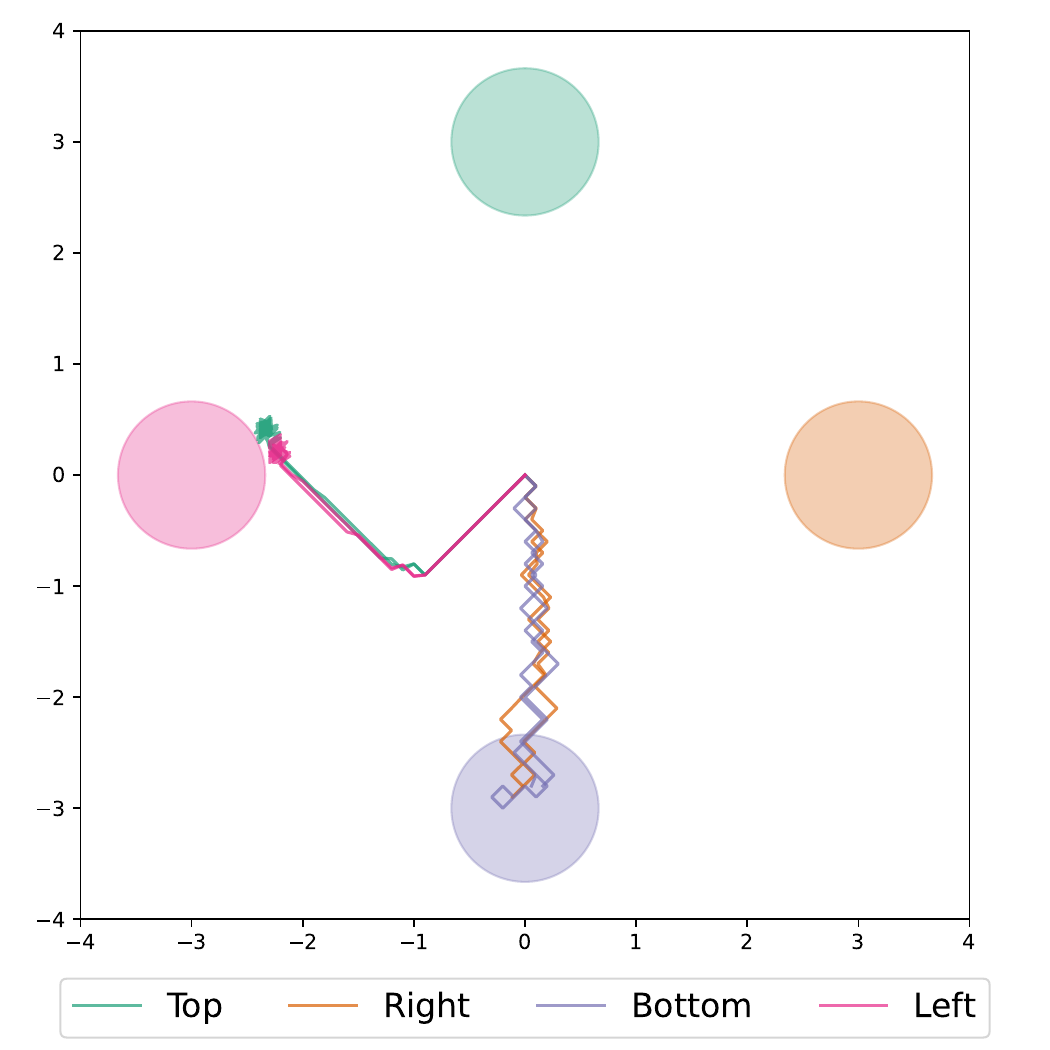} \\
        Goal: Left & Goal: Right & Goal: Top & Goal: Bottom\\
        & & & 
    \end{tabular}
    \label{fig:z1_importance_pointmass}}
    
    \subfigure[The effect of $z_2$ on the behavior of the environment. As we can notice from the above plots, $z_2$ seems to have little to no effect on the behavior of the agent. We hypothesize that the $z_2$ feature could have low importance and unnecessary, or could be a basic skill used across all tasks. This would explain why perturbation of the $z_2$ feature does not lead to any difference in behavior, i.e. not a discriminating feature across tasks.]{
    \begin{tabular}{cccc}
        \includegraphics[width=0.2\linewidth]{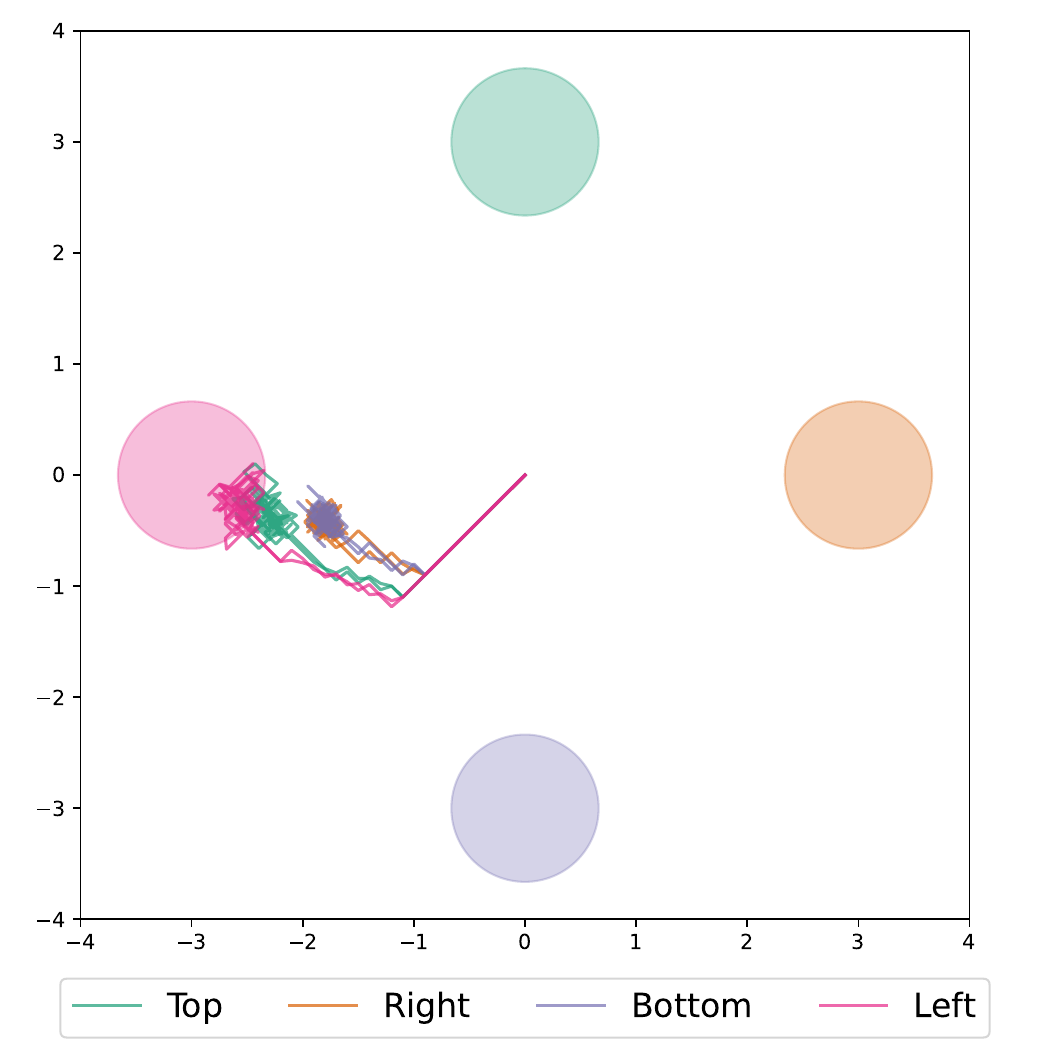} & 
        \includegraphics[width=0.2\linewidth]{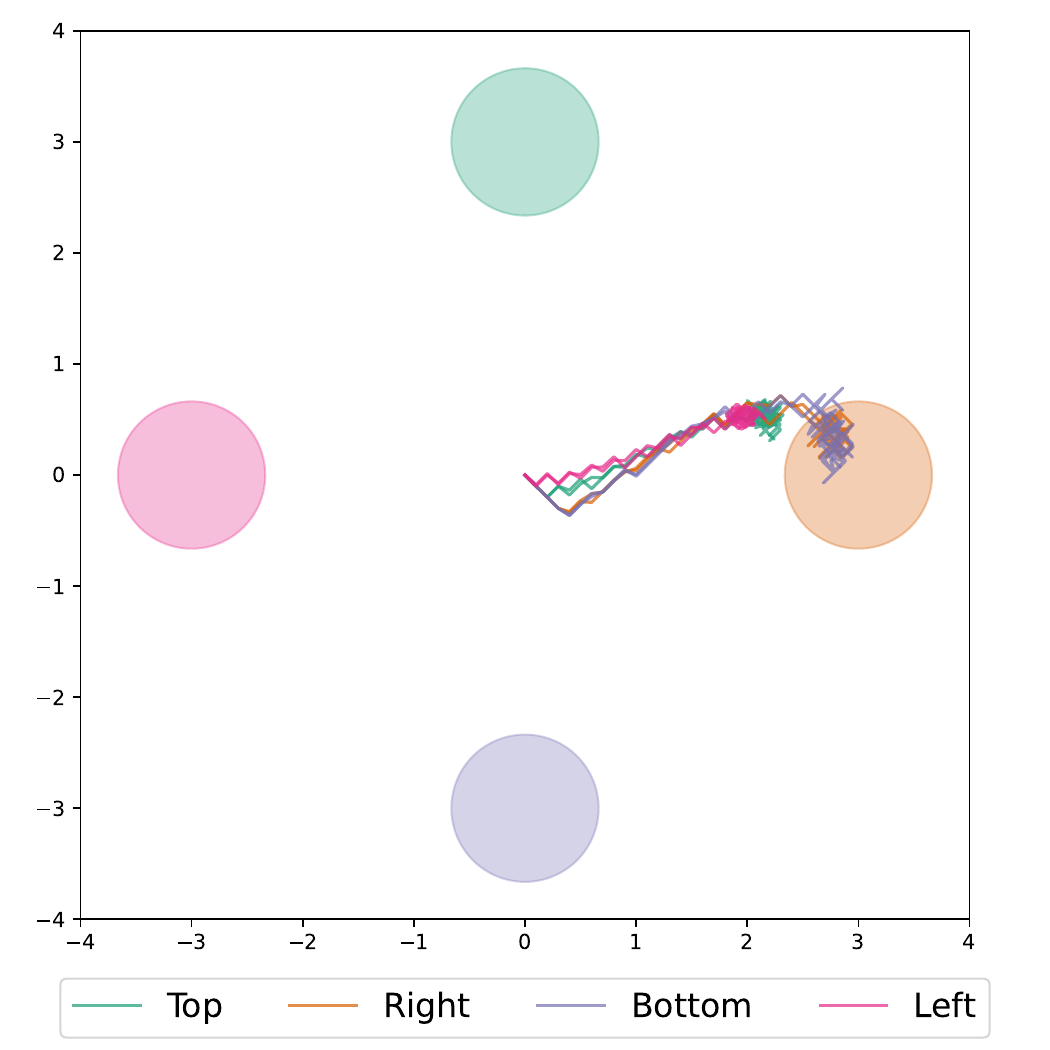} &
        \includegraphics[width=0.2\linewidth]{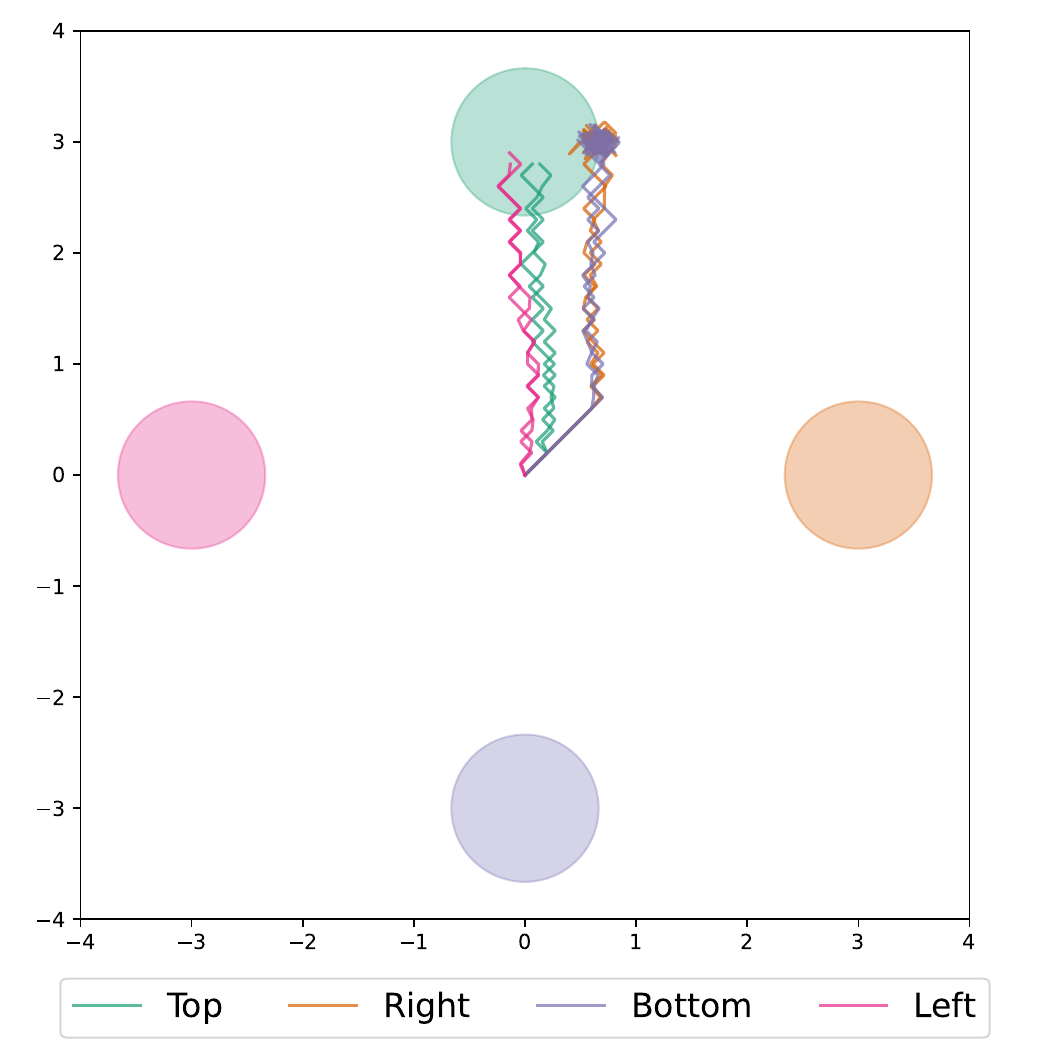} &
        \includegraphics[width=0.2\linewidth]{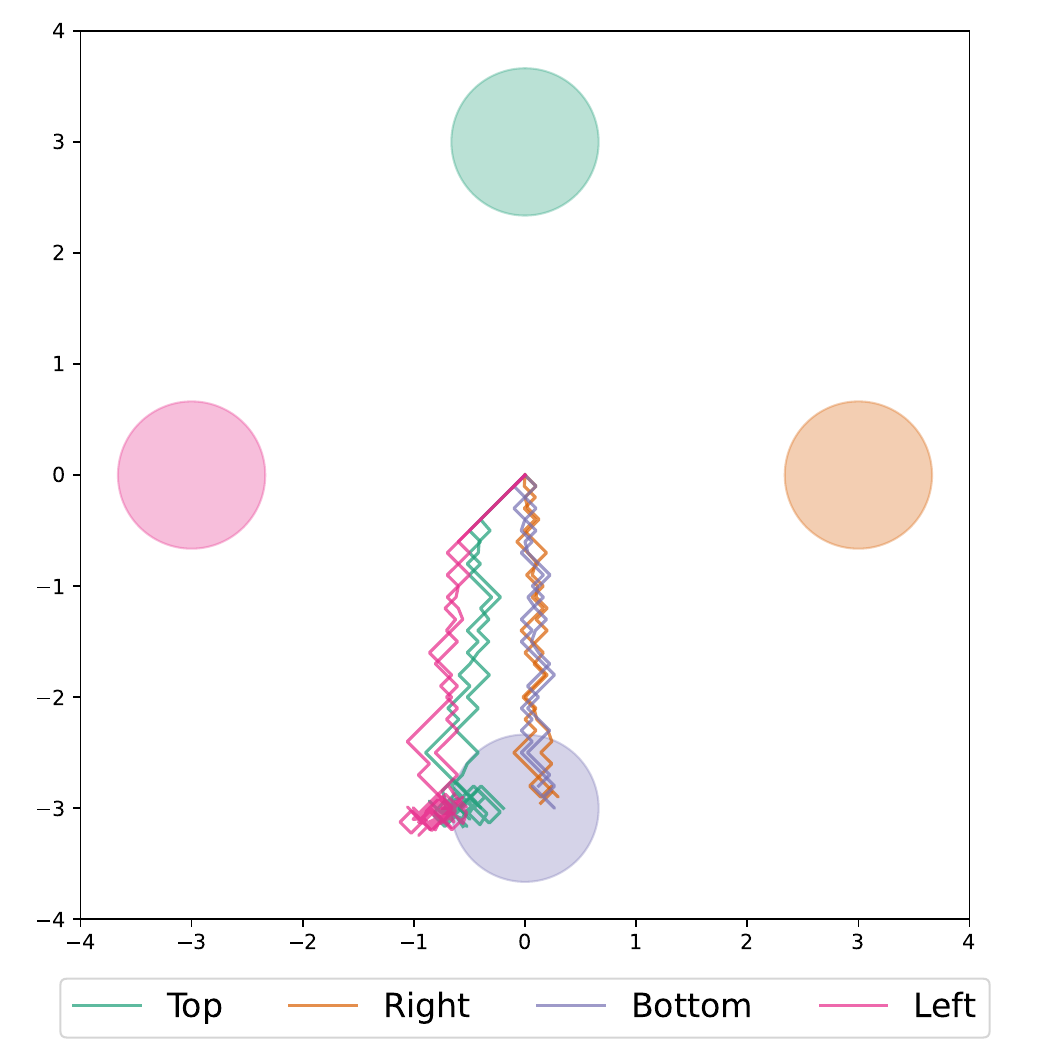} \\
        Goal: Left & Goal: Right & Goal: Top & Goal: Bottom\\
        & & & 
    \end{tabular}
    \label{fig:z2_importance_pointmass}}
    
    \subfigure[The effect of $z_3$ on the behavior of the environment. As we can notice from the above plots, $z_3$ seems to have a significant effect on the behavior of the agent. We hypothesize that the $z_3$ feature is used to determine the top-right and bottom-left segment of tasks. This would explain why some trajectories, specifically Top/Right and Bottom/Left are similar to each other.]{
    \begin{tabular}{cccc}
        \includegraphics[width=0.2\linewidth]{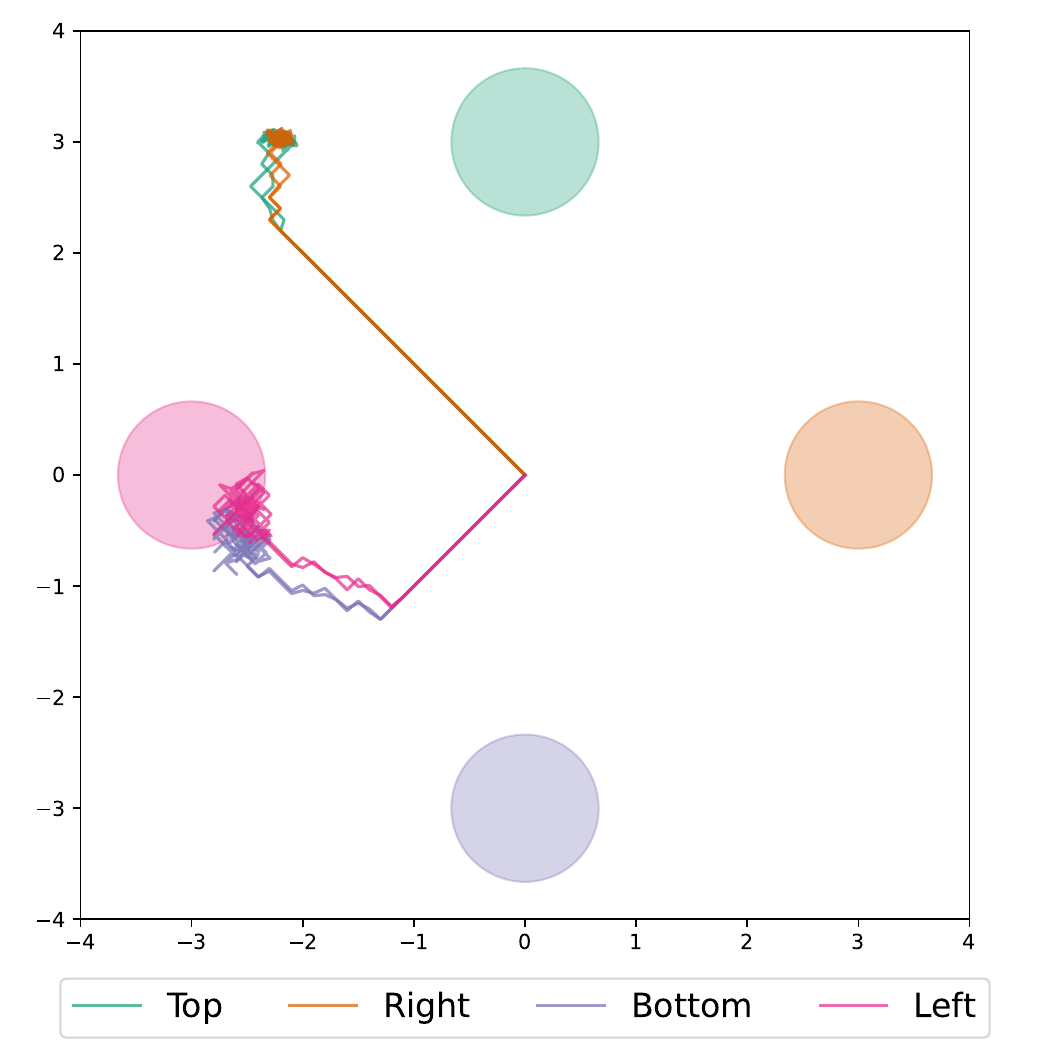} & 
        \includegraphics[width=0.2\linewidth]{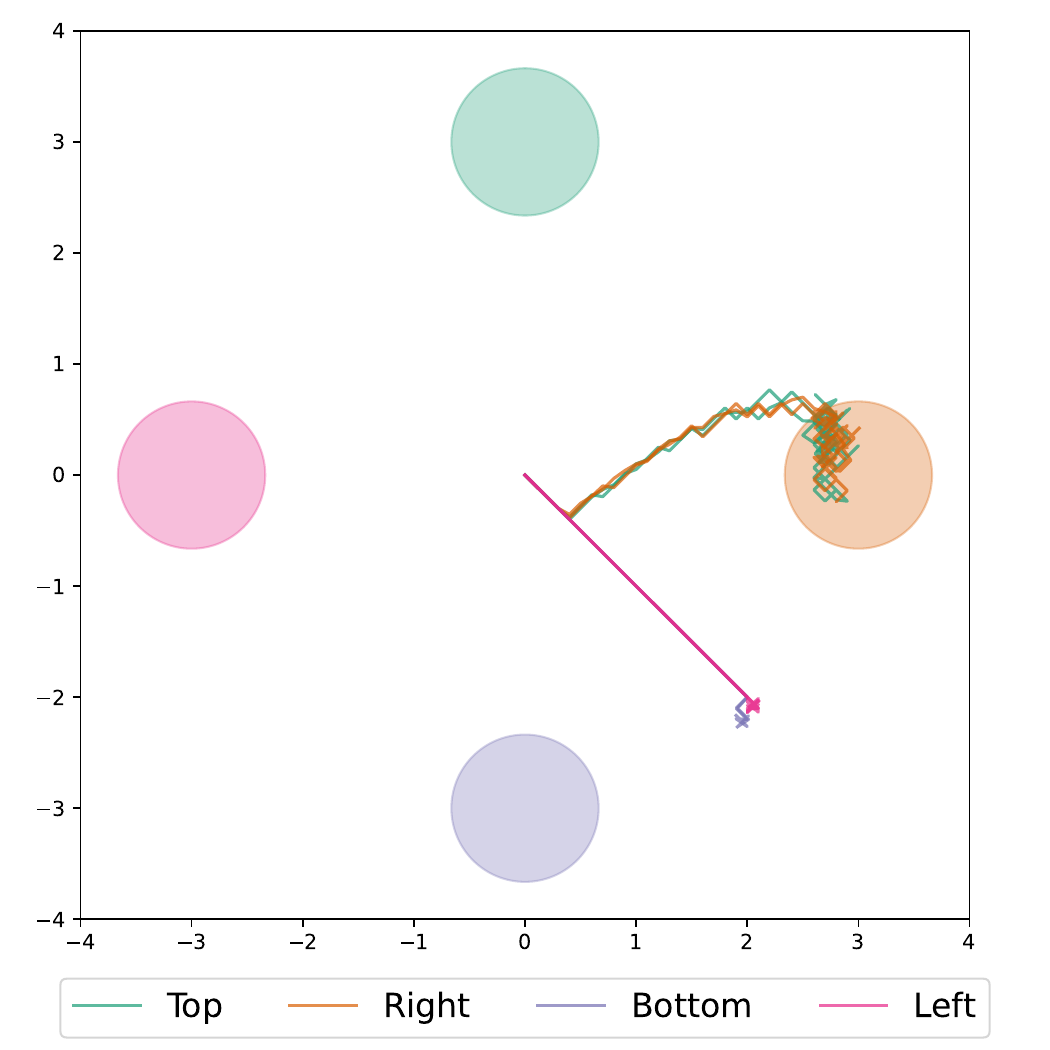} &
        \includegraphics[width=0.2\linewidth]{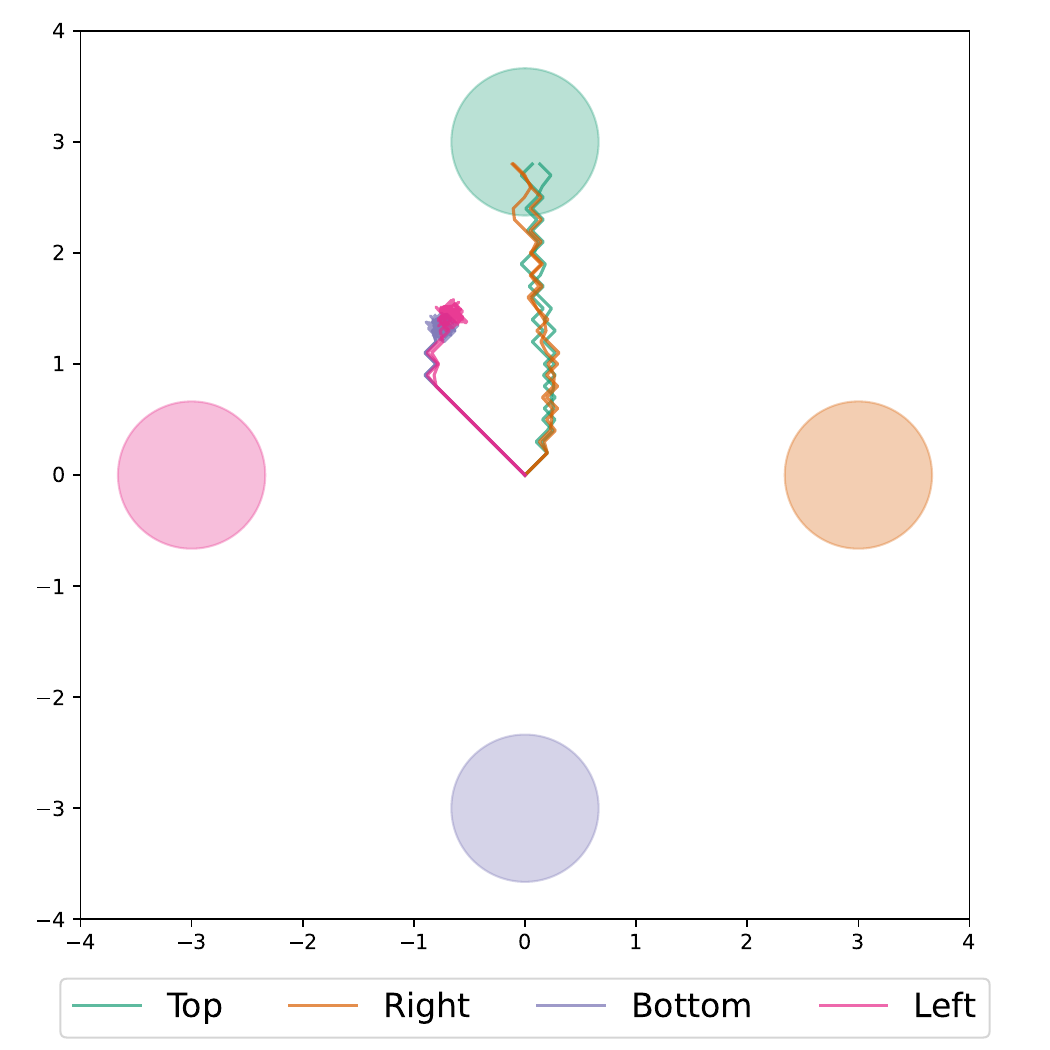} &
        \includegraphics[width=0.2\linewidth]{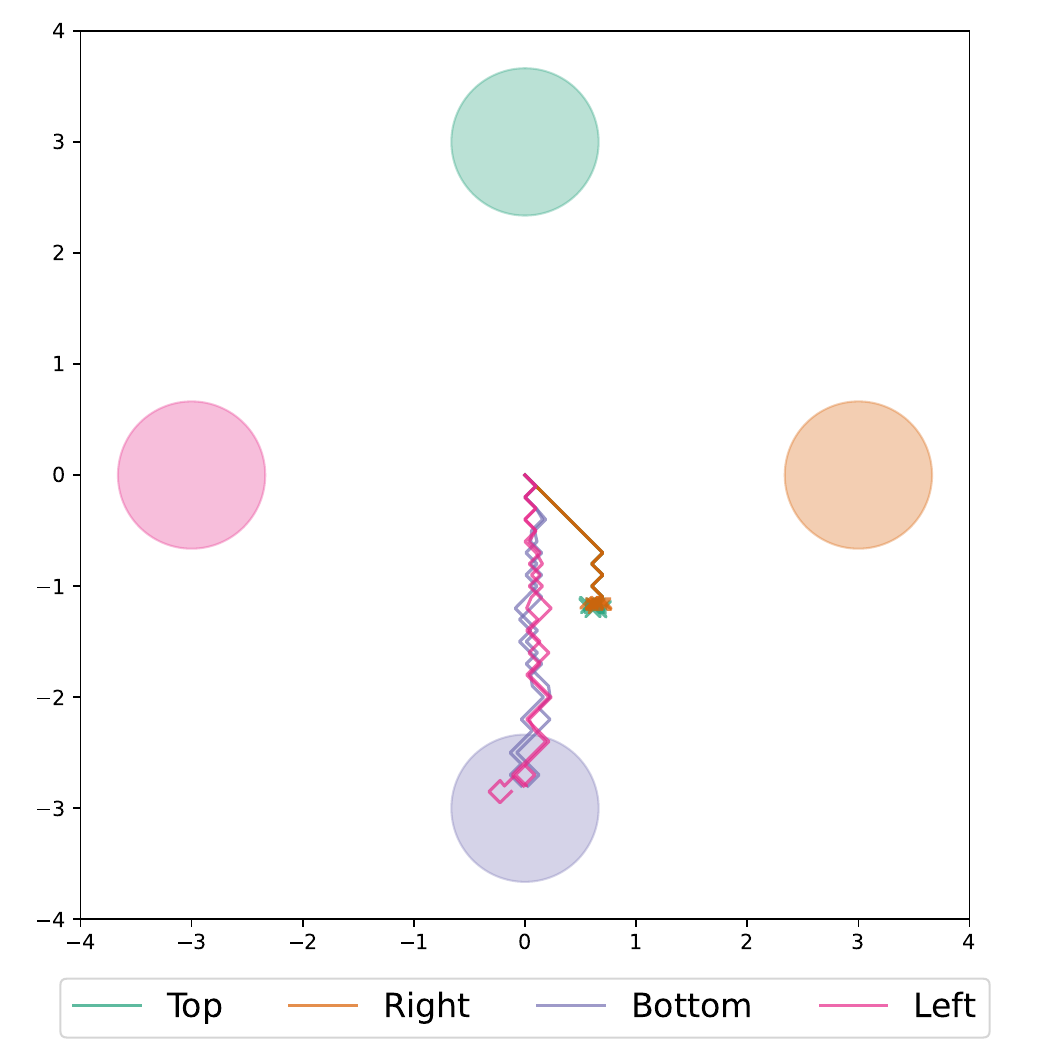} \\
        Goal: Left & Goal: Right & Goal: Top & Goal: Bottom\\
        & & & 
    \end{tabular}
    \label{fig:z3_importance_pointmass}}
    
    \subfigure[The effect of $z_4$ on the behavior of the environment. As we can notice from the above plots, $z_4$ seems to have a similar purpose as the $z_3$ feature. However, we notice a slight difference in the trajectory - $z_4$ enforces a shorter distance before the sudden turn in Left/Right, but a longer one in Top/Bottom. Whereas, $z_3$ does the exact opposite.]{
    \begin{tabular}{cccc}
        \includegraphics[width=0.2\linewidth]{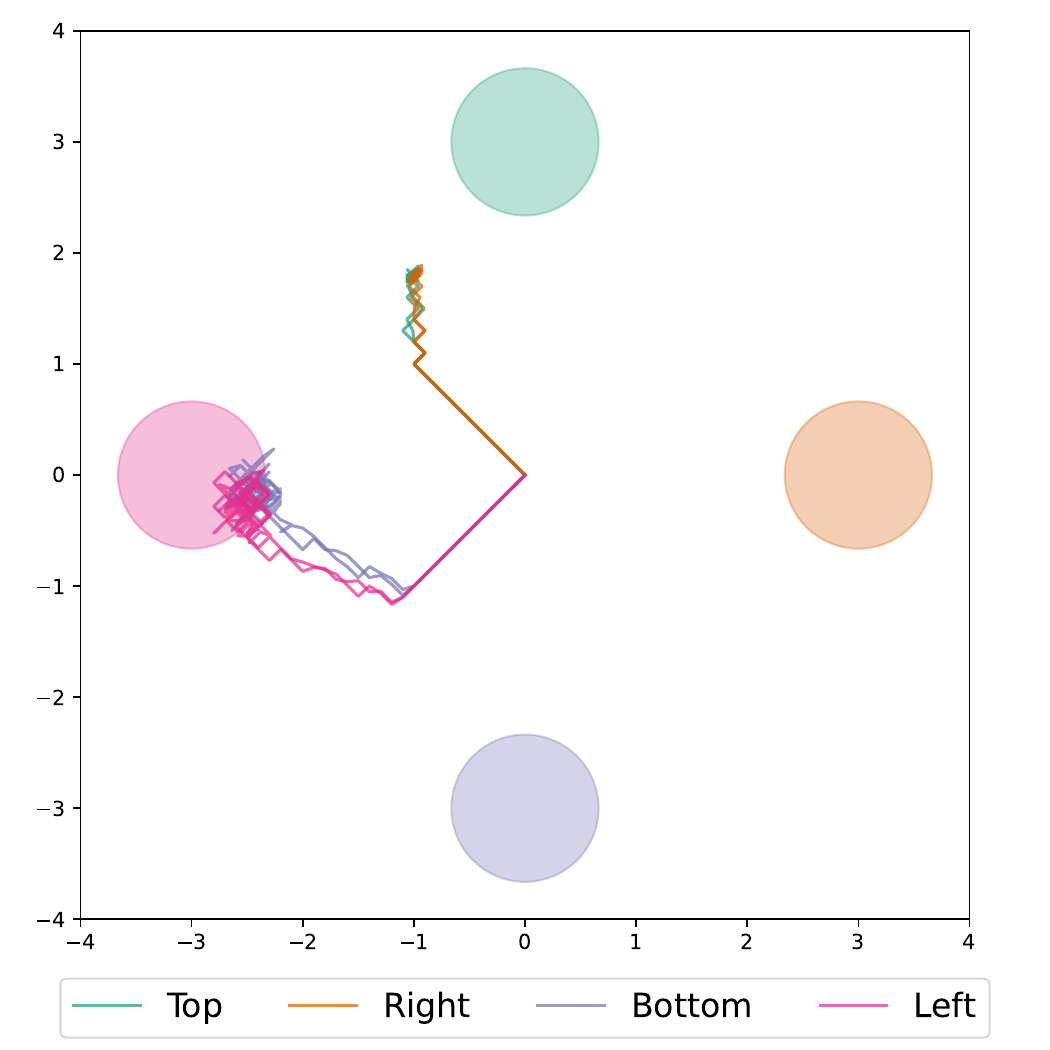} & 
        \includegraphics[width=0.2\linewidth]{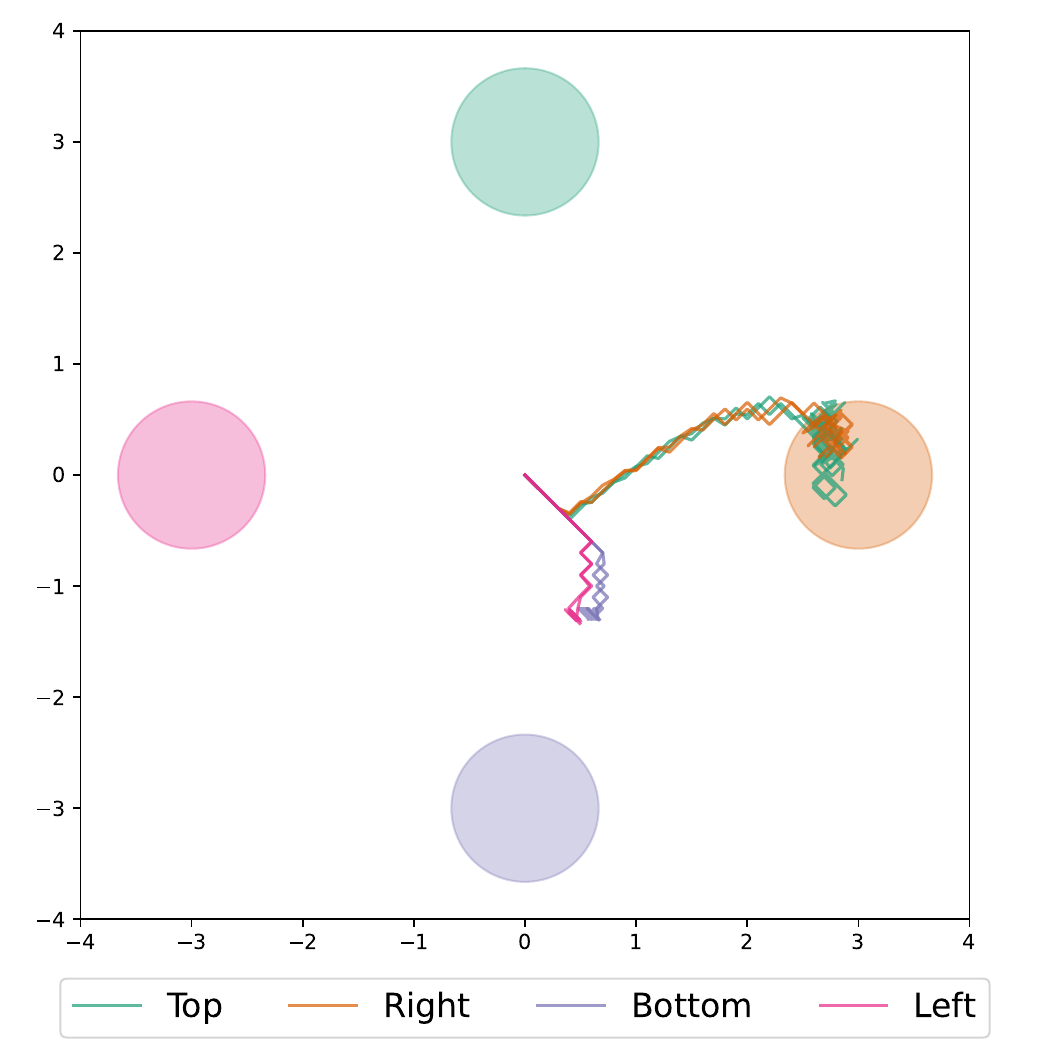} &
        \includegraphics[width=0.2\linewidth]{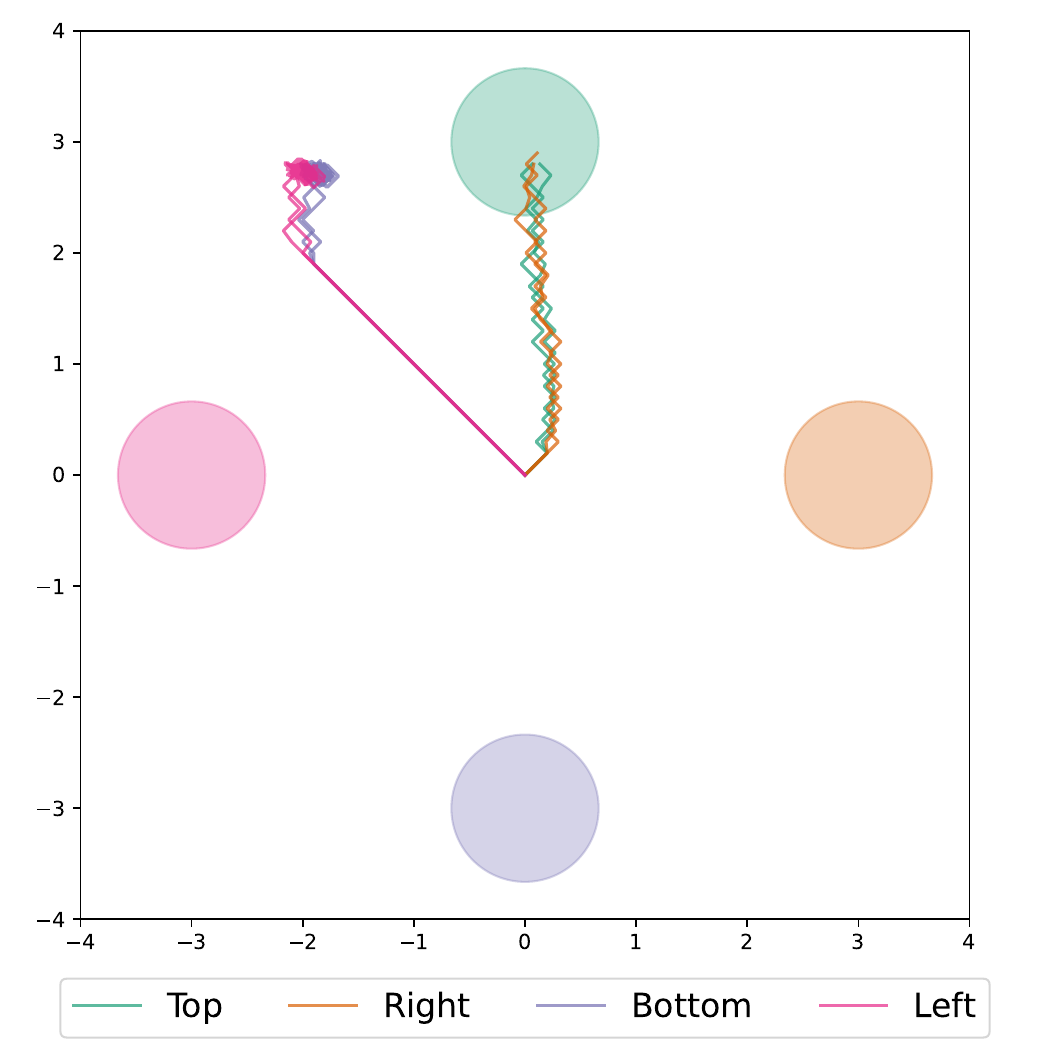} &
        \includegraphics[width=0.2\linewidth]{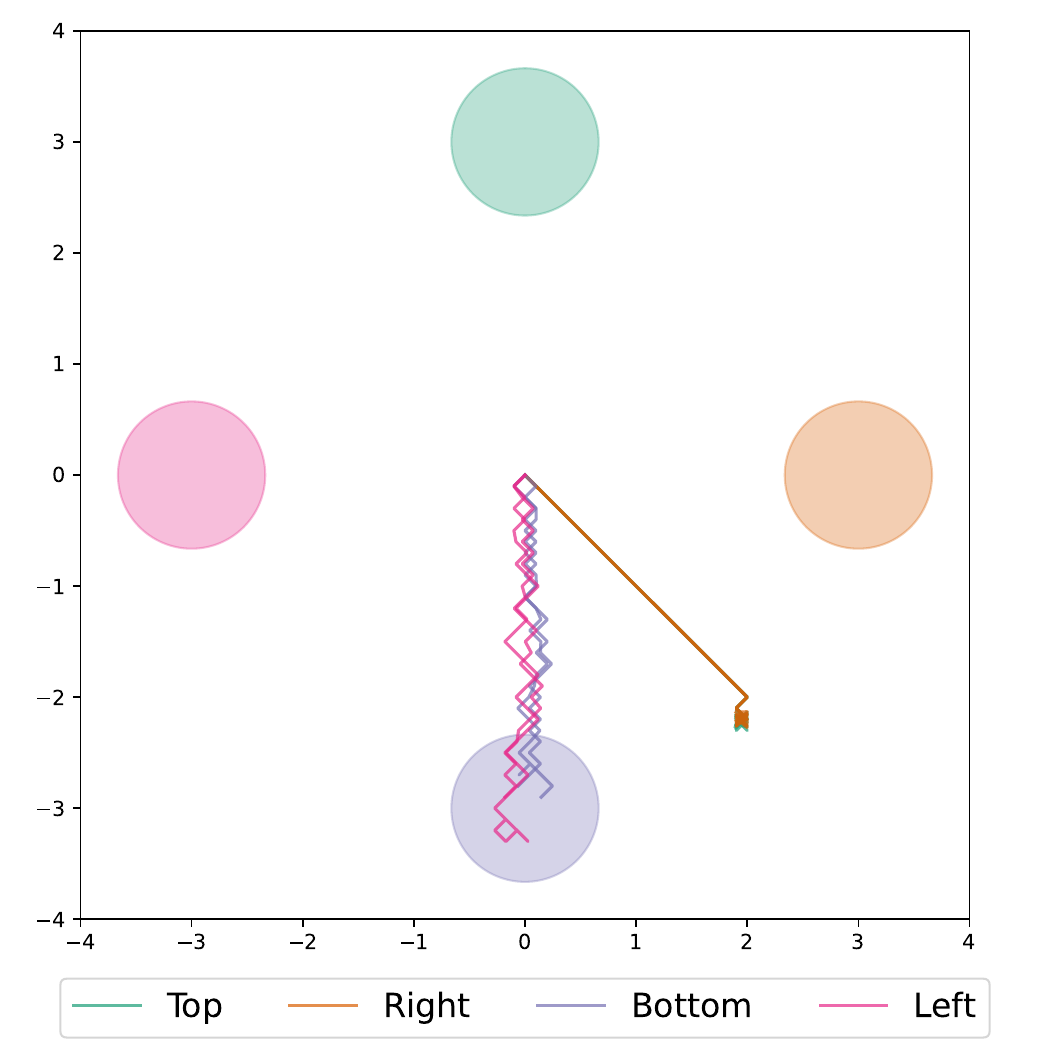} \\
        Goal: Left & Goal: Right & Goal: Top & Goal: Bottom\\
        & & & 
    \end{tabular}
    \label{fig:z4_importance_pointmass}}

    \caption{Input perturbation results on PointMass environment}
    \label{fig:input_perturb_pointmass}
\end{figure}

To better understand the effect of each feature on the behavior of the model, we use an input perturbation method. Here we fix all features to a given task: say Top, but vary a given feature within the skill to study the change in the agent's behavior. Figure~\ref{fig:input_perturb_pointmass} depicts the results from our input perturbation experiment.

\subsection{2-D Navigation Environment}
\label{appendix:navigation}

\begin{figure}[hbt!]
\centering
\subfigure[Goal: Right]{%
\includegraphics[width=0.22\linewidth]{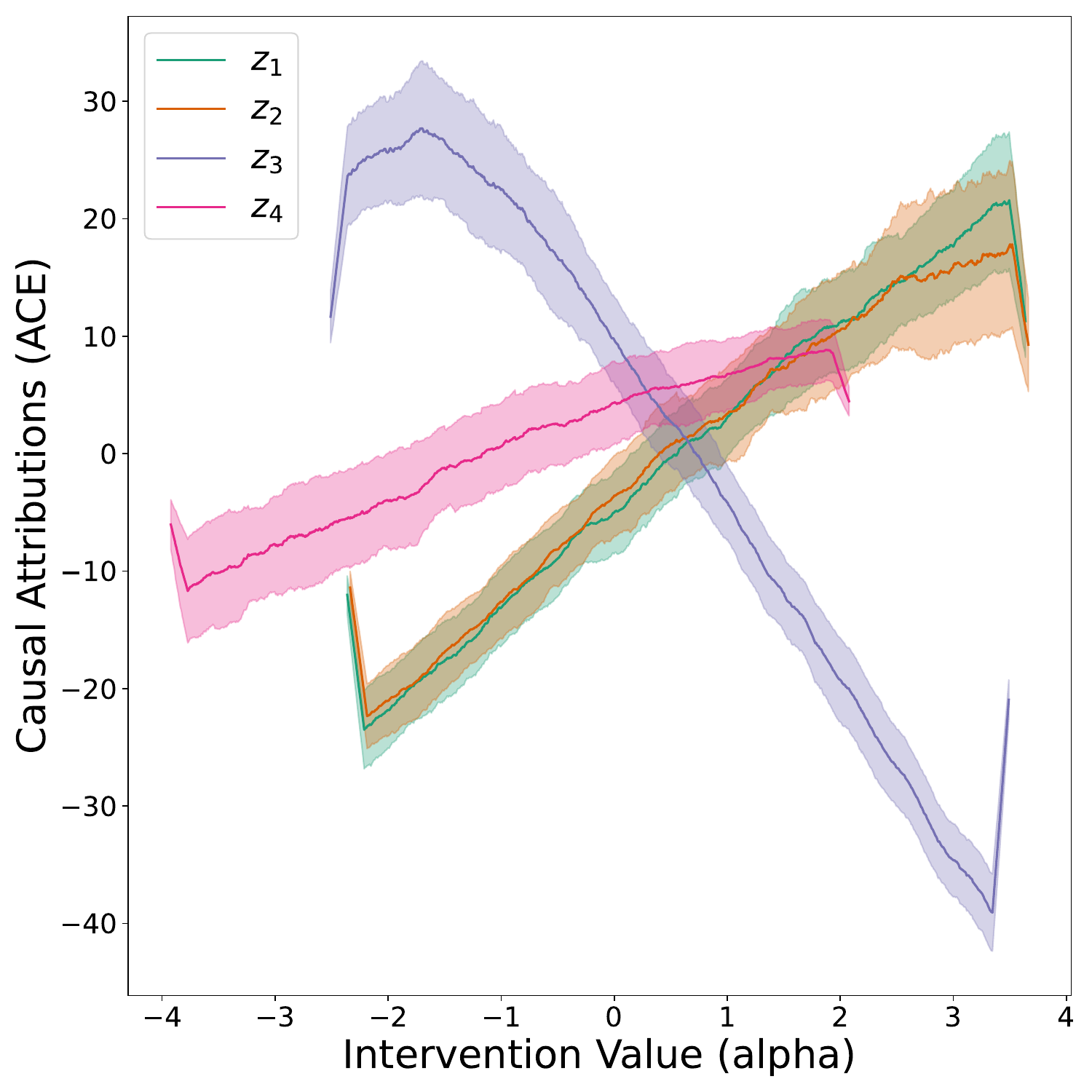}
\label{fig:ate_nav_right}}
\quad
\subfigure[Goal: Top]{%
\includegraphics[width=0.22\linewidth]{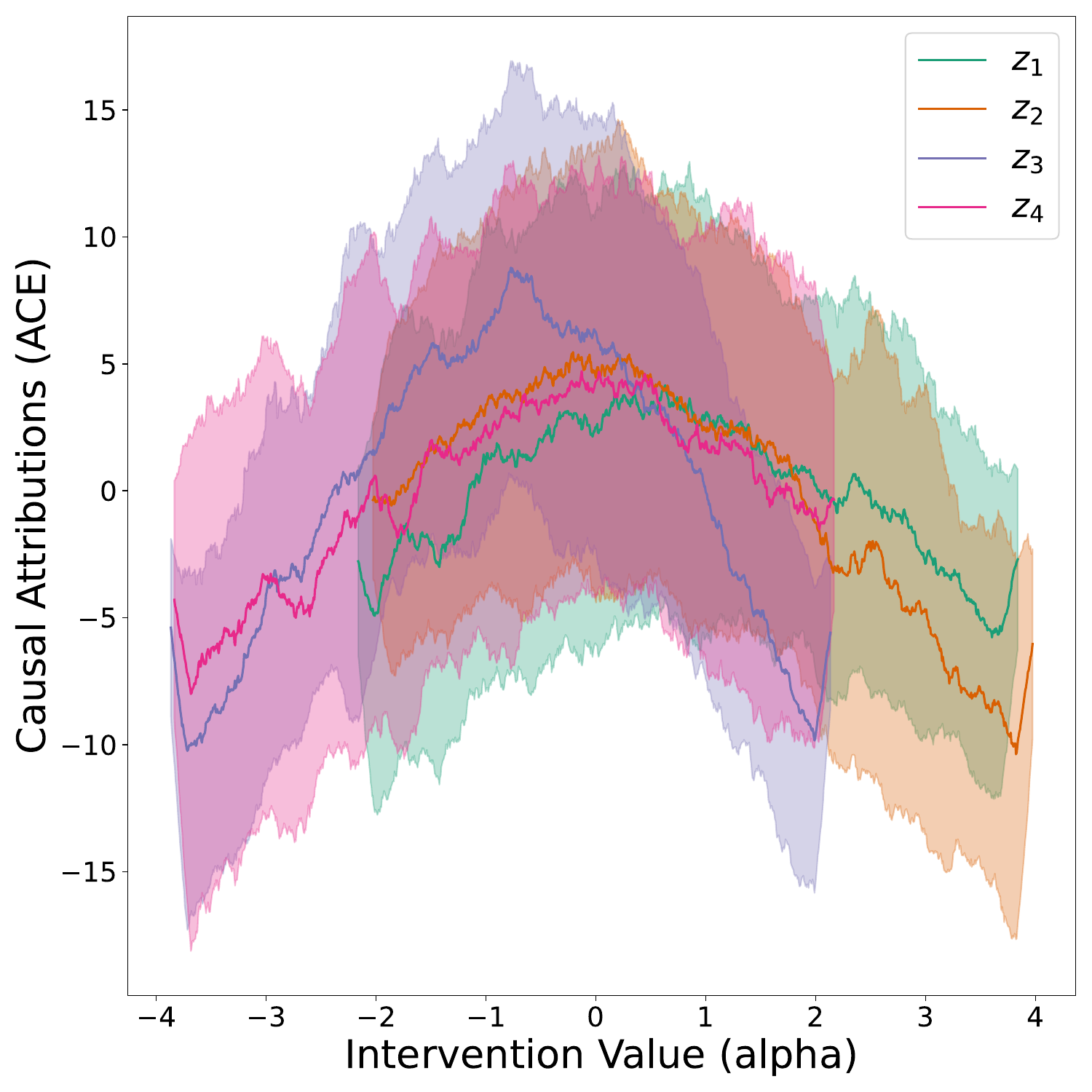}
\label{fig:ate_nav_top}}
\quad
\subfigure[Goal: Bottom]{%
\includegraphics[width=0.22\linewidth]{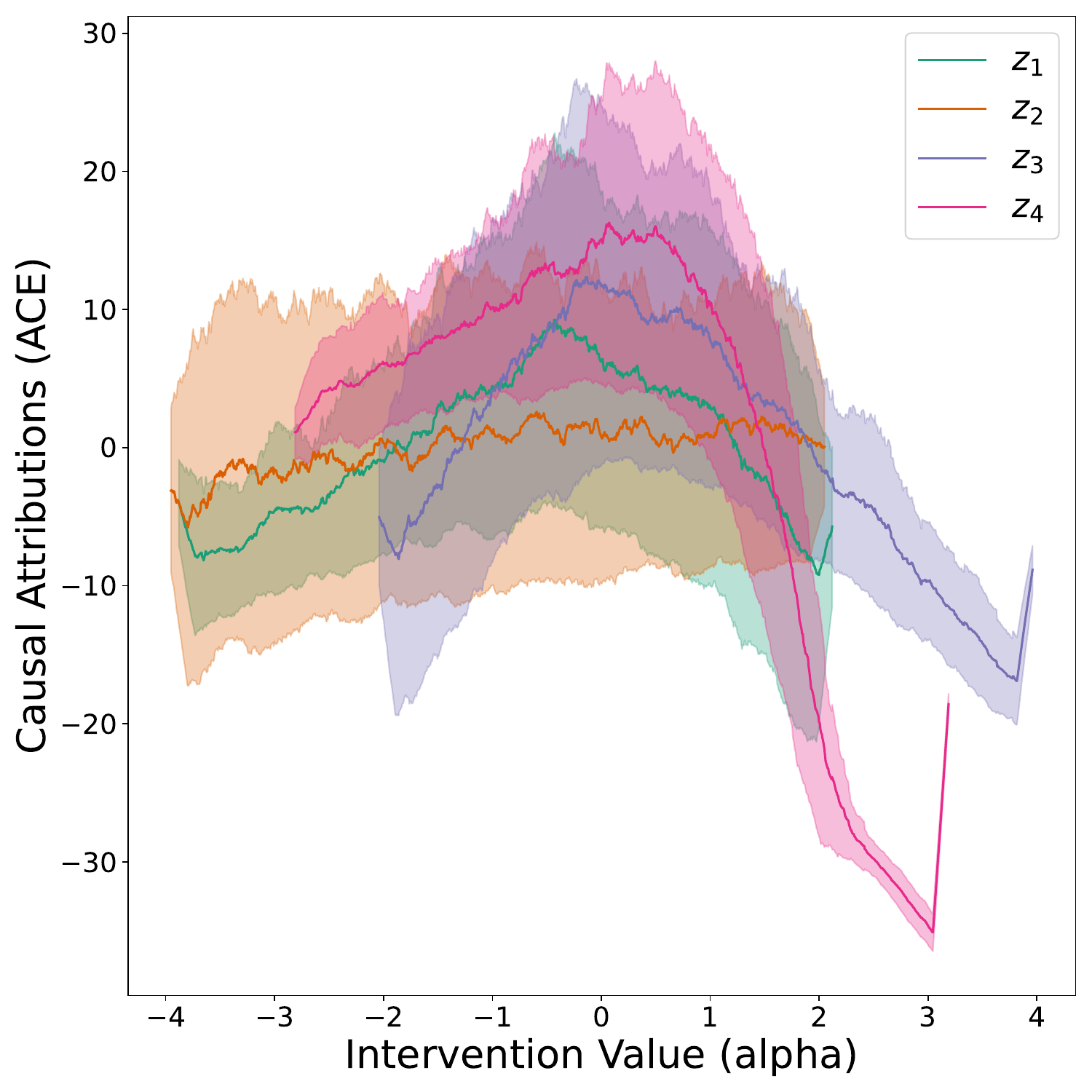}
\label{fig:ate_nav_bottom}}

\caption{The Average Causal Effect of the latent skills for different tasks in the 2-D Navigation Environment. From the above plots, we notice that there is a lot of overlap and noise in the ACE, especially in the case of Figure~\ref{fig:ate_nav_top} and Figure~\ref{fig:ate_nav_bottom}. This is expected since the model needs to learn to diverge from its initial path after a few time steps, bringing exploration only after it crosses the corridor. We hypothesize that we do not observe the same noise in Figure~\ref{fig:ate_nav_right} since there is no divergence from the path.}
\label{fig:ace_navigation}
\end{figure}

\begin{figure}[hbt!]
\centering
\includegraphics[width=0.5\linewidth]{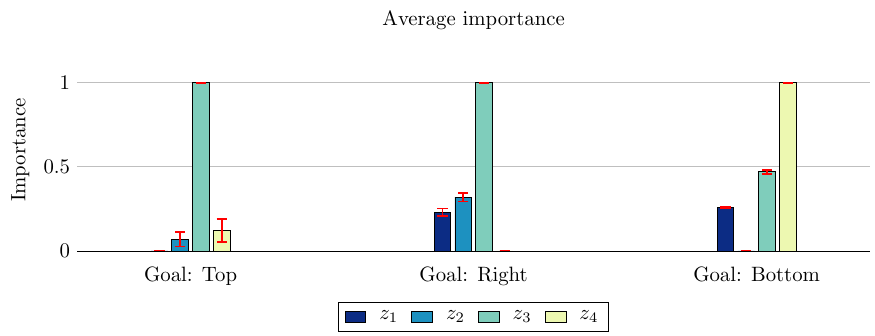}
\caption{The normalized importance of each latent feature on the 2-D navigation tasks across 5 different seeds.}
\label{fig:imp_navigation}
\end{figure}

The average importance of each feature is shown in Figure~\ref{fig:imp_navigation}.

From the ACE plots, we can successfully assert that feature $z_2$ is an auxiliary variable and does not play any significant role in the model's behavior since the ACE value is close to 0. Furthermore, we notice that the $z_3$ and $z_4$ causal analysis are pretty similar, and $z_1$ seems to be an essential feature in the pool of latent features. 

\begin{figure}[hbt!]
    \centering
    \subfigure[The effect of $z_1$ on the behavior of the environment. As we can notice from the above plots, $z_1$ does not seem to have a significant effect on the behavior of the agent. We hypothesize that the $z_1$ feature could have low importance and unnecessary, or could be a basic skill used across all tasks. This would explain why perturbation of the $z_1$ feature does not lead to any difference in behavior, i.e. not a discriminating feature across tasks.]{
    \begin{tabular}{ccc}
        \includegraphics[width=0.15\linewidth]{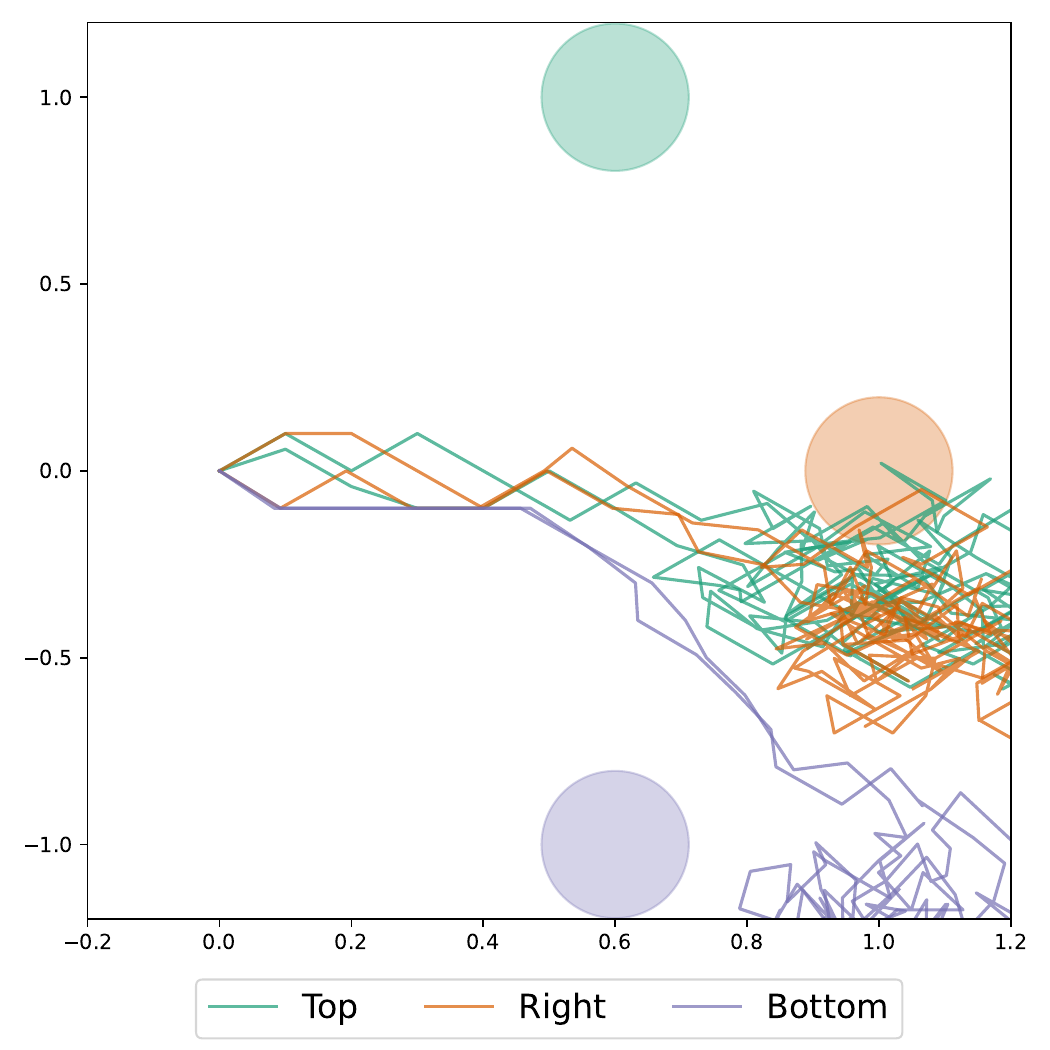} &
        \includegraphics[width=0.15\linewidth]{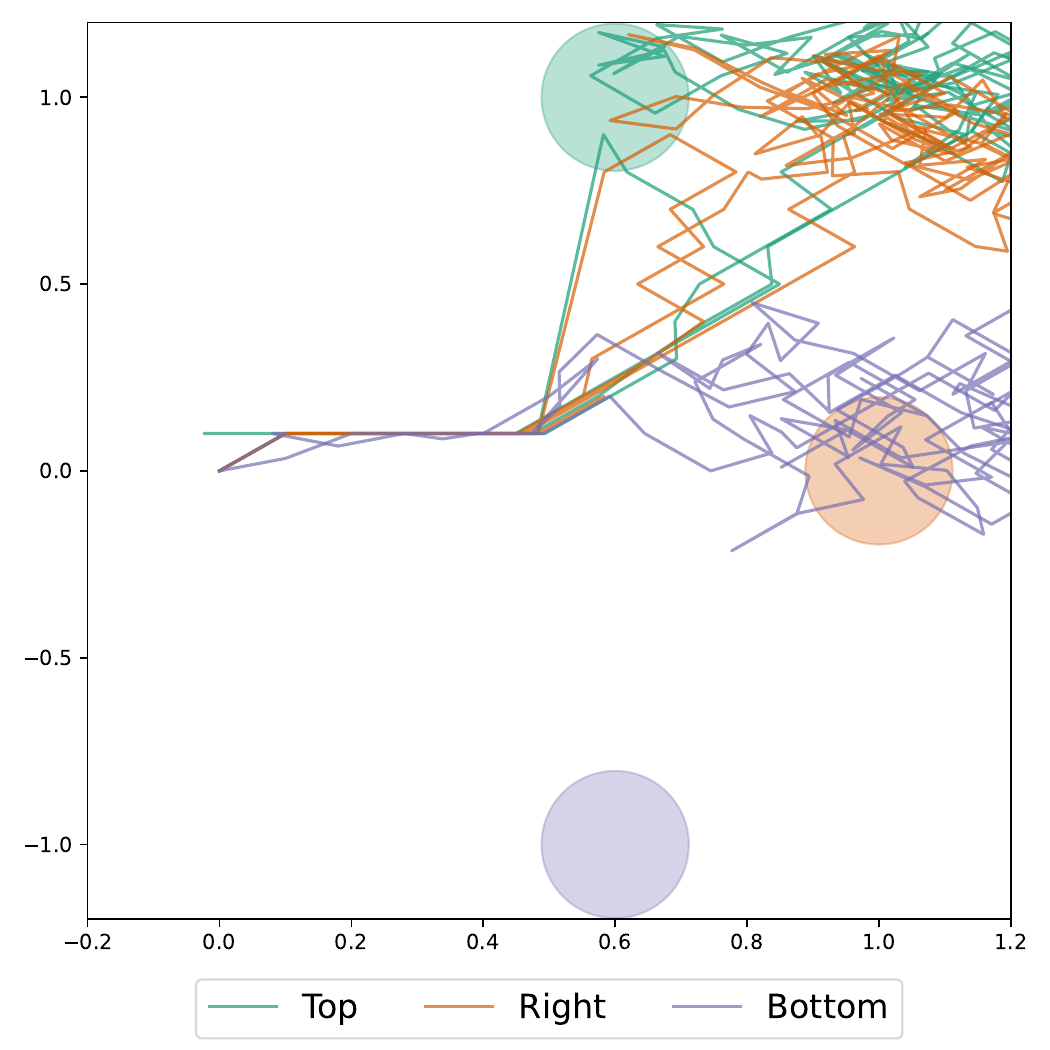} &
        \includegraphics[width=0.15\linewidth]{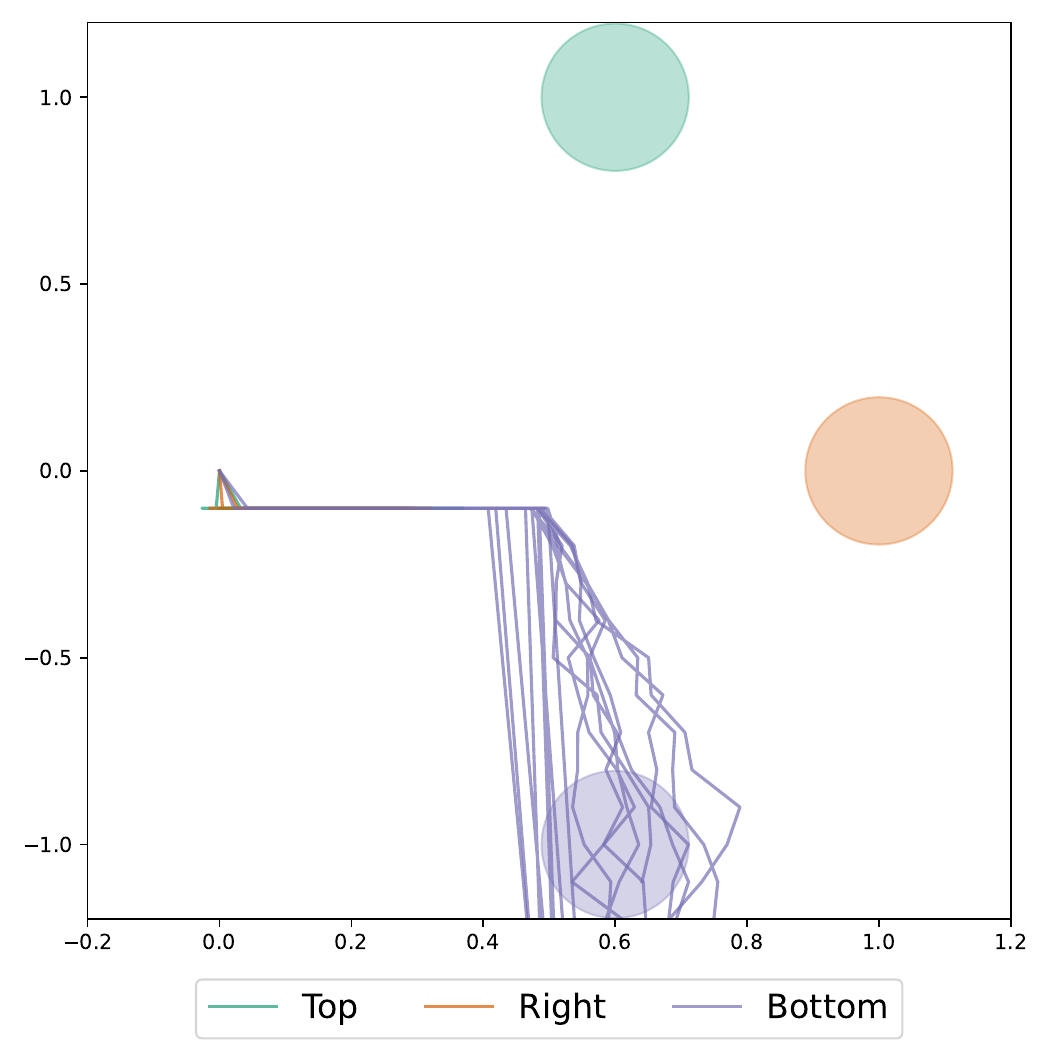} \\
        Goal: Right & Goal: Top & Goal: Bottom\\
        & & 
    \end{tabular}
    \label{fig:z1_importance_navigation}}
    
    \subfigure[The effect of $z_2$ on the behavior of the environment. As we can notice from the above plots, $z_2$ seems to have little to no effect on the behavior of the agent. This feature seems to play a very similar role to $z_1$ as discussed earlier.]{
    \begin{tabular}{ccc}
        \includegraphics[width=0.15\linewidth]{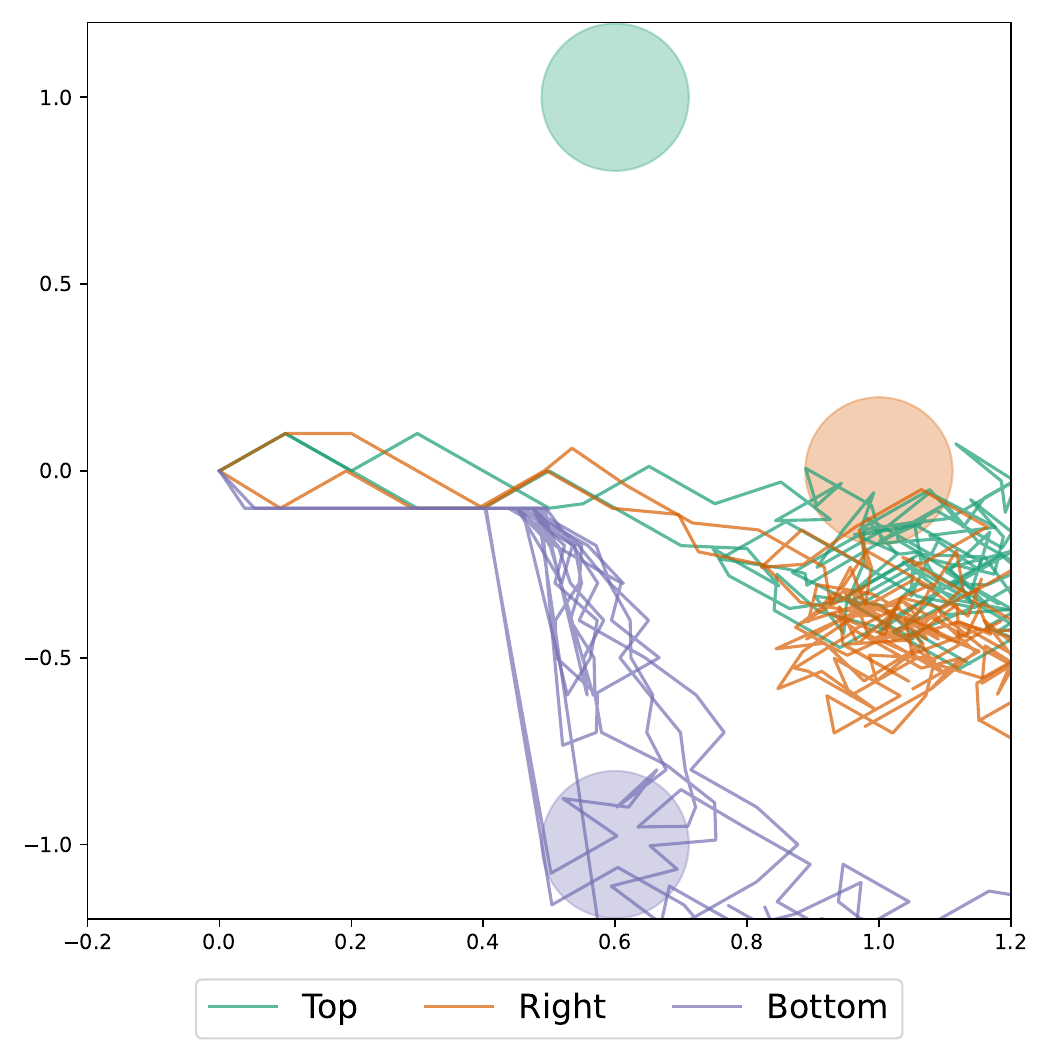} &
        \includegraphics[width=0.15\linewidth]{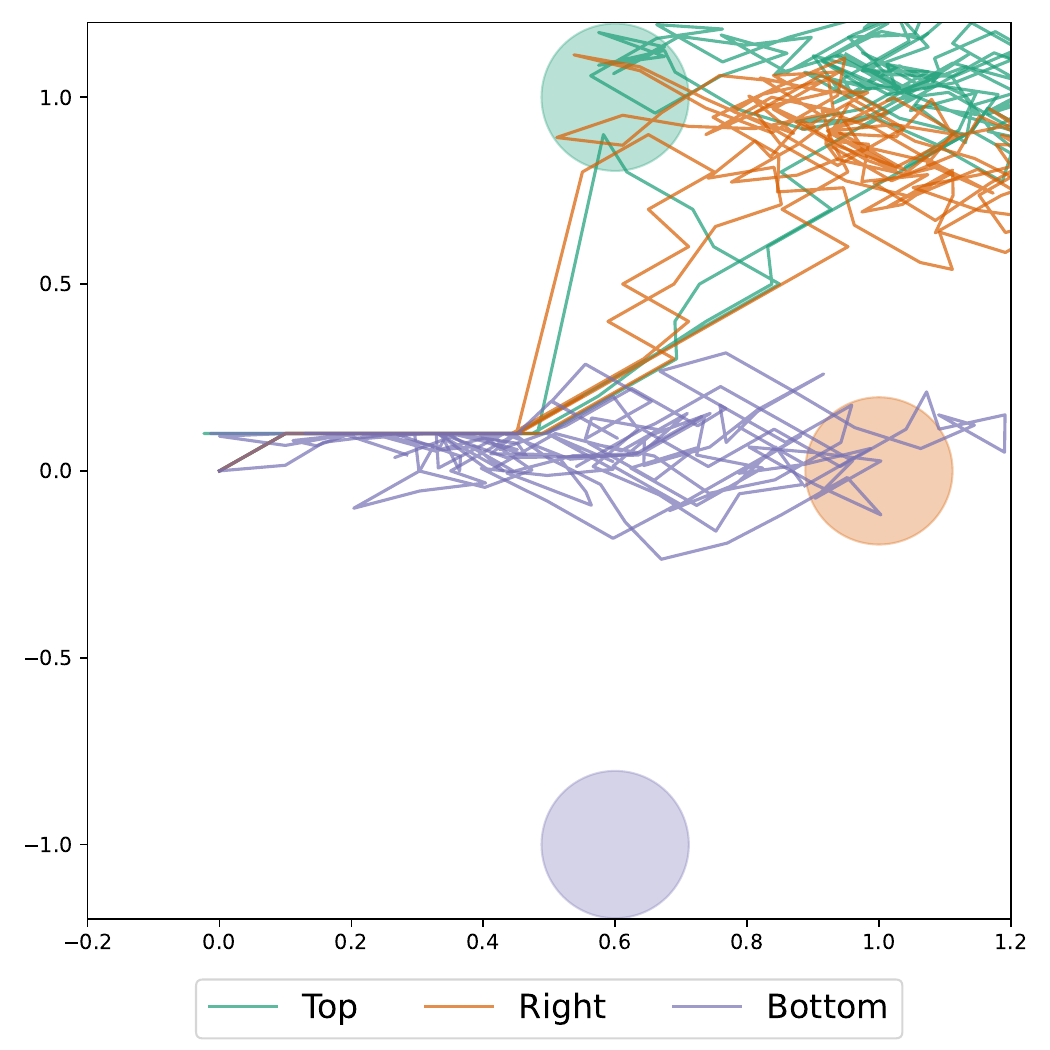} &
        \includegraphics[width=0.15\linewidth]{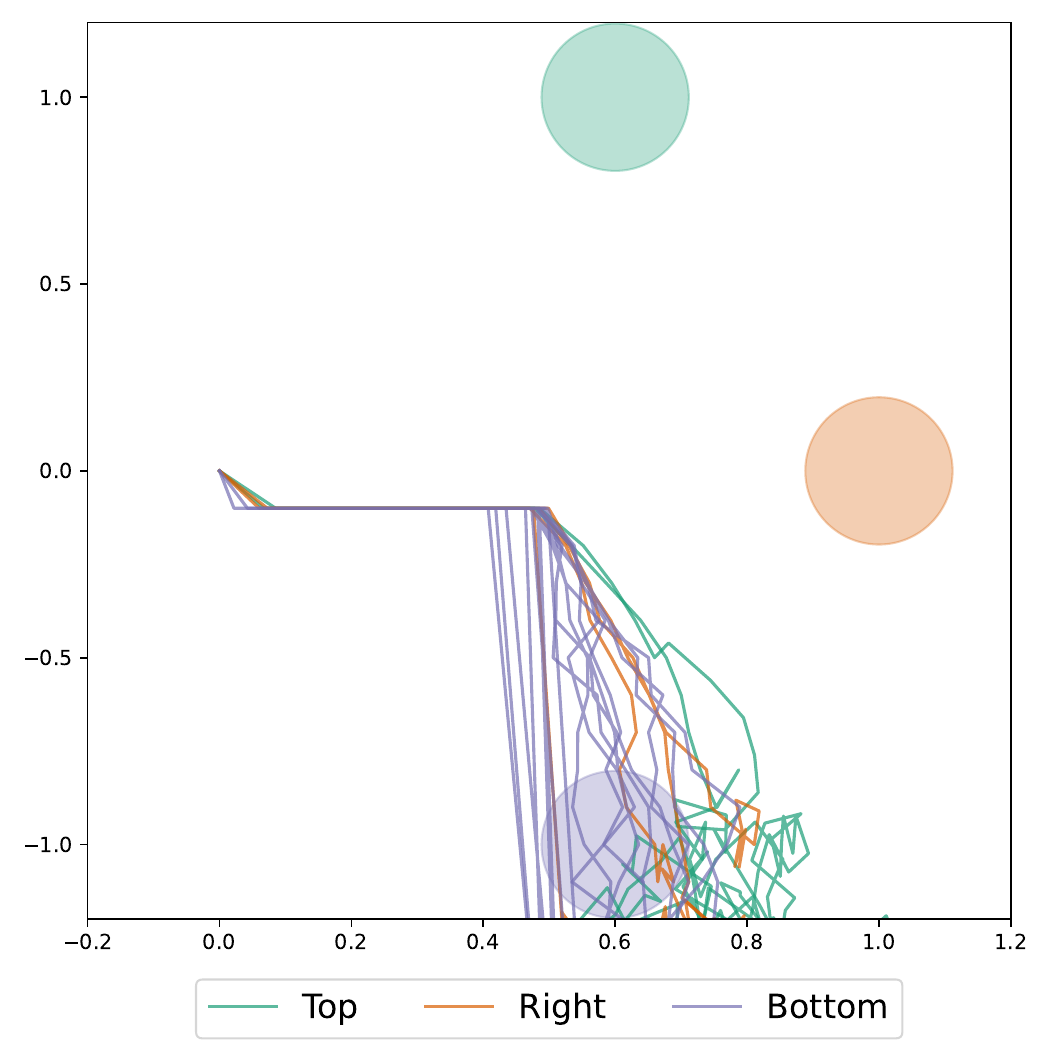} \\
        Goal: Right & Goal: Top & Goal: Bottom\\
        & & 
    \end{tabular}
    \label{fig:z2_importance_navigation}}
    
    \subfigure[The effect of $z_3$ on the behavior of the environment. As we can notice from the above plots, $z_3$ seems to have a significant effect on the behavior of the agent. We hypothesize that the $z_3$ feature helps decide the behavior of the agent for the Top and Right goals. However, when it comes to the Bottom task, a feature other than $z_3$ overwrites this behavior and forces the trajectory to go towards the Bottom goal.]{
    \begin{tabular}{ccc}
        \includegraphics[width=0.15\linewidth]{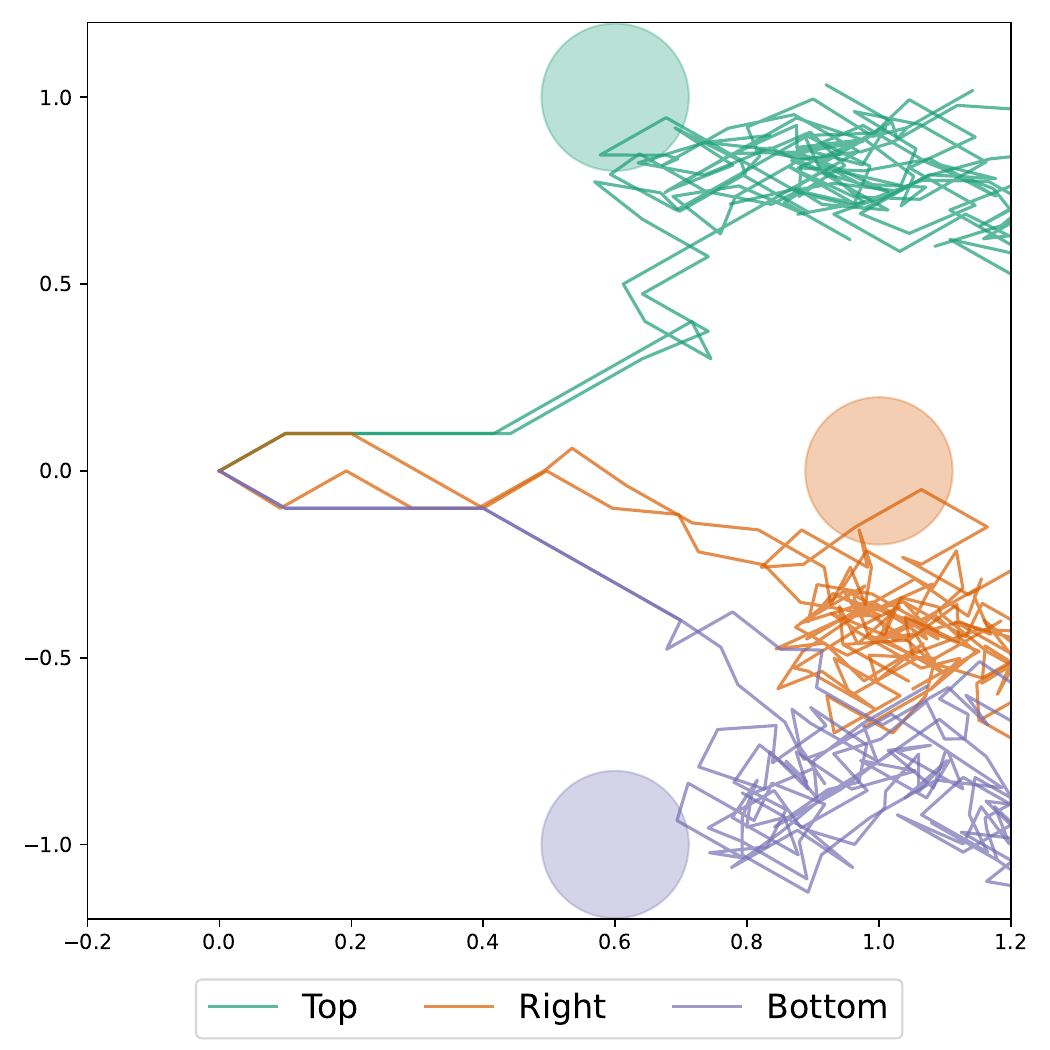} &
        \includegraphics[width=0.15\linewidth]{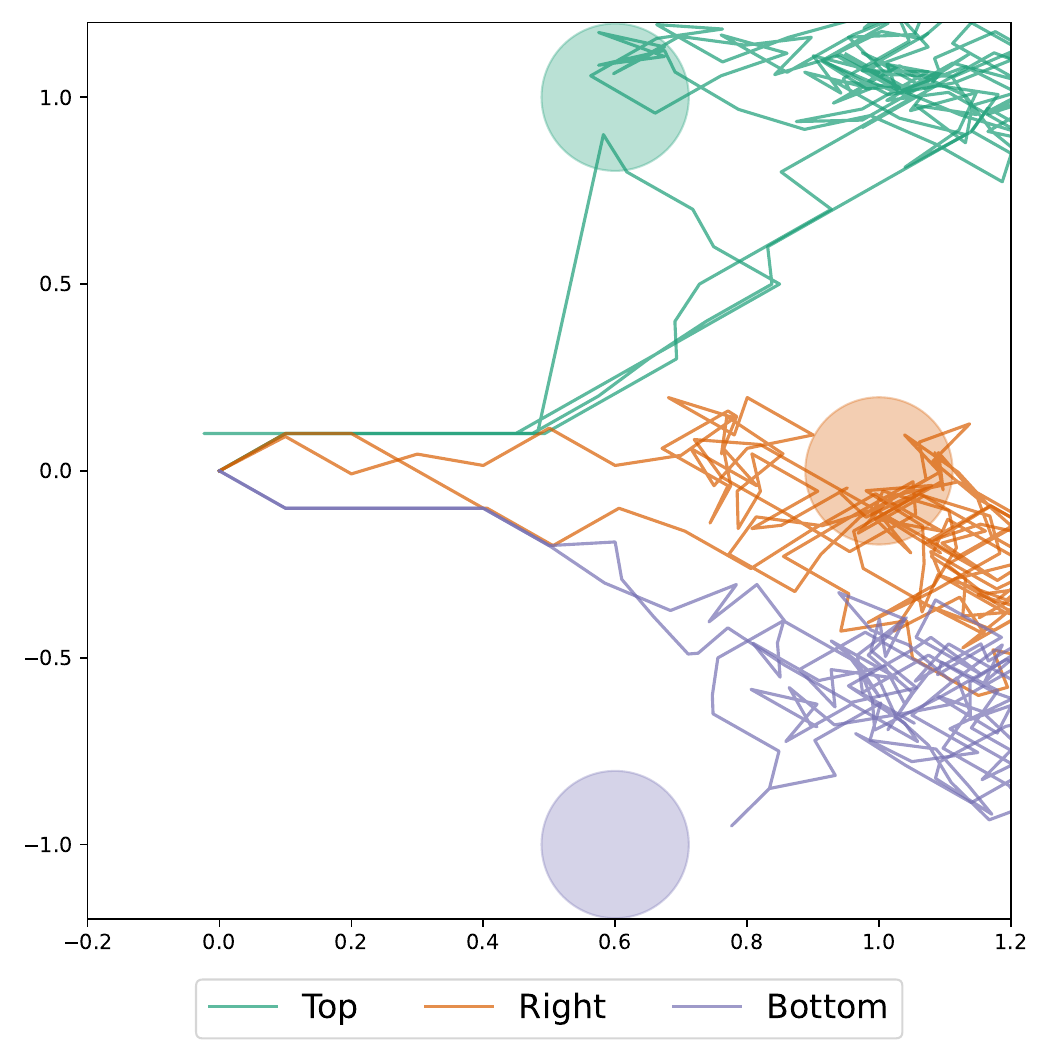} &
        \includegraphics[width=0.15\linewidth]{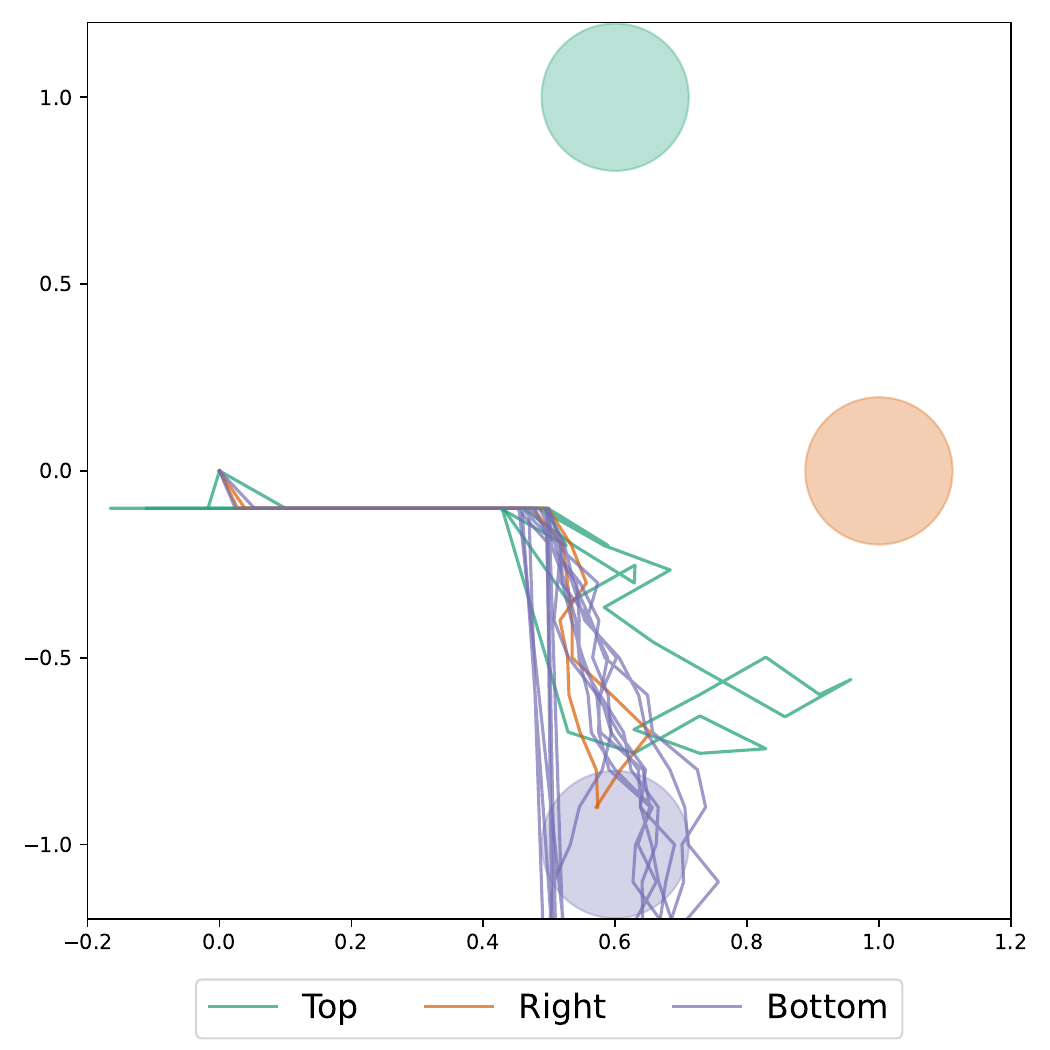} \\
        Goal: Right & Goal: Top & Goal: Bottom\\
        & & 
    \end{tabular}
    \label{fig:z3_importance_navigation}}
    
    \subfigure[The effect of $z_4$ on the behavior of the environment. As we can notice from the above plots, $z_4$ seems to be the only feature that helps propel the trajectory towards the Bottom goal. This is the feature that overwrites the effect of $z_3$ feature and forces the action to be towards the destined goal.]{
    \begin{tabular}{ccc}
        \includegraphics[width=0.15\linewidth]{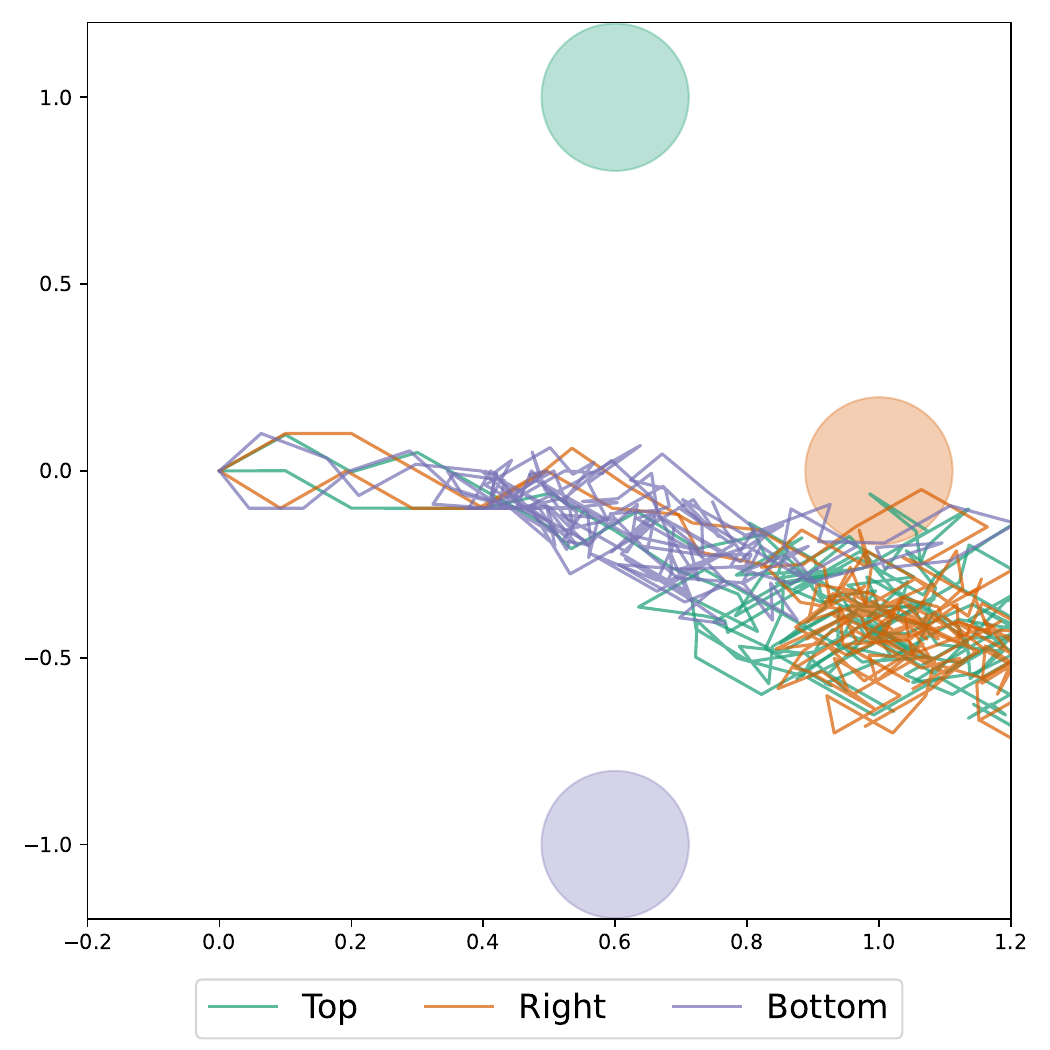} &
        \includegraphics[width=0.15\linewidth]{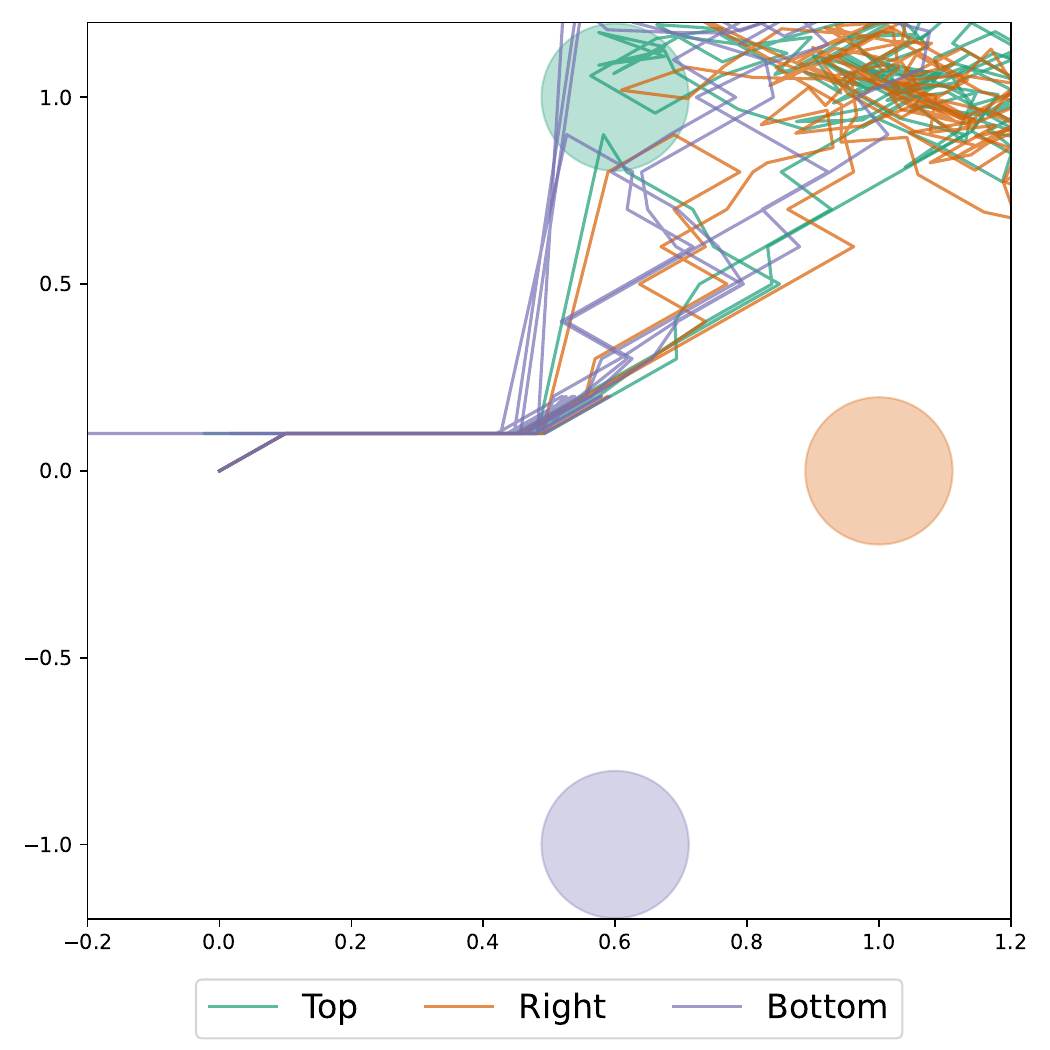} &
        \includegraphics[width=0.15\linewidth]{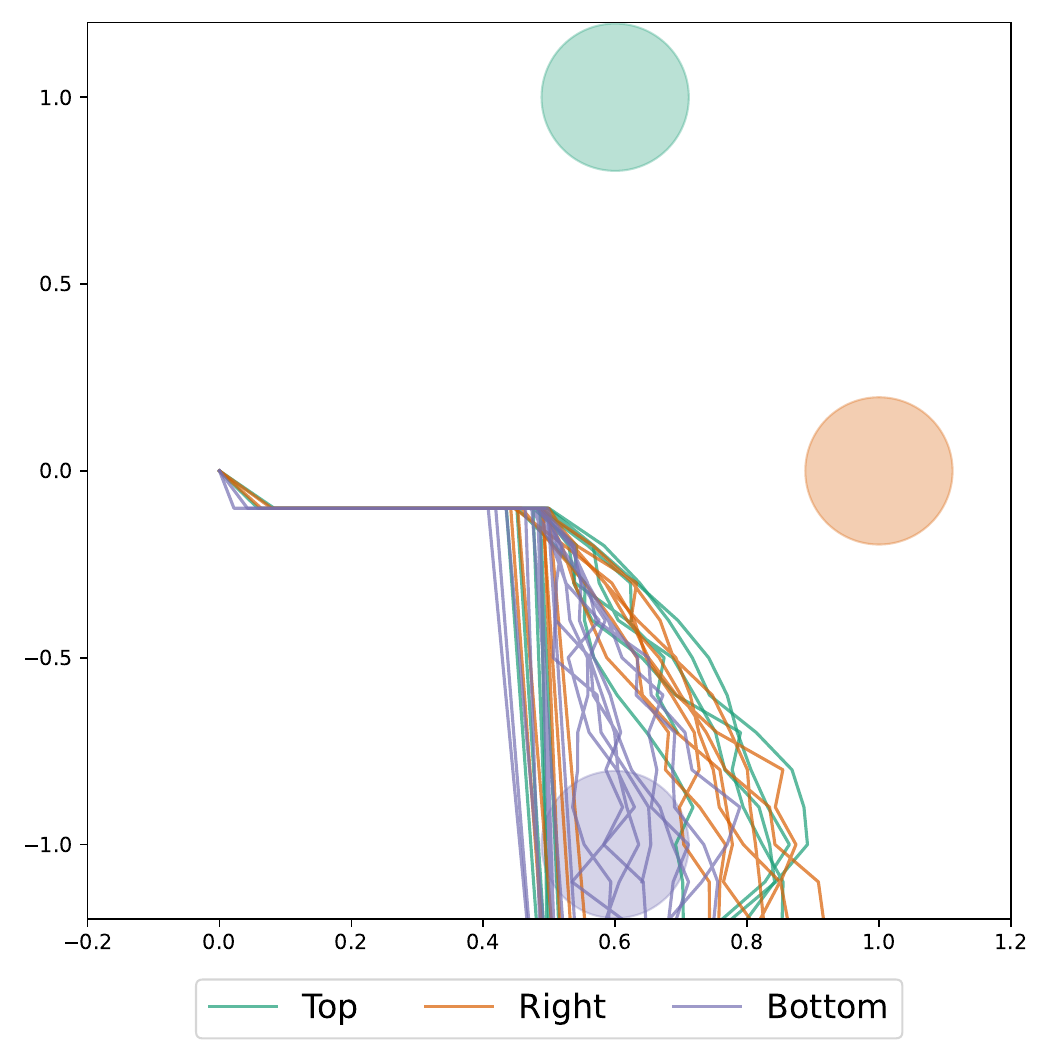} \\
        Goal: Right & Goal: Top & Goal: Bottom\\
        & & 
    \end{tabular}
    \label{fig:z4_importance_navigation}}

    \caption{Input perturbation results on 2-D Navigation environment}
    \label{fig:input_perturb_navigation}
\end{figure}

To better understand the effect of each feature on the behavior of the model, we use an input perturbation method. Here we fix all features to a given task: say Top, but vary a given feature within the skill to study the change in the agent's behavior. Figure~\ref{fig:input_perturb_navigation} depicts the results from our input perturbation experiment.

\subsection{Meta-World(MT5) Environment}
\label{appendix:mt5}

\begin{figure}[hbt!]
\centering
\subfigure[Push]{%
\includegraphics[width=0.16\linewidth]{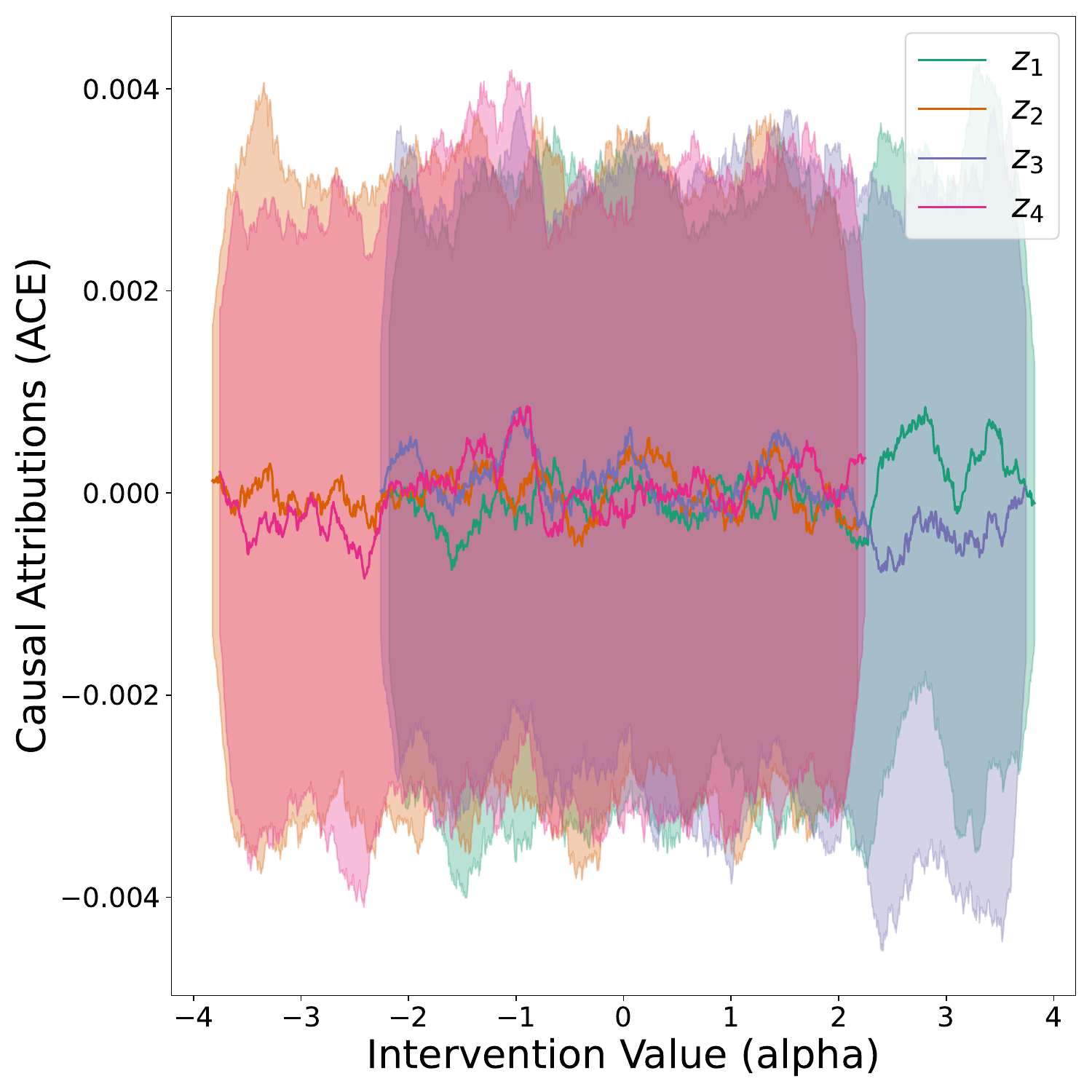}
\label{fig:ate_mt5_push}}
\quad
\subfigure[Window Open]{%
\includegraphics[width=0.16\linewidth]{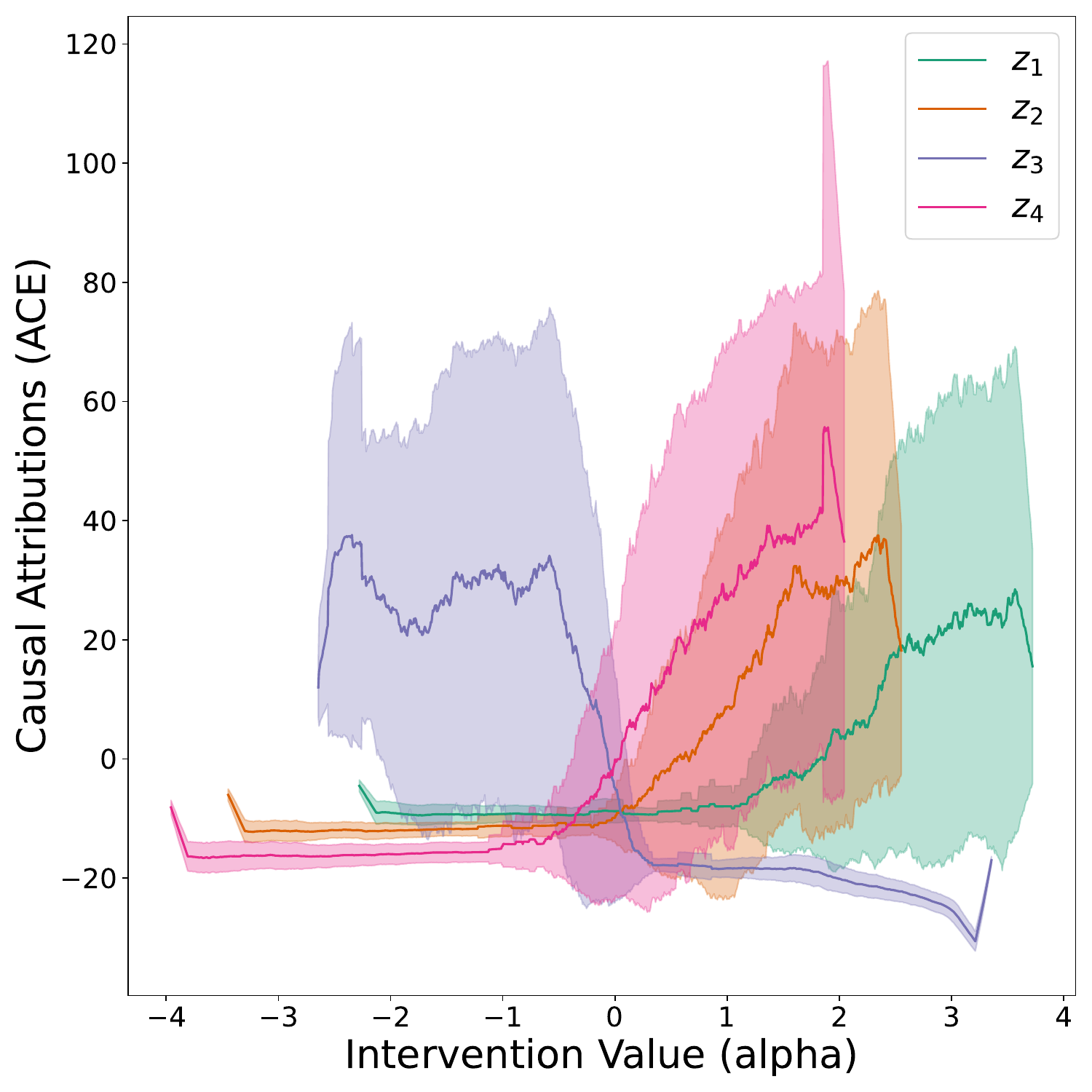}
\label{fig:ate_mt5_window_open}}
\quad
\subfigure[Window Close]{%
\includegraphics[width=0.16\linewidth]{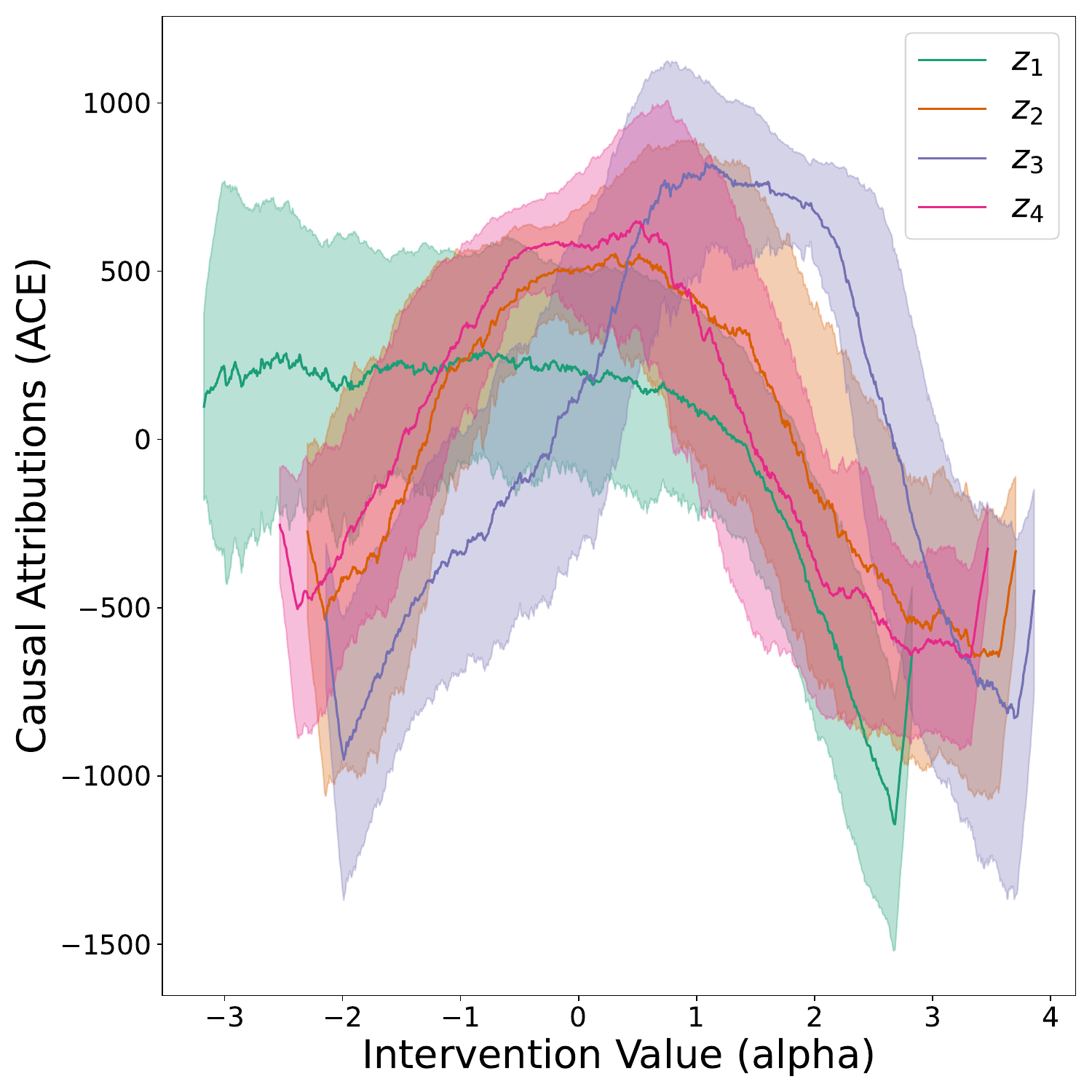}
\label{fig:ate_mt5_window_close}}
\quad
\subfigure[Drawer Open]{%
\includegraphics[width=0.16\linewidth]{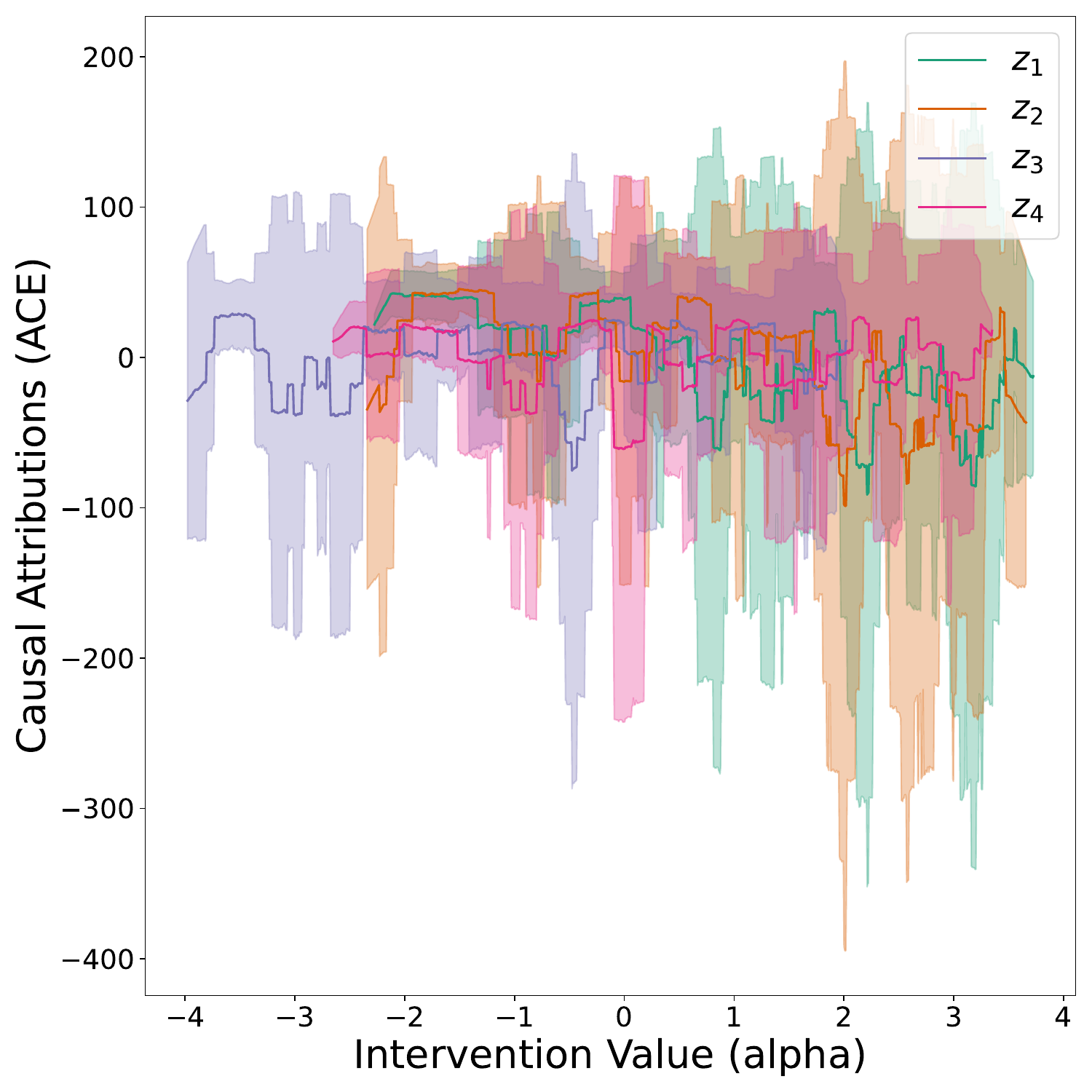}
\label{fig:ate_mt5_drawer_open}}
\quad
\subfigure[Drawer Close]{%
\includegraphics[width=0.16\linewidth]{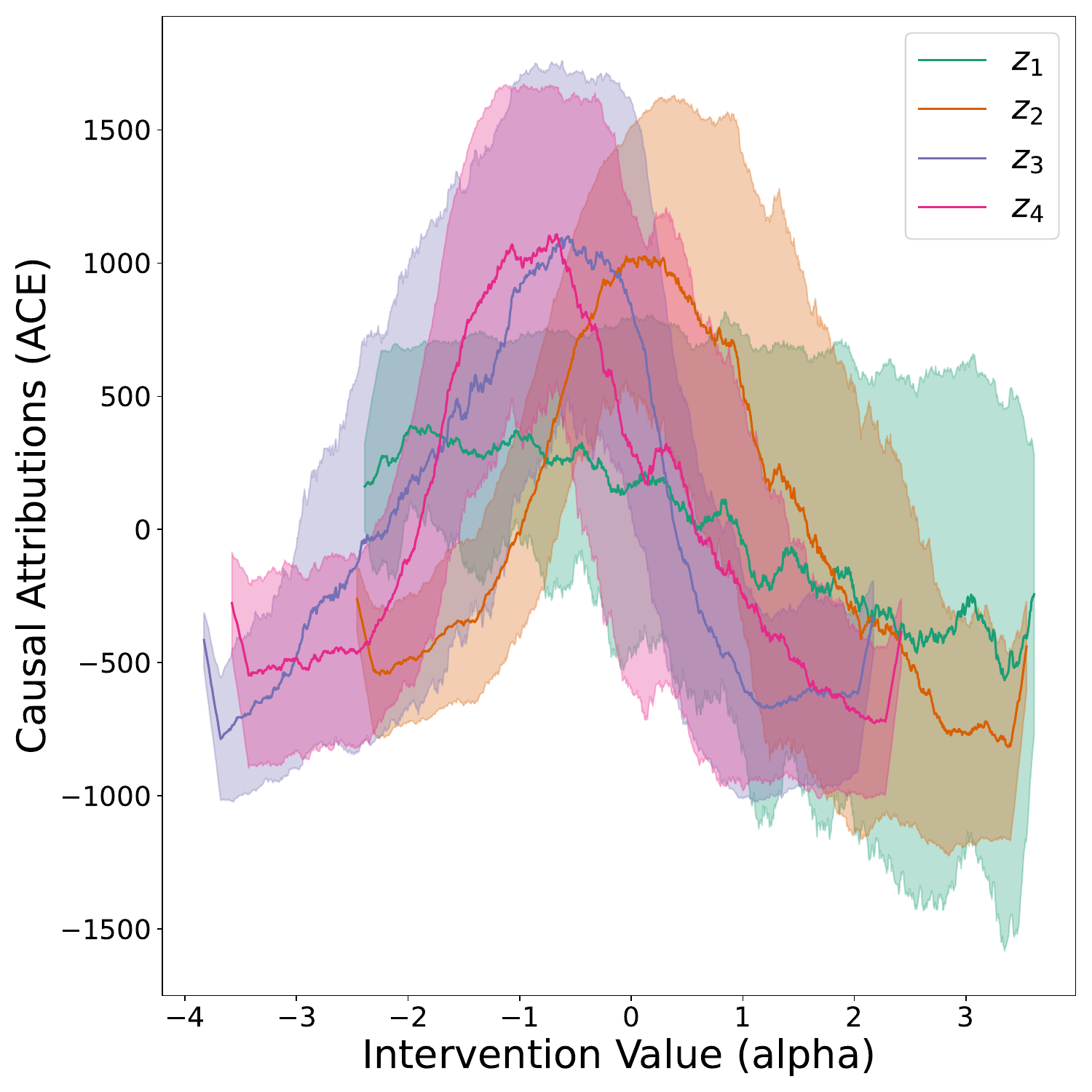}
\label{fig:ate_mt5_drawer_close}}

\caption{The Average Causal Effect of the latent skills for different tasks in the Metaworld (MT5) Environment. From the above plots, we notice that there is a lot of overlap and noise in the ACE. Furthermore, we notice that the ACE of the tasks: Push and Drawer Open, is roughly zero across all latents, and our model has prioritized learning other tasks.}
\label{fig:ace_mt5}
\end{figure}

\begin{figure}[hbt!]
\centering
\includegraphics[width=0.5\linewidth]{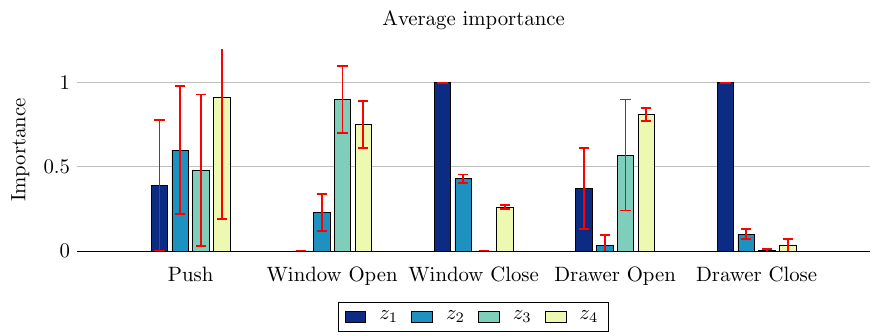}
\caption{The normalized importance of each latent feature on the Metaworld (MT5) tasks across 5 different seeds.}
\label{fig:imp_mt5}
\end{figure}

The average importance of each feature is shown in Figure~\ref{fig:imp_mt5}.

From the ACE plots, we can successfully assert that feature $z_2$ is an auxiliary variable and does not play any significant role in the model's behavior since the ACE value is close to 0. Furthermore, we notice that the $z_1$ variable is significant for tasks that involve closing, whereas $z_2$ and $z_3$ variable are significant when tasks involve the action of opening.

Performing an input-perturbation method, similar to previous sections might not be feasible for the MetaWorld environment, as it involves much higher complexity of observation and action space for a simple visualization.

\FloatBarrier

\newpage

\section{Model and Hyperparameters}
\label{hyperparam}
This section discusses the model and hyperparameters used in our experiments.

Table~\ref{tab:pm_hyper} depicts the hyperparameters we used for training our TE-PPO and ATE-PPO algorithm on the PointMass environment.

Similarly, Table~\ref{tab:nav_hyper} depicts the hyperparameters we used for training our TE-PPO and ATE-PPO algorithms in the 2-D navigation environment.

Finally, Table~\ref{tab:mt5_hyper} depicts the hyperparameters we used for training our TE-PPO and ATE-PPO algorithm in the Meta-World (MT5) environment.

\begin{table}[htbp]
    \centering
    \caption{Hyperparameters used for training TE-PPO and ATE-PPO on PointMass environment.}
    \vspace*{2mm}
    \label{tab:pm_hyper}
    \begin{tabularx}{0.9\textwidth}{bsss}
        \toprule
        \textbf{Description}     & \textbf{TE-PPO}     & \textbf{ATE-PPO} & \texttt{argument\_name}     \\ 
        \midrule
        General Hyperparameters \\
        \midrule
        Discount & $0.99$ & $0.99$ & \texttt{discount}\\
        Batch size & $4096$ & $4096$ & \texttt{batch\_size}\\
        Number of epochs & $600$ & $600$ & \texttt{n\_epochs}\\
        \midrule
        Algorithm-Specific Hyperparameters \\
        \midrule
        Encoder hidden sizes & $(20,20)$  & $(20,20)$ & \texttt{enc\_hidden\_sizes} \\
        Inference hidden sizes & $(20,20)$  & $(20,20)$ & \texttt{inf\_hidden\_sizes} \\
        Policy hidden sizes & $(32,16)$  & $(32,16)$ & \texttt{pol\_hidden\_sizes} \\
        Activation function of hidden layers & tanh & tanh & \texttt{hidden\_nonlinearity} \\
        Likelihood ratio clip range & $0.2$ & $0.2$ & \texttt{lr\_clip\_range} \\
        Latent dimension & $2$ & $4$ & \texttt{latent\_length} \\
        Inference window length & $6$ & $6$ & \texttt{inference\_window} \\ 
        Embedding maximum standard deviation & $0.2$ & $0.2$ &
        \texttt{embedding\_max\_std} \\
        Policy entropy coefficient & $1e^{-3}$ & $1e^{-3}$ & \texttt{policy\_ent\_coeff} \\
        Encoder entropy coefficient & $1e^{-3}$ & $1e^{-3}$ & \texttt{enc\_ent\_coeff} \\
        Inference entropy coefficient & $5e^{-2}$ & $5e^{-2}$ & \texttt{inf\_ent\_coeff} \\
        \midrule
        Optimizer-Specific Hyperparameters\\
        \midrule
        Protagonist mini-batch size & $32$ & $64$ & \texttt{pr\_batch\_size} \\
        Adversary mini-batch size & $-$ & $64$ & \texttt{ad\_batch\_size} \\
        Inference mini-batch size & $32$ & $64$ & \texttt{inf\_batch\_size} \\
        Protagonist learning rate & $1e^{-4}$ & $1e^{-3}$ & \texttt{pr\_lr} \\
        Adversary learning rate & $-$ & $1e^{-4}$ & \texttt{ad\_lr} \\
        Inference learning rate & $1e^{-3}$ & $1e^{-3}$ & \texttt{inf\_lr} \\
        \bottomrule
    \end{tabularx}
\end{table}

\begin{table}[htbp]
    \centering
    \caption{Hyperparameters used for training TE-PPO and ATE-PPO on 2-D Navigation environment.}
    \vspace*{2mm}
    \label{tab:nav_hyper}
    \begin{tabularx}{0.9\textwidth}{bsss}
        \toprule
        \textbf{Description}     & \textbf{TE-PPO}     & \textbf{ATE-PPO} & \texttt{argument\_name}     \\ 
        \midrule
        General Hyperparameters \\
        \midrule
        Discount & $0.99$ & $0.99$ & \texttt{discount}\\
        Batch size & $3072$ & $3072$ & \texttt{batch\_size}\\
        Number of epochs & $400$ & $400$ & \texttt{n\_epochs}\\
        \midrule
        Algorithm-Specific Hyperparameters \\
        \midrule
        Encoder hidden sizes & $(20,20)$  & $(20,20)$ & \texttt{enc\_hidden\_sizes} \\
        Inference hidden sizes & $(20,20)$  & $(20,20)$ & \texttt{inf\_hidden\_sizes} \\
        Policy hidden sizes & $(32,16)$  & $(32,16)$ & \texttt{pol\_hidden\_sizes} \\
        Activation function of hidden layers & tanh & tanh & \texttt{hidden\_nonlinearity} \\
        Likelihood ratio clip range & $0.2$ & $0.2$ & \texttt{lr\_clip\_range} \\
        Latent dimension & $4$ & $4$ & \texttt{latent\_length} \\
        Inference window length & $6$ & $6$ & \texttt{inference\_window} \\ 
        Embedding maximum standard deviation & $0.2$ & $0.2$ &
        \texttt{embedding\_max\_std} \\
        Policy entropy coefficient & $1e^{-3}$ & $1e^{-3}$ & \texttt{policy\_ent\_coeff} \\
        Encoder entropy coefficient & $1e^{-3}$ & $1e^{-3}$ & \texttt{enc\_ent\_coeff} \\
        Inference entropy coefficient & $5e^{-2}$ & $5e^{-2}$ & \texttt{inf\_ent\_coeff} \\
        \midrule
        Optimizer-Specific Hyperparameters\\
        \midrule
        Protagonist mini-batch size & $32$ & $64$ & \texttt{pr\_batch\_size} \\
        Adversary mini-batch size & $-$ & $32$ & \texttt{ad\_batch\_size} \\
        Inference mini-batch size & $32$ & $64$ & \texttt{inf\_batch\_size} \\
        Protagonist learning rate & $1e^{-4}$ & $5e^{-4}$ & \texttt{pr\_lr} \\
        Adversary learning rate & $-$ & $1e^{-4}$ & \texttt{ad\_lr} \\
        Inference learning rate & $1e^{-3}$ & $5e^{-4}$ & \texttt{inf\_lr} \\
        \bottomrule
    \end{tabularx}
\end{table}

\begin{table}[htbp]
    \centering
    \caption{Hyperparameters used for training TE-PPO and ATE-PPO on MT5 environment.}
    \vspace*{2mm}
    \label{tab:mt5_hyper}
    \begin{tabularx}{0.9\textwidth}{bsss}
        \toprule
        \textbf{Description}     & \textbf{TE-PPO}     & \textbf{ATE-PPO} & \texttt{argument\_name}     \\ 
        \midrule
        General Hyperparameters \\
        \midrule
        Discount & $0.99$ & $0.99$ & \texttt{discount}\\
        Batch size & $25000$ & $25000$ & \texttt{batch\_size}\\
        Number of epochs & $1000$ & $1000$ & \texttt{n\_epochs}\\
        \midrule
        Algorithm-Specific Hyperparameters \\
        \midrule
        Encoder hidden sizes & $(20,20)$  & $(20,20)$ & \texttt{enc\_hidden\_sizes} \\
        Inference hidden sizes & $(20,20)$  & $(20,20)$ & \texttt{inf\_hidden\_sizes} \\
        Policy hidden sizes & $(32,16)$  & $(32,16)$ & \texttt{pol\_hidden\_sizes} \\
        Activation function of hidden layers & tanh & tanh & \texttt{hidden\_nonlinearity} \\
        Likelihood ratio clip range & $0.2$ & $0.2$ & \texttt{lr\_clip\_range} \\
        Latent dimension & $4$ & $4$ & \texttt{latent\_length} \\
        Inference window length & $6$ & $6$ & \texttt{inference\_window} \\ 
        Embedding maximum standard deviation & $0.2$ & $0.2$ &
        \texttt{embedding\_max\_std} \\
        Policy entropy coefficient & $2e^{-}2$ & $2e^{-2}$ & \texttt{policy\_ent\_coeff} \\
        Encoder entropy coefficient & $2e^{-2}$ & $2e^{-2}$ & \texttt{enc\_ent\_coeff} \\
        Inference entropy coefficient & $5e^{-2}$ & $5e^{-2}$ & \texttt{inf\_ent\_coeff} \\
        \midrule
        Optimizer-Specific Hyperparameters\\
        \midrule
        Protagonist mini-batch size & $256$ & $256$ & \texttt{pr\_batch\_size} \\
        Adversary mini-batch size & $-$ & $256$ & \texttt{ad\_batch\_size} \\
        Inference mini-batch size & $256$ & $256$ & \texttt{inf\_batch\_size} \\
        Protagonist learning rate & $1e^{-3}$ & $5e^{-4}$ & \texttt{pr\_lr} \\
        Adversary learning rate & $-$ & $1e^{-4}$ & \texttt{ad\_lr} \\
        Inference learning rate & $1e^{-3}$ & $5e^{-4}$ & \texttt{inf\_lr} \\
        \bottomrule
    \end{tabularx}
\end{table}

\end{document}